\documentclass[journal]{IEEEtran}

\usepackage{calc,amsfonts,amssymb,amsmath,bm,url,color,graphicx,cite,epstopdf,nicefrac,bbold, amsthm}
\usepackage{psfrag,subfigure,float}
\usepackage[algoruled,linesnumbered]{algorithm2e}
\usepackage{multirow}

\usepackage{algorithmic}
\usepackage[normalem]{ulem}
\usepackage{stmaryrd}
\usepackage{xcolor}
\usepackage{psfrag,framed,mdframed}
\usepackage{comment}
\usepackage{mdframed}
\usepackage{cleveref}
\usepackage[utf8]{inputenc}

\usepackage{mathtools}

\definecolor{orange}{RGB}{255,107,0}
\definecolor{green}{RGB}{0,100,0}
\definecolor{purple}{RGB}{255, 0, 255}

\makeatletter
\newcommand{\nosemic}{\renewcommand{\@endalgocfline}{\relax}}% Drop semi-colon ;
\newcommand{\dosemic}{\renewcommand{\@endalgocfline}{\algocf@endline}}% Reinstate semi-colon ;
\let\oldnl\nl% Store \nl in \oldnl
\newcommand{\nonl}{\renewcommand{\nl}{\let\nl\oldnl}}% Remove line number for one line
\makeatother

\newlength\myindent
\usepackage{steinmetz}

\newcommand{\g}{\boldsymbol{g}}

\newcommand{\q}{\boldsymbol{q}}

\renewcommand{\P}{\boldsymbol{P}}

\newcommand{\x}{\boldsymbol{x}}

\newcommand{\btheta}{\boldsymbol{\theta}}
\newcommand{\bLambda}{\boldsymbol{\Lambda}}
\newcommand{\blambda}{\boldsymbol{\lambda}}

\newcommand{\zero}{\boldsymbol{0}}
\newcommand{\one}{\boldsymbol{1}}
\newcommand{\I}{\boldsymbol{I}}

\newcommand{\W}{\boldsymbol{W}}
\newcommand{\Y}{\boldsymbol{Y}}
\newcommand{\G}{\boldsymbol{G}}
\newcommand{\Q}{\boldsymbol{Q}}
\newcommand{\X}{\boldsymbol{X}}

\newcommand{\E}{\boldsymbol{E}}
\newcommand{\U}{\boldsymbol{U}}
\renewcommand{\H}{\boldsymbol{H}}
\newcommand{\M}{\boldsymbol{M}}
\newcommand{\A}{\boldsymbol{A}}

\newcommand{\V}{\boldsymbol{V}}

\newcommand{\Z}{\boldsymbol{Z}}

\newcommand{\T}{{\!\top\!}}
\newcommand{\cB}{\mathcal{B}}
\newcommand{\cC}{\mathcal{C}}

\newcommand{\cE}{\mathcal{E}}
\newcommand{\cF}{\mathcal{F}}

\newcommand{\cO}{\mathcal{O}}

\newcommand{\cQ}{\mathcal{Q}}
\newcommand{\cR}{\mathcal{R}}

\newcommand{\bDelta}{\bm{\varDelta}}

\newcommand{\bbR}{\mathbb{R}}
\newcommand{\bbE}{\mathbb{E}}

\DeclareMathOperator*{\minimize}{\textrm{minimize}}

\newtheorem{theorem}{Theorem}
\newtheorem{assumption}{Assumption}
\newtheorem{fact}{Fact}
\newtheorem{lemma}{Lemma}

\newtheorem{definition}{Definition}

\usepackage{xcolor}
\usepackage{psfrag,framed}
\usepackage{lipsum}
\usepackage{chapterbib}
\PassOptionsToPackage{normalem}{ulem}
\usepackage{ulem}
\definecolor{shadecolor}{RGB}{220,220,220}
\begin{document}

% \begin{bibunit}[IEEEtran]
	\title{Communication-Efficient {Federated} Linear and Deep Generalized Canonical Correlation Analysis}
	
	\author{Sagar Shrestha and Xiao Fu
		\thanks{
		This work is supported in part by the Army Research Office (ARO) under project W911NF-21-1-0227, and the National Science Foundation (NSF) under project NSF-ECCS 1808159.
		
		The authors are with the School of Electrical Engineering and Computer Science, Oregon State University, OR 97331, USA. Email: (shressag,xiao.fu)@oregonstate.edu}}
	
	\maketitle
	\begin{abstract}
		Classic and deep learning-based generalized canonical correlation analysis (GCCA) algorithms seek low-dimensional common representations of data entities from multiple ``views'' (e.g., audio and image) using linear transformations and neural networks, respectively. {When the views are acquired and stored at different computing agents (e.g., organizations and edge devices) and data sharing is undesired due to privacy or communication cost considerations, {\it federated learning}-based GCCA is well-motivated. In federated learning, the views are kept locally at the agents and only derived, limited information exchange with a central server is allowed. However, applying existing GCCA algorithms onto such federated learning setting may incur prohibitively high communication overhead.}
		This work puts forth a communication-efficient {federated learning} framework for both linear and deep GCCA under the maximum variance (MAX-VAR) {formulation}.
		The overhead issue is addressed by aggressively compressing (via quantization) the exchanging information between the computing agents and a central controller. Compared to the unquantized version, {our empirical study shows that} the proposed algorithm { enjoys a substantial reduction of communication overheads} with virtually no loss in accuracy and convergence speed. 
		Rigorous convergence analyses are also presented, which is a nontrivial effort. {Generic federated optimization results do not} cover the special problem structure of GCCA, {which is a manifold constrained multi-block nonconvex eigen problem}.
		Our result shows that the proposed algorithms for both linear and deep GCCA converge to critical points in a sublinear rate, even under heavy quantization and stochastic approximations. 
		In addition, in the linear MAX-VAR case, the quantized algorithm approaches a {\it global optimum} in a {\it geometric} rate under reasonable conditions. Synthetic and real-data experiments are used to showcase the effectiveness of the proposed approach.
	\end{abstract}
	
	\IEEEpeerreviewmaketitle

	\section{Introduction}
	{\it Canonical correlation analysis} (CCA) aims at learning low-dimensional latent representations from two different ``views"  of data entities---e.g., an image and an audio clip of a cat are considered two different views of the entity ``cat''. 
	The benefits of using more than one view to learn representations of data entities have been long recognized by the machine learning and signal processing communities, which include robustness against strong but nonessential components in the data{ \cite{lyu2020nonlinear,ibrahim2020reliable, lyu2021understanding}}  and resilience to unknown colored noise \cite{bach2005probabilistic}.
	A natural extension of CCA is to leverage information from more than two views, which leads to the technique called {\it generalized canonical correlation analysis} (GCCA) \cite{horst1961generalized}. Intuitively, increasing the number of views of the same entities should make it easier to distinguish common information from the irrelevant details that are specific to the views. This has also found theoretical and empirical supports; see, e.g., \cite{sorensen2021generalized}. 
	
	Classic GCCA is restricted to learning a linear transformation for each view. In recent years, such {\it linear GCCA} formulations were extended to the nonlinear regime---e.g., by using kernel functions or deep neural networks to replace the linear operators. Particularly, the line of work called {\it deep GCCA} in the literature has shown promising results \cite{andrew2013deep, benton2019deep,lyu2020nonlinear}.
	GCCA has many different criteria in the literature, e.g., \textit{sum of correlations} (SUMCORR), \textit{maximum variance} (MAX-VAR), \textit{minimum variance} (MIN-VAR), just to name a few \cite{kettenring1971canonical}. Among these criteria, linear transformation-based MAX-VAR is considered more ``computation-friendly'' due to its tractable nature and promising performance in many applications \cite{rastogi2015multiview, fu2017scalable, benton2019deep}. The deep neural network-based MAX-VAR also enjoys computational convenience due to its special constraint structure; see \cite{lyu2020nonlinear}.

	{Triggered by the big data deluge and the ubiquity of multiview data in modern machine learning, there has been a renewed interest in GCCA, particularly, efficient computation of GCCA in the presence of a large number of big views; see{ \cite{rastogi2015multiview, fu2017scalable,benton2019deep,fu2018efficient,kanatsoulis2018structured,fu2016efficient}}. 
	{ To advance large-scale multiview analysis to accommodate modern machine learning scenarios, considering federated learning-based GCCA is particularly meaningful. Here, federated learning refers to a special distributed computing paradigm where different organizations or edge devices collaboratively train statistical model using locally held data, often with the help of a central server \cite{kairouz2021advances, li2020federated}. Federated learning is considered a core machine learning paradigm in the era of mobile computing, which was first proposed in \cite{konevcny2016federated} to train machine learning models by keeping data ``local'' in edge devices.
    Such federated settings of GCCA are also} well-motivated when the views are acquired and stored by individual organizations or devices, while data sharing is either costly or not allowed due to legal/privacy restrictions. 
	
    \noindent
    {\bf Communication Overhead Challenge.}	
	In recent years, a number of distributed linear GCCA algorithms appeared in the literature \cite{fu2017scalable, fu2018efficient,  bertrand2015distributed, hovine2021distributed}. {Their settings often have a central node to coordinate among the computing agents, and thus are closely related to federated learning.}
	The algorithm in \cite{lyu2020nonlinear} for a deep MAX-VAR GCCA criterion can also be implemented in a distributed fashion.
	These methods are plausible, since the major computations are carried out in the computing agents and only derived information from data is exchanged in the iterations---which circumvents raw data exchange. However, there are also critical challenges remaining.
	In particular, when the views are of large size, exchanging such derived information may still entail large communication overhead. 
	High communication overhead {may lead to many issues such as high latency, costly communication data consumption, and fast drainage of device batteries. These issues} hinder the possibility of using {federated} GCCA in important wireless scenarios, e.g., edge computing and internet of things (IoT). 
	
	{ In this work, we propose a quantization-based algorithmic framework for communication-efficient federated GCCA.} 
    Designing {such a framework} with convergence guarantees turns out to be a nontrivial task. 
	A natural idea is to employ some existing general-purpose distributed optimization frameworks with information compression or quantization, {e.g., \cite{rabbat2005quantized, nedic2008distributed, pu2017quantization, yuan2012distributed, bernstein2018signsgd, alistarh2017qsgd, stich2018sparsified, basu2019qsparse, karimireddy2019error, reisizadeh2020fedpaq, rumelhart1986learning, shlezinger2021uveqfed}.
	However, directly using these frameworks for linear/deep GCCA algorithms may not be viable. 
	These frameworks deal with unconstrained or convex set-constrained generic optimization problems, yet GCCA has a special multi-block manifold-constrained eigen problem structure.}
	Hence, tailored algorithm design and custom convergence analyses are needed for communication-efficient {federated} GCCA. 

	}
	
		\noindent
	\textbf{Contributions.} {Our interest lies in
	designing a federated learning algorithm tailored to accommodate GCCA's special problem structure, while ensuring certain convergence guarantees.}
	The main contributions are summarized as follows:
	
	\noindent
	$\bullet$ \textbf{Algorithm Design} We design a communication-efficient computational framework for both linear and deep MAX-VAR GCCA.  Under the proposed framework, each computing agent stores a (potentially large) view that is not shared with other agents. In each iteration, the agent is responsible for updating the view-specific transformation operators (e.g., a transformation matrix or a neural network) locally. The updated latent representations of the views are then compressed and sent to a central node, where simple operations are performed. Under our design, the manifold constraint of GCCA is always respected. {Our empirical study shows that the resulting framework reduces the communication overhead {substantially}} 
    ---with virtually no accuracy or speed losses compared to the unquantized version.

	\noindent
	$\bullet$ \textbf{Convergence Analysis} 
    We present custom convergence analyses for communication-efficient MAX-VAR GCCA under both the linear and the deep transformation operators. 
    We show that, for both cases, our method converges to a critical point and can attain a sublinear rate. Our technical approach leverages the {\it error feedback} (EF) mechanism and the so-called $\delta$-compressors for quantizating the exchanged information. Both EF and $\delta$-compressors were proposed for gradient compression unconstrained single-block distributed optimization \cite{basu2019qsparse, karimireddy2019error}. Nonetheless, we show that using them to compress learned representations other than gradients still guarantees convergence, even for multi-block manifold-constrained GCCA problems.
	Moreover, we show that for the linear GCCA case, the algorithm converges to a neighborhood of the {\it global optimal solution} with a {\it geometric} convergence rate. 
	Notably, existing gradient compression techniques often only compress uplink information (see \cite{bernstein2018signsgd, alistarh2017qsgd, basu2019qsparse}), but our algorithm compresses both uplink and downlink transmissions with convergence assurances---which is even more economical in terms of overhead.

\smallskip	
Part of the work was {published at} IEEE ICASSP 2022 \cite{icassp2022submission}. The conference version focused on linear MAX-VAR GCCA, {where the convergence theorem (regarding the global optimality of the linear case) was stated {\it without} proof. 
The journal version has the following substantial additions:
(i) The detailed global optimality analysis of the proposed federated linear MAX-VAR GCCA algorithm is presented.
(ii) A federated deep GCCA algorithm design is proposed. (iii) A stationary-point convergence analysis with convergence rate characterizations for the deep GCCA case is presented.
(iv) More synthetic and real data experiments are also included.}
	
\noindent
    	\textbf{Related Work}.
    We should mention that many multimodal learning frameworks exist beyond GCCA, e.g., deep generative models based multimodal learning \cite{srivastava2012learning, wu2018multimodal, vedantam2017generative, tsai2019learning},  multimodal contrastive learning \cite{oord2018representation}, multimodal autoencoders \cite{ngiam2011multimodal}, multimodal deep belief networks \cite{suzuki2016joint}, etc.
    Instead of exploring a wide spectrum of multimodal models, this work focuses on GCCA. Our effort is motivated by its wide applicability to many important tasks, e.g., multilingual word embedding \cite{rastogi2015multiview}, sensor fusion in wireless sensor networks \cite{hovine2021distributed}, self-supervised representation learning \cite{liu2021self}, multimodal representation learning \cite{guo2019deep}. In addition, the GCCA problem presents an interesting and challenging optimization problem with manifold constraints, whose computational aspects are worth studying. Nonetheless, the algorithm design principles of our method can be used for other multimodal models with proper modifications---which we leave for future work.

	\noindent
    	\textbf{Notation}. {$x$, $\x$, and $\X$ denote a scalar, a vector and a matrix, respectively. $\x_i$ and $\X(i,:)$ denote the $i$th column and the $i$th row of $\X$, respectively. Both $\X(i,j)$ and $[\X]_{i,j}$ denote the $(i,j)$-th element of $\X$. $\|\X\|_2$, $\|\X\|_{\rm F}$ and $\| \X\|_{\rm max}$ denote the spectral norm, the Frobenius norm, and the maximum absolute value of the elements of $\X$, respectively. $\sigma_{\rm max}(\X)$ and $\sigma_{\rm min}(\X)$ denote the largest and the smallest singular values of $\X$, respectively. $\X^\T$, $\X^\dagger$, and ${\rm Tr}(\X)$ denote the transpose, the pseudo-inverse, and the trace of $\X$, respectively. ${\rm vec}(\X) = [\x_1^\T \dots \x_N^\T]^\T $ denotes the vectorized version of $\X \in \bbR^{M \times N}$.  ${\rm sgn}(x)$ denotes the sign operator which returns $+1$ if $x \geq 0$ and $-1$ otherwise. $\bbE[\cdot]$ denotes the expectation operator, $\bbE[X|Y]$ the conditional expectation of $X$ given $Y$, and $\bbE_{ab}[\cdot] = \bbE_a[\bbE_b[\cdot]]$. $<\cdot, \cdot>$ denotes the matrix inner product. $\cR(\X)$ denotes the range space of $\X$. For any $I \in \mathbb{N}$, $[I] = \{ 1, 2, \dots, I\}$.
        }
        \color{black}

	\section{Background}
	In this section, we briefly introduce the preliminaries of the problem of interest.
	
	\subsection{Classic Linear GCCA under MAX-VAR Criterion}
	Let $\X_i \in \bbR^{J \times N_i}$ be the view held by the $i$th computing agent (or node), where $i=1,\ldots,I$ and $I$ is the number of views, $J$ is the number of data entities, and $N_i$ represents the feature dimension of the $i$th view. The row vector $\X_i (j, :)$ denotes $i$th view of the $j$th data entity. Ideally, $\X_i (j, :)$ and $\X_{i'} (j, :)$ for $i\neq i'$ are expected to share the same learned latent representation, since they are two views of the same entity. The goal of GCCA is to find such $K$-dimensional shared representations across views for all the entities. In classic two-view CCA, this is achieved via solving the following optimization criterion \cite{hardoon2004canonical,lu2014large,ge2016efficient,hotelling1992relations}:
	\begin{align}\label{eq:sumcor_gcca}
		\minimize_{\Q_1, \Q_2} \quad & \| \X_1 \Q_1 - \X_2 \Q_2 \|_{\rm F}^2,  \nonumber\\
		\text{subject~to } & \Q_i^{\T} \X_i^{\T} \X_i \Q_i = \bm I,  \quad i \in \{1,2\},
	\end{align}
	where $\Q_i\in\mathbb{R}^{ N_i\times K}$ for $i=1,2$ are two linear operators that transform $\X_i(j,:)\in\mathbb{R}^{1\times N_i}$ for $i=1,2$ such that the resulting $\X_i(j,:)\Q_i\in\mathbb{R}^{1\times K}$ are as similar as possible. The constraint is to prevent trivial solutions, e.g., $\Q_i=\bm 0$, from happening. 
	When $I>2$ views are present, a {\it generalized} CCA (GCCA) formulation is to change the objective function to
	$\sum_{i\neq j}  \| \X_i \Q_i - \X_j \Q_j \|_{\rm F}^2,$
	which is the so-called \textit{sum-of-correlations} (SUMCORR) formulation for GCCA \cite{hardoon2004canonical,fu2018efficient}. The SUMCORR formulation has been shown to be NP-hard \cite{rupnik2013comparison}. A more tractable formulation is MAX-VAR GCCA. The idea of MAX-VAR 
	{ is to introduce a slack variable $\G$ and rewrite the shared representation-finding problem as follows:}
    \begin{align}\label{eq:maxvar_gcca}
		\minimize_{\{\Q_i\}_{i=1}^I, \G} \quad & \sum_{i=1}^I \frac{1}{2}\left\| \X_i \Q_i - \G \right\|_{\rm F}^2,  \nonumber\\
		\text{subject~to } &  \G^\T \G = {\bm I}.
    \end{align} 
	{ Let} $\P = \sum_{i=1}^I \X_i \X_i^\dagger$, then the optimal solution, $\G_{\rm opt}$, is obtained via extracting the first $K$ principal eigenvectors of $\P$ \cite{golub2013matrix,fu2017scalable}.
	This solution is not easy to implement when the views are large, but efficient and scalable algorithms were proposed to tackle this challenge; see, e.g., \cite{fu2017scalable,rastogi2015multiview}. The algorithm in \cite{fu2017scalable} can be summarized as follows:
		\begin{subequations}\label{eq:oldalgo}
			\begin{align}
				\Q_i^{(r+1)} &\leftarrow \arg\min_{\Q_i} \frac{1}{2}\left\| \X_i \Q_i - \G^{(r)} \right\|_{\rm F}^2,~\forall i\in[I],\label{eq:oldQupdate}\\
				\G^{(r+1)} &\leftarrow \arg\min_{\G^\T\G=\bm I}~\sum_{i=1}^I\left\|\X_i \Q_i^{(r+1)} - \G \right\|_{\rm F}^2\label{eq:oldGupdate}.
			\end{align}
		\end{subequations}
		The work in \cite{fu2017scalable} showed that {the above} operations are fairly scalable if $\X_i$'s are sparse.
	
	\subsection{Deep GCCA}\label{sec:background_dgcca}
	Many attempts were made towards nonlinear (G)CCA for enhanced expressive power, by replacing $\Q_i$'s with nonlinear operators such as kernel functions \cite{akaho2001kernel, hardoon2004canonical} and deep neural networks (DNNs) \cite{andrew2013deep,benton2019deep,lyu2020nonlinear,wang2015deep}. In particular, deep (G)CCA frameworks have drawn considerable attention due to a good balance between efficiency and effectiveness. In \cite{benton2019deep}, the MAX-VAR GCCA formulation is integrated with DNNs as follows. Let us define {an operator $\q: \bbR^{N_i} \to \bbR^K$ for view $i$ such that: 
	\begin{align}\label{eq:notation_transf}
		\bm z_j^{(i)}={\bm q}(\bm x_j^{(i)}; \btheta_i),
	\end{align}
	where $\x_j^{(i)}=\X_i(j,:)^\T$.} In deep GCCA, this operator is a deep neural network, i.e.,
	\begin{align}
		({\sf Deep~Case}) \quad  \bm z_j^{(i)}=   \bm \sigma\left(\bm W_L^{(i)} \ldots \bm \sigma\left( {\bm W}^{(i)}_1\bm x_j^{(i)}\right)\right),
	\end{align}
	\begin{figure}[t]
		\centering
		\includegraphics[width=0.7\linewidth]{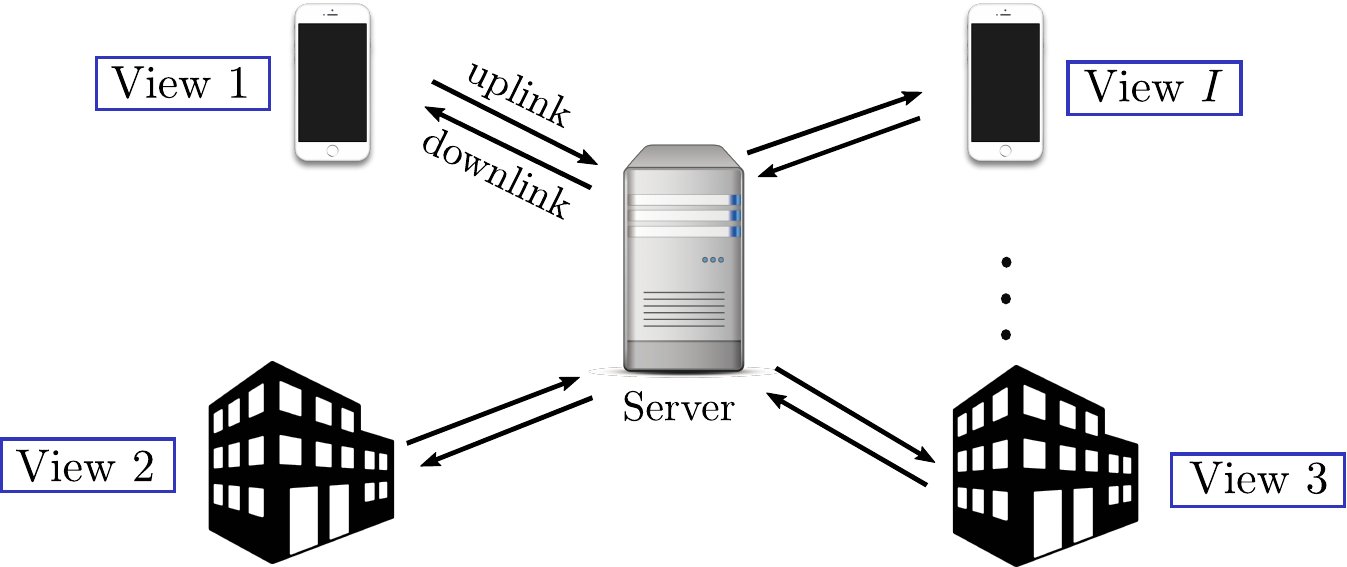}
		\caption{Illustration of the {federated} GCCA scenario.  }
		\label{fig:scenario}
	\end{figure}
	where $\bm W^{(i)}_\ell$ is the linear mixing system of the $\ell$th layer and $\bm \sigma(\x)$ imposes a nonlinear activation function [e.g., sigmoid and rectified linear unit (ReLU)] on each dimension of $\x$ and
	$  \bm \theta_i =[ {\rm vec}(\W_1^{(i)})^\T,\ldots,{\rm vec}(\bm W_L^{(i)})^\T ]^\T. $
	Note that the same notation in \eqref{eq:notation_transf} can also represent the linear operator, in which we have
	\begin{align}
		({\sf Linear~Case}) \quad  \bm z_j^{(i)}=   \Q_i^\T\x_j^{(i)},
	\end{align}
	and $\bm \theta_i={\rm vec}(\bm Q_i)$.

	With the notations defined, the classic MAX-VAR formulation in \eqref{eq:maxvar_gcca} can be generalized as follows:
	\begin{subequations}\label{eq:maxvar_general}
		\begin{align}
			\minimize_{\G, \{\btheta_i\}_{i=1}^I} \quad &  \sum_{i=1}^I \frac{1}{2} \left\| {\cal Q }(\X_i;\bm \theta_i) - \G \right\|_{\rm F}^2,  \\
			\text{subject~to } \quad &\G^{\T} \G = \bm I, \quad \nicefrac{\G^\T \one}{J} = \mathbf{0},\label{eq:zeromean}
		\end{align}  
	\end{subequations}
	where we have ${\cal Q }(\X_i;\bm \theta_i) = \left[\bm z_1^{(i)},\ldots,\bm z_J^{(i)}\right]^\T$.
    { Note that if the views are centered (i.e., $\bm 1^\T\X_i/J=\bm 0$), the constraint $\mathbf{1}^\T \G/J = \mathbf{0}$ is only used in the deep GCCA case to avoid trivial solutions, e.g., constant latent representations \cite{lyu2020nonlinear}. }
	{ This is because in the linear case the zero mean of $\X_i$ is preserved in $\X_i\Q_i$, whereas DNNs are capable of shifting the mean of ${\cal Q}(\X_i;\bm \theta_i)$ far away from zero. }

	\subsection{Federated GCCA: Motivations and Challenges}
	When the views are acquired and stored by different organizations, individuals, or devices, the {raw} data {(i.e., $\X_i$'s)} may be sensitive/costly to be transmitted to and collected by a central controller. 
    Such scenarios arise in machine learning and signal processing due to the ubiquitous data acquisition by edge devices and entities, and are often coped with by the {\it federated learning} paradigm \cite{li2020federated, kairouz2021advances}. In federated learning, raw data sharing is not allowed. Only derived information (e.g., model parameters or their gradients) is exchanged among the computing agents and a central controller. 
    {An array of motivating examples for federated GCCA is as follows:
    
    \noindent 
    $\bullet$ \textbf{Electronic Health Record (EHR) Data Analysis}: EHR datasets are collected locally at medical institutions and contain sensitive information. Federated learning is advocated for handling EHR data for privacy reasons; see, e.g., \cite{rieke2020future, xu2021federated, ma2021communication}. 
    Data collected at various medical institutions can naturally stand for different views of entities of interest (e.g., the medicines). Hence, federated GCCA can be used to learn entity representations---which could benefit tasks such as medicine/prescription recommendation. 
    
    \noindent $\bullet$ \textbf{Data Fusion in Wireless Sensor Networks (WSNs)}: In WSN, spatially dispersed sensors are equipped with diverse sensing devices. These sensors provide multiple views of events or objects of interest. Federated GCCA is considered a useful tool for fusing such multiple data in WSNs
    \cite{bertrand2015distributed, hovine2021distributed}.
    
    \noindent $\bullet$ \textbf{Cross-Platform Recommender Systems}: Different platforms---such as Facebook, Twitter, and TikTok---capture different activities of the same users. This creates multiple views of the users stored at the platforms.
    Using federated GCCA, one can take advantage of such cross-platform multiview data to learn representations of the users. Such representations reflect the users' invariant interest across multiple platforms, and thus may improve the performance of recommender systems \cite{zhu2021cross, berkovsky2007cross}.
    
    \noindent $\bullet$ \textbf{Parallel Computing}: 
    Federated learning can also be considered as a parallel computing paradigm \cite{tang20211}, as every computing agent (e.g., a CPU) processes their data simultaneously (on parallel) under federated learning frameworks. This was considered for training neural networks with large-scale data using multiple CPU cores \cite{teerapittayanon2017distributed}.
    The same idea can help accelerate GCCA.
    For large-scale GCCA problems,
    each computing node can store (part of) a data view and communicate with a master node for further updating. In such parallel computation frameworks, the communication overhead between the agents and the master is often the performance bottleneck \cite{tang20211}.
    }	

    Fig. \ref{fig:scenario} depicts a scenario where {federated} GCCA is well-motivated. We assume that $i$th view is held by the $i$th node (computing agent)---where a node can be a device or an organization. All nodes carry out their local computations and exchange {\it derived} information (other than the original data) with a central server that is capable of simple computations, e.g., aggregation, subtraction and SVD. The communication from a node to the server is called \textit{uplink} communication and opposite direction is called \textit{downlink} communication. The goal of {federated} GCCA is to learn the parameters $\btheta_i$ for all $i$ and $\G$. In an iterative {federated learning procedure}, the nodes and the server exchange information in every iteration.
    Hence, one of the most important considerations under this computing paradigm is the communication overhead between the nodes and the server. {In each of the motivating examples, high communication overheads between the nodes and the server may incur undesired communication latency, which hinder the system performance severely, especially when computational resources (e.g., bandwidth, energy and time) are limited; see more discussions in \cite{ma2021communication, tang20211, hovine2021distributed}.}

   In this work, our goal is to propose a {\it communication-efficient} algorithmic framework for MAX-VAR GCCA---for both the classic and the deep learning versions. We also aim to offer rigorous performance characterizations for the algorithm. 
	
	\section{Proposed Approach}\label{sec:proposed}
	As mentioned in \cite{fu2017scalable}, the algorithm in Eq.~\eqref{eq:oldalgo} is natural for {federated} MAX-VAR GCCA. However, the method may entail large communication overhead if full precision information is exchanged between the nodes and the server.
	In this work, our idea is to avoid transmitting such full precision messages but use coarsely quantized information for communications.

\subsection{A {Federated Learning} Framework for Problem \eqref{eq:maxvar_general}}
To see our idea, we first extend the algorithm in \eqref{eq:oldalgo} to cover both the deep GCCAcases and describe both cases as a unified framework. For the formulation in \eqref{eq:maxvar_general}, using the alternating optimization idea as in \eqref{eq:oldalgo}, the updates can be summarized as follows:
        \begin{subequations}\label{eq:generalalgo}
            \begin{align}
                \bm \theta_i^{(r+1)} \leftarrow \arg\min_{\bm \theta_i} \frac{1}{2}\left\| {\cal Q}(\X_i;\bm \theta_i) - \G^{(r)} \right\|_{\rm F}^2,~\forall i\in[I],\label{eq:generalQupdate} \\
                \G^{(r+1)} \leftarrow \arg\min_{ \substack{\G^\T\G=\bm I,\\ \bm 1^\T\bm G/J=\bm 0}}~\sum_{i=1}^I\left\|{\cal Q}(\X_i;\bm \theta_i^{(r+1)}) - \G \right\|_{\rm F}^2\label{eq:generalGupdate}.
            \end{align}
	\end{subequations}
	Note that the subproblem in \eqref{eq:generalQupdate} may not be solvable in the deep GCCA case. Even in the linear GCCA case, exactly solving the linear least squares problem may take too much time. Reasonable approximations may be a number of (stochastic) gradient descent iterations, as one will see later. In the deep GCCA case, the gradient w.r.t. $\bm \theta_i$ (i.e., the neural network weights) can be computed using back-propagation \cite{rumelhart1986learning}.
	
	The optimal solution to the $\G$-subproblem can be obtained using SVD, i.e.,
	\begin{equation}\label{eq:Gupdate}
		\G^{(r+1)} \leftarrow \U_{\Y^{(r+1)}} \V_{\Y^{(r+1)}}^\T,    
	\end{equation}
	where $~\bm U_{\Y^{(r+1)}}\bm \Sigma_{\Y^{(r+1)}} \V_{\Y^{(r+1)}}\leftarrow\texttt{svd}(\Y^{(r+1)},'\texttt{econ}')$ and
	\begin{equation}
		\Y^{(r+1)} := \sum_{i=1}^I \left( \I - \frac{1}{J} \mathbf{1} \mathbf{1}^\T \right) \M^{(r+1)} ,
	\end{equation} 
	where $\texttt{svd}(\cdot,'\texttt{econ}')$ means the thin SVD,
	see \cite[Lemma 1]{lyu2020nonlinear} for a proof. Note that in the linear case, $\cQ( \X_i; \btheta_i^{(r+1)} )=\X_i\Q_i$ is often centered for all $i$ via pre-processing and thus $\bm 1^\T\bm G/J=\bm 0$ is not needed. The above update boils down to the same update in \eqref{eq:oldGupdate}. 
	
	The algorithm in \eqref{eq:generalalgo} can be implemented in the following distributed manner:
	\begin{mdframed}
	\vskip -1\baselineskip
		\begin{align*}
			&{\sf node:~computes}~\eqref{eq:generalQupdate}~{\sf sends}~{\cal Q}(\X_i;\bm \theta_i^{(r+1)});\\
			&{\sf server}:~{\sf computes~}\eqref{eq:generalGupdate}~{\sf broadcasts}~{\bm G}^{(r+1)}.
		\end{align*}
	\end{mdframed}
	{In each iteration, a node communicates its transformed view ${\cal Q}(\X_i;\bm \theta_i^{(r+1)})$ to the server. Note that the transmission needs $\cO(JKq_{\rm full})$ bits, where $q_{\rm full}$ is the number of bits used to represent a real number in a full-resolution floating point system (typically, $q_{\rm full}=32$ or higher). The server broadcasts common representation $\bm G$ to each node, which costs $\cO(JKq_{\rm full})$ bits. } 
    In big data analytics, {such information exchange is often not affordable. Consider a case where $J=1,000,000$ and $K=300$}---which is not uncommon in large-scale data analytics like multilingual embedding \cite{fu2017scalable,fu2018efficient}. {Then, about} $1.2$GB of data needs to be exchanged between each node and the server in each iteration, if double precision {(i.e., $q_{\rm full}=32$)} is used for real numbers.
	This is already costly, let alone the fact that the algorithm often runs for hundreds of iterations.

	\subsection{Reducing Communication Cost}
	Our idea for reducing the communication cost of the algorithm in \eqref{eq:generalalgo} is by quantizing the uplink and downlink information. 
	
	In iteration $r$, node $i$ updates $\btheta_i^{(r)}$ to obtain $\btheta_i^{(r+1)}$. We wish to quantize $\cQ( \X_i; \btheta_i^{(r+1)})$ before sending it to the server. This may create large quantization error. Instead, a better approach is to quantize and transmit the {\it change} in $\cQ(\X_i; \btheta_i)$, i.e., $\cQ( \X_i; \btheta_i^{(r+1)}) - \cQ( \X_i; \btheta_i^{(r)} )$. 
	{ 
    The reason is intuitive: As the algorithm converges, one can expect that the change of $\cQ(\X_i; \btheta_i)$ to vanish, which in turn makes the compressed version of the change of $\cQ(\X_i; \btheta_i)$ to vanish.  This means that the ``compression error'' of the change of $\cQ(\X_i; \btheta_i)$ will approach zero upon convergence of the algorithm.}
	However, directly compressing such changes may sometimes be too aggressive. It often makes the key quantities used for the subsequent step (e.g., gradient) too noisy, and thus could hinder the convergence guarantees; see, e.g., \cite{karimireddy2019error}. A remedy is to employ the so-called {\it error feedback} (EF) mechanism introduced in \cite{karimireddy2019error}. EF can be understood as a noise reduction approach. It has also been shown to improve the convergence rate of some compressed gradient methods, e.g., \cite{alistarh2017qsgd}.

	To see how EF can be used in GCCA, let the server maintain an estimate of $\M_i^{(r+1)}=\cQ( \X_i; \btheta_i^{(r+1)} )$ for node $i$ in iteration $r$. This estimated version is denoted as $\widehat{\M}_i^{(r+1)}$ [cf. Eq.~\eqref{eq:M_update_theta}]. Similarly, every node maintains an estimated version of $\G^{(r)}$, which is denoted as $\widehat{\G}^{(r)}$. These terms will be used in the node operations and server operations.
	\subsubsection{Node Operations}	In iteration $r$, node $i$ first computes $\btheta_i$ using $\widehat{\G}^{(r)}$, i.e. 
	\begin{align}\label{eq:theta_update_compressed}
		\btheta^{(r+1)} \leftarrow \arg\min_{\bm \theta}~\frac{1}{2}\left\|{\cal Q}(\X_i;\bm \theta)-\widehat{\G}^{(r)} \right\|_{\rm F}^2.
	\end{align}
    To upload information to the server, the node computes
	the quantity $\bm \varDelta_{\bm \theta_i}^{(r)}$ defined as follows:
	\begin{align}\label{eq:deltatheta}
		& \bm \varDelta_{\bm \theta_i}^{(r)} = \cQ\left( \X_i; \btheta_i^{(r+1)} \right)  - \widehat{\M}_i^{(r)} = \\
		& \underbrace{\cQ\left( \X_i; \btheta_i^{(r+1)} \right) - \cQ\left( \X_i; \btheta_i^{(r)} \right)}_{\text{exact change}} + \underbrace{\cQ\left( \X_i; \btheta_i^{(r)} \right) - \widehat{\M}_i^{(r)}}_{\text{previous estimation error}}. \nonumber
	\end{align}
	This procedure is called EF  because the estimation error from the previous step, induced by the compression, is added to the current update.
	As mentioned, both the nodes and the server keep their own copies of $\widehat{\M}_i^{(r)}$ and $\widehat{\G}^{(r)}$, which makes implementing the EF mechanism possible.
	
	Obviously, if $\bm \varDelta_{\btheta_i}^{(r)}$ can be transmitted to the server, the server can recover $\cQ( \X_i; \btheta_i^{(r+1)} )$ without any loss. However, this is costly in terms of communication.
	To reduce the communication overhead, node $i$ quantizes $\bm \varDelta_{\btheta_i}^{(r)}$ using a compressor that is denoted by $${\bm \cC}(\cdot):\mathbb{R}^{J\times K}\rightarrow \mathbb{Q}^{J\times K},$$
	where $\mathbb{Q}\in\mathbb{R}^{J\times K}$ denotes a quantized domain that is a subset of the real-valued space $\mathbb{R}^{J\times K}$.
	Typical compressors in the literature include \texttt{SignSGD} \cite{bernstein2018signsgd}, \texttt{QSGD} \cite{alistarh2017qsgd}, and \texttt{Qsparse-local-SGD} \cite{basu2019qsparse}. Simply speaking, the compressors aim to achieve ${\bm  \cC}(\bm \varDelta_{\btheta_i}^{(r)})\approx \bm \varDelta_{\btheta_i}^{(r)},$ but using much fewer bits. Using careful design, such compressed signal, ${\bm \cC}(\bm \varDelta_{\btheta_i}^{(r)})$, may attain a more than 90\% reduction of memory compared to the uncompressed matrix, $\bm \varDelta_{\btheta_i}^{(r)}$, without slowing down the overall GCCA convergence, as will be seen later. The server also uses a similar compression strategy [cf. Eq.~\eqref{eq:deltaG}].

	After the uplink transmission in iteration $r$,
	the server receives $\bm {\bm \cC}( \bm \varDelta_{\btheta_i}^{(r)})$ and updates $\widehat{\M}_i^{(r)}$ as follows:
	\begin{equation}\label{eq:M_update_theta}
		\widehat{\M}_i^{(r+1)} \leftarrow \widehat{\M}_i^{(r)} + \bm \cC\left(\bm \varDelta_{\btheta_i}^{(r)}\right),
	\end{equation}
	and the same update is done at node $i$ for maintaining the node copy.
	
	\subsubsection{Server Operations}	The update of $\bm G$ in \eqref{eq:Gupdate} has to be changed as well, since now ${\cal Q}(\X_i,\bm \theta_i)$ is no longer available at the server end. We use the following update:
	\begin{align}\label{eq:g_subproblem_proximal}
		& \G^{(r+1)} \leftarrow 	\\
		& \arg\min_{\substack{\G^\T\G = \bm I \\\mathbf{1}^\T \G/J = \mathbf{0}} } &  \sum_{i=1}^I \frac{1}{2} \left\| \widehat{\M}_i^{(r+1)} - \G \right\|_{\rm F}^2 + \frac{1}{2\alpha_{\G}^{(r)}} \left\| \G - \G^{(r)} \right\|_{\rm F}^2 \nonumber.
	\end{align}
	Note that we deliberately add the proximal term in the above---whose importance will be clear in the convergence analysis section.
	This term also makes intuitive sense---since the server has access to the uncompressed $\G^{(r)}$, regularizing the next iterate with this ``clean'' version of $\G^{(r)}$ (as opposed to the noisy estimate $\widehat{\bm M}_i^{(r+1)}$) may safeguard the algorithm from going with an undesired direction.

	Adding the proximal term does not increase the difficulty of solving the subproblem.
	The optimal solution of \eqref{eq:g_subproblem_proximal} can still be found using the thin SVD of the following matrix:
	\begin{equation}\label{eq:Y_for_g_update}
		\Y^{(r+1)} =  \sum_{i=1}^I \left( \I - \frac{1}{J} \mathbf{1} \mathbf{1}^\T \right)\widehat{\M}_i^{(r+1)} + \frac{1}{\alpha_{\G}}\G^{(r)}  .
	\end{equation}
	% 	\end{fact}
	\begin{align}\label{eq:g_svd_update_compressed}
		\text{And  } \G^{(r+1)} = \U_{\Y^{(r+1)}} \V_{\Y^{(r+1)}}^\T,
	\end{align}
	{ where $\U_{\Y^{(r+1)}}\in \mathbb{R}^{J\times K}$ and $  \V_{\Y^{(r+1)}}\in\mathbb{R}^{K\times K}$.}
	The proof of \eqref{eq:Y_for_g_update} follows \cite[Lemma 1]{lyu2020nonlinear}; see Appendix~\ref{app:g_svd}.
	{ The computational complexity of the {thin} SVD is $\cO(JK^2)$ \cite{golub2013matrix}, where $K\ll J$ often holds---the step has a linear complexity in the number of samples and thus is not hard to be carried out by the aggregation server.}
	
	Next, the server computes $ \bDelta_{\G}^{(r)}$ as follows:
	\begin{equation}\label{eq:deltaG}
		\bm \varDelta_{\bm G}^{(r)}= \bm G^{(r+1)} -\widehat{\G}^{(r)}.
	\end{equation}
	The downlink message ${\bm \cC}(\bDelta_{\G}^{(r)})$ is then broadcast to the nodes. After downlink transmission, both the server and the nodes update $\widehat{\G}^{(r)}$ as follows:
	$$ \widehat{\G}^{(r+1)} \leftarrow \widehat{\G}^{(r)} + {\bm \cC}(\bDelta_{\G}^{(r)}).$$
	
    Note that under the above algorithmic structure, only quantized information, i.e., ${\bm \cC}(\bm \varDelta_{\bm \theta_i}^{(r)})$ and ${\bm \cC}(\bDelta_{\G}^{(r)})$, is exchanged in both uplink and downlink communications. As one will see {in Section \ref{sec:numerical_results}}, with carefully designed compressors, {this can reduce the communication overhead significantly in practice.}

	\subsection{Implementation Considerations}
	\subsubsection{Initialization}\label{subsec:estimate}
	The whole procedure requires proper initialization, which is particularly important for establishing convergence.
	To this end,
	one can randomly initialize $\btheta_i^{(0)}$ at node $i$, compute $\cQ(\X_i; \btheta_i^{(0)})$, and transmit it to the server using {\it full precision} without compression. 	
	This makes the server and the nodes able to start from the same initialization.
    The server then computes $\G^{(0)}$ using \eqref{eq:Gupdate} and broadcasts it to the nodes using {\it full precision} representation. Note that this is only full precision communication round that is needed in our algorithm.
	Since the initialization is transmitted with full precision, we set 
	\begin{align*}
		\widehat{\M}_i^{(0)} & \leftarrow \cQ\left( \X_i; \btheta_i^{(0)} \right),~ \forall i,   ~
		\widehat{\G}^{(0)}  \leftarrow \G^{(0)}.
	\end{align*} 
	
	For linear MAX-VAR GCCA, there are light approximation algorithms, e.g., those in \cite{rastogi2015multiview}, that can also be executed in a distributed manner with a single-round ${\cal O}(J\widetilde{K})$ (where $\widetilde{K}>K$) communication. {This is done by computing GCCA on heavily rank-reduced version of the views (i.e., of rank $\widetilde{K}$)}. Such initialization strategies can also be used under our framework; see \cite{fu2017scalable}. 
	This way, the proposed algorithm is warm-started, and often enjoys quicker convergence.
	\subsubsection{Choice of Compressors} \label{sec:compressor_choice}
	\begin{figure}[t]
		\centering
		\includegraphics[width=0.6\linewidth]{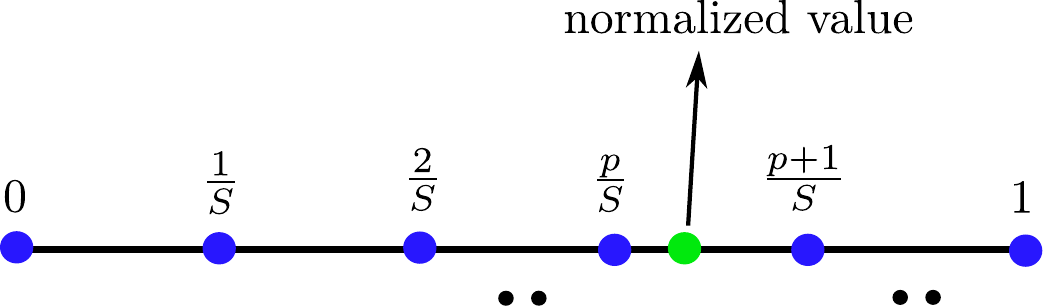}
		\caption{Illustration of the Compression operation. Normalized value corresponds to $|\bDelta(j, k)|/\|\bDelta\|_{\rm max}$.}
		\label{fig:compression}
	\end{figure}
	In principle, we hope to use the so-called $\delta$-compressors \cite{stich2018sparsified}, i.e.,
	\begin{definition}
	    [$\delta$-Approximate Compressor]\label{def:delta_compressor}
		The operator ${\bm \cC}(\cdot)$ is a $\delta$-approximate compressor for $\delta \in (0,1]$ if $\forall \bDelta \in \bbR^{L \times K}$,
		$$ \bbE_{\zeta}\left[\|{\bm \cC}(\bDelta) - \bDelta\|_{\rm F}^2\right] \leq (1-\delta)\|\bDelta\|_{\rm F}^2,$$
		where $\zeta$ is the random variable used in the compressor.
	\end{definition}
	Such compressors retain much information of $\bm \varDelta$, if $\delta$ is close to one.
	There are several options for the compressor ${\bm \cC}(\cdot)$. For example, the compressor can be quantization based \cite{alistarh2017qsgd, bernstein2018signsgd}, or sparsification based \cite{basu2019qsparse, stich2018sparsified}. In this work, we select a quantization based stochastic compression scheme introduced in \cite{alistarh2017qsgd}, because it flexibly allows multiple-precision levels, incurs low computation cost---and works well in practice. {
	Moreover, the compressor has been successfully adopted in various works \cite{reisizadeh2020fedpaq, koloskova2019decentralized} for designing communication-efficient SGD algorithms. A limitation of the compressor is that it does not support extremely heavy compression (e.g., by using only 1 bit for a real-valued scalar) as it requires at least 2 bits for encoding a real number. Nonetheless, it strikes a good balance between compression ratio and performance.}
	
	The compression scheme used in our method is illustrated in Fig. \ref{fig:compression}. Let $\bDelta \in \bbR^{J \times K}$, $\bDelta \not = \mathbf{0}$, be the unquantified data. Given the number of levels of quantization, i.e., $S\in \mathbb{N}$, we first divide the range from $0$ to $1$ into $S$ intervals. For any element $\bDelta(j, k)$, one can find a corresponding interval, $[p/S, (p+1)/S]$, such that the normalized value $|\bDelta(j, k)| / \|\bDelta\|_{\rm max} \in [p/S, (p+1)/S]$ where $p\in\{0,\ldots,S-1\}$. Let $ h(\bDelta(j, k), S)$ follow the following quantization rule:
	\begin{equation}\label{eq:compression_prob}
		h(\bDelta(j, k), S) = \begin{cases}
			p/S \quad \text{w.p.} \quad 1 - \left( \frac{|\bDelta(j, k)|}{ \|\bDelta\|_{\rm max}}S - p \right) \\
			(p+1)/S \quad \text{otherwise.}
		\end{cases} 
	\end{equation}
	The result is then multiplied with  $\|\bDelta\|_{\rm max} {\rm sgn}(\bDelta(j, k))$ to obtain $\widetilde{\bm \cC}(\bDelta)$ as follows:
	\begin{equation}\label{eq:qsgd}
		\begin{aligned}
			&[\widetilde{\bm \cC}(\bDelta)]_{j,k}	= \|\bDelta\|_{\rm max} {\rm sgn}(\bDelta(j, k)) \cdot h(\bDelta(j,k), s),
		\end{aligned}
	\end{equation}
	where the last step is for including the sign and magnitude information. In order to make this compressor a $\delta$-compressor, we apply appropriate scaling to $\widetilde{\bm \cC}(\bDelta)$, and obtain the final result as follows:
	\begin{equation}\label{eq:quantization}
		{\bm \cC}(\bDelta) = (1/u) \widetilde{\bm \cC}(\bDelta),
	\end{equation}
	where $u$ is such that $\bbE\|\widetilde{\bm \cC}(\bDelta)\|_{\rm F}^2 \leq u \| \bDelta \|_{\rm F}^2$.
	We also define ${\bm \cC}(\bDelta) = \mathbf{0}$ when $\bDelta = \mathbf{0}$. 
	
	\begin{fact}\label{fact:delta_compressor}
	The	random compressor in \eqref{eq:quantization} is a $\delta$-compressor if $u \geq  1 + (\nicefrac{JK}{4S^2})(\nicefrac{\|\bDelta\|_{\rm max}^2}{\|\bDelta\|_{\rm F}^2})$. 
	\end{fact}
	
	Fact \ref{fact:delta_compressor} follows from \cite[Lemma 3.1]{alistarh2017qsgd} and \cite[Remark 5]{karimireddy2019error}. However, for completeness, we provide the proof in Appendix \ref{app:facts}.
   Fact~\ref{fact:delta_compressor} presents a somewhat conservative bound due to the worst-case nature of the analysis. In practice, we observe that letting ${\bm \cC} (\cdot) = \widetilde{\bm \cC}(\cdot)$ often leads to the best performance. Indeed, one can show that under some more conditions, $\widetilde{\bm \cC}(\cdot)$ is also a $\delta$-compressor:
	\begin{fact}\label{fact:delta_compressor_condition}
	If $\bDelta \in \bbR^{J \times K}$ is such that
		\begin{align}\label{eq:equalengergy}
		    		\| \bDelta \|_{\rm F}^2 > (JK/(4S^2)) \|\bDelta\|_{\rm max}^2,
		\end{align}
		 $\widetilde{\bm \cC}(\cdot)$ is a $\delta$-compressor.
	\end{fact}
    The proof of Fact \ref{fact:delta_compressor_condition} is relegated to Appendix \ref{app:facts}.  The condition in \eqref{eq:equalengergy} may not be hard to meet in our case as $\bDelta$ captures the estimation error of the change of the optimization variables---whose energy is often evenly spread over the entries. 
    
	\subsubsection{Reducing Computation at the Nodes via SGD}\label{sec:sgd_computation}
	There are many off-the-shelf algorithms for solving or inexactly solving the $\bm \theta_i$-subproblem. The most natural choice may be gradient descent as in \cite{fu2017scalable}, which has shown powerful in handling large sparse data in the linear GCCA case.
	However, for very large and dense datasets or for the deep GCCA case, computing the full gradient at the nodes can be computationally costly. This is more so when nodes are edge devices with limited computing power. One natural way for circumventing this challenge is to use SGD at the nodes for updating $\bm \theta_i$'s.
	
	To be precise, in each outer iteration $r$, node $i$ runs $T$ inner iterations of SGD. Hence, in iteration $t$ of the SGD, the node randomly selects a minibatch of entities indexed by $\cF^{(r,t)}_i \in \{1, \dots, J\}$.
	The minibatch is used to compute a gradient estimation $\g_{\btheta_i}^{(r,t)}$, which is the gradient of the partial objective $$\left\| \cQ\left( \X_i\left(\cF^{(r,t)}_i, :\right) ; \btheta_i^{(r,t)}\right) - \widehat{\G}^{(r)}\left( \cF^{(r,t)}_i, : \right) \right\|_{\rm F}^2.$$ Then, the stochastic gradient update is given by:
	$$\btheta_i^{(r,t+1)} \leftarrow \btheta_i^{(r,t)} - \alpha_{\btheta}^{(r,t)} \g_{\btheta_i}^{(r,t)}.$$
	The size of $|{\cal F}^{(r,t)}_i|$ can be flexibly chosen according to the computational capacity of the nodes. In the deep GCCA case, many empirically powerful SGD variants such as 
	\texttt{AdaGrad} \cite{duchi2011adaptive,fu2019block} and 
	\texttt{Adam} \cite{kingma2015adam} can also be used for the $\bm \theta_i$-subproblem.
	
	\subsection{The \texttt{CuteMaxVar} Algorithm}
	Based on the above discussions,
	the overall algorithm is termed as {\it \uline{c}ommunication-q\uline{u}an\uline{t}iz\uline{e}d MAX-VAR} (\texttt{CuteMaxVar}) and summarized in Algorithm \ref{algo:dist_gcca}.
	One can see that the nodes and the server always exchange compressed information, which is the key for overhead reduction.
	Note that we used plain-vanilla SGD for \eqref{eq:theta_update_compressed} in Algorithm~\ref{algo:dist_gcca}. In practice, any off-the-shelf nonlinear programming algorithm can be employed to handle the $\bm \theta_i$ problem, e.g., variants of SGD such as \texttt{Adam} and \texttt{Adagrad}, full gradient descent, and higher-order algorithms like Gauss-Newton, whichever is appropriate for the problem structure.	
	
	From Algorithm~\ref{algo:dist_gcca}, one can see that the communication cost of \texttt{CuteMaxVar} can be fairly low. The only full precision transmission is in the initialization stage, which takes ${\cal O}(q_{\rm full}JK)$ bits overhead (where $q_{\rm full}=$32 if a 32-bit floating point is used for a real number) for both uplink and downlink signaling. When $r\geq 1$, only ${\cal O}(qJK+q_{\rm full})$ bits are used for ${\bm \cC}(\bm \Delta_{\bm \theta_i}^{(r)})$, ${\bm \cC}(\bm \Delta_{\bm G}^{(r)})$, and their sign and magnitude information, where $q={\cal O}(\log_2(S))$ is the number of bits that we use to encode the compressed information, which can often be $q\leq 3$. Ideally, if \texttt{CuteMaxVar} and the unquantized version use a similar number of iterations to reach the same accuracy level, then the communication saving is about $\approx(1-q/q_{\rm full} )\times 100\%$. This turns out to be the case in practice.

    {In terms of computational complexity, the SGD based update in the nodes requires $\cO(T\widetilde{J}K^2)$ flops, where $\widetilde{J}$ is the batch size in each inner iteration. As mentioned, the thin SVD operation requires $\cO(JK^2)$ flops. Moreover, the quantization takes $\cO(JK)$ flops. One can see that all operations have fairly low complexity as $K$ and $\widetilde{J}$ are often not large.}

	\begin{algorithm}[t]\label{algo:dist_gcca}
		\footnotesize
		\SetAlgoLined
		\tcp{At the nodes}
		\For{$i \leftarrow 1:I$}{
		Initialize $\btheta_i^{(0)}$;\\
		Initialize $\widehat{\M}_i^{(0)} \leftarrow \cQ(\X_i ; \btheta_i^{(0)})$; \\
		Transmit $\cQ(\X_i ; \btheta_i^{(0)})$ to the server using full precision; \\
		}
		\tcp{At the server}
		$\widehat{\M}_i^{(0)} \leftarrow \cQ(\X_i ; \btheta_i^{(0)}), \forall i$;  \quad \tcp{received from the nodes with full precision}
		Compute $\G^{(0)}$ using \eqref{eq:g_svd_update_compressed}; \\
		Initialize $\widehat{\G}^{(0)} \leftarrow \G^{(0)}$;\\
		Broadcast $\G^{(0)}$ to the nodes using full precision; \\
		
		$r \leftarrow 0$ ;\\
		\While{some stopping criteria is not met}{
			\tcp{At the Nodes}
			\eIf{r = 0}{
			    $\widehat{\G}^{(0)} \leftarrow \G^{(0)}$;  \quad \tcp{received from the server with full precision}
			    }
			    {
			    Receive ${\bm \cC} \left( \bDelta_{\G}^{(r-1)} \right)$ from the server; \\
			    $\widehat{\G}^{(r)} \leftarrow \widehat{\G}^{(r-1)} + {\bm \cC} \left( \bDelta_{\G}^{(r-1)} \right) $ ;\\
                }
			\For{$i \leftarrow 1:I$}{
				\For{$t \leftarrow 0:T-1$}{
					Sample $\cF^{(r,t)}$ and compute $\g_{\btheta_i}^{(r,t)}$; \\
					$\btheta_i^{(r, t+1)} \leftarrow \btheta_i^{(r, t)} - \alpha_{\btheta}^{(r,t)} \g_{\btheta_i}^{(r,t)} $ ;\\
				}
				$\bDelta_{\btheta_i}^{(r)} \leftarrow \cQ\left( \X_i; \btheta_i^{(r+1)} \right) - \widehat{\M}_i^{(r)}$ ;\\
				Transmit ${\bm \cC} \left( \bDelta_{\btheta_i}^{(r)} \right)$ to the server ;\\
				$\widehat{\M}_i^{(r+1)} \leftarrow \widehat{\M}_i^{(r)} + {\bm \cC} \left( \bDelta_{\btheta_i}^{(r)} \right) $ ; \quad \tcp{Node's copy of the server's estimate of $\M_i^{(r+1)}$} 
			}
			\tcp{At the Server}
			Receive ${\bm \cC} \left( \bDelta_{\btheta_i}^{(r)} \right), \forall i$ from the nodes ;\\
			$\widehat{\M}_i^{(r+1)} \leftarrow \widehat{\M}_i^{(r)} + {\bm \cC} \left( \bDelta_{\btheta_i}^{(r)} \right) $ ;\\
			Compute $\G^{(r+1)}$ using \eqref{eq:g_svd_update_compressed}; \\
			$\bDelta_{\G}^{(r)} \leftarrow \G^{(r+1)}- \widehat{\G}^{(r)}$ ;\\
			Transmit ${\bm \cC} \left( \bDelta_{\G}^{(r)} \right)$ to the nodes ;\\
			$\widehat{\G}^{(r+1)} \leftarrow \widehat{\G}^{(r)} + {\bm \cC} \left( \bDelta_{\G}^{(r)} \right) $ ; \quad \tcp{Server's copy of the node's estimate of $\G^{(r+1)}$}
			$ r \leftarrow r+1$ ;\\
		}
		{\bf Output:}{$\G^{(r)}$, $\{\btheta_i^{(r)}\}_{i=1}^I$.}
		\caption{\texttt{CuteMaxVar}}
	\end{algorithm}	
	\setlength{\textfloatsep}{10pt}

	\section{Convergence Analysis}
	In this section, we provide convergence characterizations of \texttt{CuteMaxVar}.
	Note that existing analysis for gradient-compression based { federated learning and distributed algorithms} are not applicable in our case.
	For example, \cite{karimireddy2019error} and \cite{basu2019qsparse} provided analysis for EF based distributed algorithms, but only considered a single-block smooth unconstrained non-convex optimization problem. Their algorithms also only used compression in the uplink direction. The work in \cite{ma2021communication} extended the analysis in \cite{karimireddy2019error} to cover the multi-block case. However, \cite{ma2021communication} still deals with unconstrained optimization; the compression was also only for the uplink. Note that our problem involves multiple blocks, manifold constraints, and compression of both downlink and uplink communications.
	In addition, analytical methods in \cite{bernstein2018signsgd, alistarh2017qsgd, stich2018sparsified, basu2019qsparse, karimireddy2019error} are based on distributed gradient descent with gradient compression. However, the considered {federated} MAX-VAR GCCA, even without compression, does not belong to the genre of {these methods}.
	Hence,
	nontrivial custom analysis is required to establish convergence of \texttt{CuteMaxVar}.
	
	{ To proceed, let us denote $f(\bm \theta,\bm G)=\sum_{i=1}^I f_i(\bm \theta_i,\bm G)$ in \eqref{eq:maxvar_general},} where $f_i(\bm \theta_i,\bm G)= \frac{1}{2}\| {\cal Q }(\X_i;\bm \theta_i) - \G\|_{\rm F}^2$ and $\bm \theta=[\bm \theta_1^\T,\ldots,\bm \theta_I^\T]^\T$.
	Using the above notation,
	we first define the critical point, or a Karush--Kuhn--Tucker (KKT) point of \eqref{eq:maxvar_general}:
	\begin{definition}[Critical Point]
		$(\btheta^\ast, \G^\ast)$ is a critical point of Problem \eqref{eq:maxvar_general} if it satisfies the following first order conditions:
		\begin{align}
		    \begin{cases}
				&\nabla_{\btheta_i} f_i(\btheta_i^\ast, \G^{\ast})  = \mathbf{0}, \quad \forall i\in[I], \nonumber \\
				&\sum_{i=1}^I \nabla_{\G} f_i(\btheta_i^\ast, \G^{\ast}) + \G^\ast \bLambda^\ast + \blambda^\ast \one^\T = \mathbf{0}, \nonumber \\
				&(\G^\ast)^\T \G^\ast = \I, \quad (\G^\ast)^\T \one/J = \zero \nonumber
			\end{cases}
		\end{align}
		where $\bLambda {\in \mathbb{R}^{K \times K}}$ and $\blambda {\in \mathbb{R}^{J}}$ are the Lagrange multiplier associated with the equality constraints $\G^\T \G = \I$ and $\G^\T \one/J = \zero$, respectively.  
	\end{definition}
	
	We proceed to prove convergence by assuming that each node uses SGD to update $\bm \theta_i$ as described in Algorithm~\ref{algo:dist_gcca}. The proof can be extended to cover the case where the nodes use full gradient in a straightforward manner.

	We first consider the following facts and assumptions which are common in stochastic gradient methods.
	\begin{fact}[Unbiased SGD]\label{fact:unbiased_gq}
	    Assume that the SGD samples data uniformly at random. Then, the stochastic gradient is an unbiased estimate of the true gradient:
		$$\bbE_{\xi^{(r,t)}} \left[ g_{\btheta_i}^{(r,t)} \big| \cE^{(r,t)}, \cB^{(r)} \right] = \nabla_{\btheta_i} f_i(\btheta_i^{(r,t)}, \widehat{\G}^{(r)}),$$
		where ${\cal E}^{(r,t)}$ is the filtration of the $\bm \theta_i$-updating algorithm before iteration $t$ in the outer iteration $r$, and ${\cal B}^{(r)}$ is the overall filtration before iteration $r$ \footnote{To be precise, let $\xi^{(r,t)}$ contain random variables (RVs) related to SGD at $(r,t)$ for all views. Then, $\cE^{(r,t)} = \{ \xi^{(r,0)}, \dots, \xi_i^{(r,t-1)} \}$ and $\cB^{(r)} = \{\cE^{(0)}, \zeta_{\btheta}^{(0)}, \zeta_{\G}^{(0)}, \dots, \cE^{(r-1)}, \zeta_{\btheta}^{(r-1)}, \zeta_{\G}^{(r-1)}\}$, where $\zeta_{\btheta}^{(r)}$ and $\zeta_{\G}^{(r)}$ are the RVs related to the respective random compressions.}.
	\end{fact}
	
\begin{IEEEproof}
    The proof is straightforward and thus omitted.
\end{IEEEproof}

   To proceed, we make the following assumptions:
	\begin{assumption}[Uniform Gradient Lipchitzness]\label{ass:lipschitz}
		There is a uniform Lipschitz constatnt $L \geq 0$ such that
		\begin{align*}
			\left\|\nabla_{\btheta_i} f(\btheta_i, \G) - \nabla_{\btheta_i} f(\btheta_i', \G') \right\|_{\rm F} & \leq L \|(\btheta_i, \G) - (\btheta_i', \G')\|_{\rm F} \quad \forall i.
		\end{align*}
	\end{assumption}
	\begin{assumption}[Second Moment bound]\label{ass:sgd_variance}
	 For all $t,r$, we have
		$$ \bbE_{\xi^{(r,t)}}  \left[\left\| g_{\btheta_i}^{(r,t)} \right\|_{\rm F}^2 \big| \cE^{(r,t)}, \cB^{(r)} \right] \leq \left(\sigma^{(r,t)}\right)^2 \leq \sigma^2.$$
	\end{assumption}
	
	\begin{assumption}\label{ass:delta_compressor}
		The operator ${\bm \cC}(\cdot)$ is a $\delta$-approximate compressor for $\delta \in (0,1]$.
	\end{assumption}
	Note that Assumption~\ref{ass:lipschitz} is easy to be satisfied in our case, if $\bm \theta_i^{(r)}$'s are bounded. The boundedness of $\bm \theta_i$ can be readily established by the boundedness of $\G^{(r)}$ and the fact that our algorithm produces a non-increasing objective sequence, under mild conditions; see similar arguments in \cite{fu2016robust}.
	Assumption~\ref{ass:sgd_variance} naturally holds if $\bm \theta_i$'s are bounded.

	Next, we establish a lemma that bounds the compression error terms:
	\begin{subequations}\label{eq:compression_errors}
		\begin{align}
			\Z_{\btheta_i}^{(r)}  &= {\bm \cC}\left( \bDelta_{\btheta_i}^{(r)} \right) - \bDelta_{\btheta_i}^{(r)}  \label{eq:compression_error_theta} \\
			\Z_{\G}^{(r)}  &= {\bm \cC}\left( \bDelta_{\G}^{(r)} \right) - \bDelta_{\G}^{(r)}. \label{eq:compression_error_g} 
		\end{align}
	\end{subequations}
	
	\begin{lemma}[Compression Error Bound]\label{lemma:compression_error}
		Assume that a $\delta$-compressor is employed. Then, we have
		\begin{align*}
		\bbE \left\| \Z_{\btheta_i}^{(r)} \right\|_{\rm F}^2 & \leq \frac{4(1-\delta)}{\delta^2} \sum_{t=0}^{T-1} \left( \alpha_{\btheta}^{(r,t)} \right)^2  \sigma^2 \quad \text{and} \\
		\bbE \| \Z_{\G}^{(r)} \|_{\rm F}^2 & \leq \frac{4(1-\delta)}{\delta^2} \bbE \left\| \G^{(r+1)} - \G^{(r)}\right\|_{\rm F}^2, 
		\end{align*}
		where the expectation is taken with respect to $\cB^{(r+1)}$.
	\end{lemma}
	
	\subsection{Convergence to Critical Points}
First, we show that \texttt{CuteMaxVar} converges to a critical point of Problem~\eqref{eq:maxvar_general} for both the linear and deep GCCA cases. Consider the following matrix:
	\begin{equation}\label{eq:subdifferential}
		\bm \varPhi(\btheta^\ast, \G^\ast) = \begin{bmatrix}
			\nabla_{\btheta_1} f_1(\btheta_1^\ast, \G^{\ast}) \\
			\vdots \\
			\nabla_{\btheta_I} f_I(\btheta_I^\ast, \G^{\ast}) \\
			\nabla_{\G} f(\btheta^\ast, \G^{\ast}) + \G^\ast \bLambda^\ast + \blambda^\ast \one^\T
		\end{bmatrix}.	
	\end{equation}
	Clearly, ${\bm \varPhi}(\btheta^\ast, \G^\ast) = \mathbf{0}$ implies that $(\btheta^\ast, \G^\ast)$ is a critical point of Problem \eqref{eq:maxvar_general}. With this, we show the following theorem:

    \begin{theorem}(Asymptotic Convergence)\label{thm:critical_point}
		Under Assumptions~\ref{ass:lipschitz}-\ref{ass:delta_compressor},
		assume that the step sizes $\alpha_{\btheta}^{(r,t)} \leq 1/L$  for all $r,t$ and $(\alpha_{\G}^{(r)})^2 v(r,t) > 0, \forall r,t$, where $v(r,t) = ( \nicefrac{1}{2\alpha_{\G}^{(r)}} - {I} - \sum_{t=0}^{T-1} \nicefrac{2 \alpha_{\btheta}^{(r+1,t)} I L^2 (1-\delta)}{\delta^2} )$. Further assume that $  \alpha_{\btheta}^{(r,t)}=\alpha_{\btheta}^{(r)}$ for all $t\in[T]$ (i.e., the $\bm \theta_i$-updates use constant step size), and that $\{ \alpha_{\btheta}^{(r)}  \}$ and $\{(\alpha_{\G}^{(r)})^2 v(r,t)\}$ satisfy the Robinson-Monroe rule. Then, on average, every limit point of the solution sequence is a KKT point of \eqref{eq:maxvar_general}; i.e.,
		$$\liminf_{{r \to \infty} } \bbE \left\| \bm \varPhi\left(\btheta^{(r)}, \G^{(r)}\right) \right\|_{\rm F}^2 = 0, $$
		where the expectation is taken over all the randomness in the solution sequence.
	\end{theorem}
	    The proof of Theorem \ref{thm:critical_point} is relegated to Appendix \ref{app:proofthm_critical}.
	The step sizes condition in Theorem \ref{thm:critical_point} is not hard to satisfy.
	A sequence $\{s^{(r)}\}$ satisfies the Robinson-Monroe rule if 
	\begin{align}\label{eq:robins_monroe}
	\sum_{r=0}^\infty s^{(r)} = \infty,\quad \sum_{r=0}^\infty \left( s^{(r)} \right)^2 < \infty.  
	\end{align}
	We have the following fact:
   	\begin{fact}\label{fact:step_size}
     For any $\alpha^{(r)} < 1$ satisfying \eqref{eq:robins_monroe}, setting $\alpha_{\btheta}^{(r,t)} = \alpha_{\G}^{(r)} $ $= \alpha^{(r)} (\nicefrac{I}{c} (\sqrt{1 + \nicefrac{c}{I^2}} - 1)){\rm min}\{ 1/L, 1 \}$,  where $c = \nicefrac{4TIL^2 (1-\delta)}{\delta^2}$, makes $\{(\bm \alpha_{\bm G}^{(r)})^2v(r,t)\}$ and $\{  \bm \alpha_{\bm \theta}^{(r)} \}$ satisfy the condition in \eqref{eq:robins_monroe} as well. 
	\end{fact}
	The proof of Fact~\ref{fact:step_size} is provided in Appendix~\ref{app:facts}.

    Theorem~\ref{thm:critical_point} works under a fairly general {step size selection rule}, but the convergence is asymptotic. {Next, we show that a stronger convergence result can be obtained under a more stringent step size specification.}
	{ To this end}, we define a potential function $\Gamma^{(r)}$ as follows:
	\begin{align}
		& \Gamma^{(r)} = \sum_{t=0}^{T-1} \sum_{i=0}^I \left\| \nabla_{\btheta_i} f_i(\btheta_i^{(r,t)}, \G^{(r)})\right\|_{\rm F}^2  +  \bigg\|I \G^{(r)}  \\
		& - \sum_{i=1}^I \cQ\left( \X_i; \btheta_i^{(r+1)} \right) + \G^{(r+1)} \bLambda^{(r+1)} + \blambda^{(r+1)} \one^\T \bigg\|_{\rm F}^2. \nonumber
	\end{align}
	It can be seen that the value of the potential function can be used to measure convergence:
	\begin{fact}\label{fact:potential_func}
	    When $\Gamma^{(r)} \to 0$, $(\btheta^{(r)}, \G^{(r)})$ converges to a KKT point of Problem \eqref{eq:maxvar_general}, i.e., $\| \bm \varPhi\left(\btheta^{(r)}, \G^{(r)}\right) \|_{\rm F}^2 \to 0$.
	\end{fact}
	The proof of Fact~\ref{fact:potential_func} is relegated to Appendix~\ref{app:facts}. The following theorem shows the convergence rate of the algorithm using this relation:
	\begin{theorem}(Convergence Rate)\label{thm:convergence_rate}
	    Under Assumptions \ref{ass:lipschitz}-\ref{ass:delta_compressor}, 
		let $\alpha_{\btheta} = \alpha_{\G}  =  \nicefrac{1}{\sqrt{R+1}} (\nicefrac{I}{c} (\sqrt{1 + \nicefrac{c}{I^2}} - 1)){\rm min}\{ 1/L, 1 \}$, where $c = \nicefrac{4TIL^2 (1-\delta)}{\delta^2}$ . Then, the following holds:
		\begin{align*}
			& \frac{1}{R+1}\sum_{r=0}^R \bbE[\Gamma^{(r)}] \leq \\
			& \cO\left(\frac{1}{\sqrt{R+1}}\right) \left(f\left( \btheta^{(0)}, \G^{(0)} \right) - \bbE \left[ f\left( \btheta^{(R+1)}, \G^{(R+1)} \right) \right] \right) + \\
			& \cO\left(\frac{1}{\sqrt{R+1}}\right)\left(\frac{IT \sigma^2 (\delta^2 L + 8(1-\delta))}{2\delta^2}  +  \frac{2^{I+1}IT(1-\delta)\sigma^2}{\delta^2} \right).	
		\end{align*}
	\end{theorem}
	The proof of Theorem \ref{thm:convergence_rate} can be found in Appendix \ref{app:proofthm_rate}. It shows that the potential function decreases at a {\it sublinear} rate. 
	
	\subsection{Global Optimality and Geometric Rate of The Linear Case}\label{sec:global_conv}
	In linear case, the objective in \eqref{eq:maxvar_general} is equivalent to \eqref{eq:maxvar_gcca}, with $\cQ(\X_i; \btheta_i) = \X_i \Q_i$. 
	It is known that the uncompressed version of \texttt{CuteMaxVar} (i.e., the \texttt{AltMaxVar} algorithm in \cite{fu2017scalable}) is a global optimal algorithm of \eqref{eq:maxvar_gcca}. 
In particular, the
	\texttt{AltMaxVar} algorithm in \cite{fu2017scalable} is shown to attain the global optimum of \eqref{eq:maxvar_gcca} at a geometric rate under the assumption that the $\Q_i$-update is sufficiently accurate. With aggressive uplink and downlink compression introduced in \texttt{CuteMaxVar}, does such optimality still hold?
	In this section, we show that \texttt{CuteMaxVar} converges to a {\it neighborhood} of the global optimum at the same geometric rate as in \texttt{AltMaxVar}.

	Note that the linear MAX-VAR GCCA problem amounts to computing the subspace spanned by the $K$ principal eigenvectors of $\bm P = \sum_{i=1}^I \X_i (\X_i^\T \X_i)^{-1} \X_i^\T \in \bbR^{J \times J}$. 
	Hence,
	to measure the progress of the iterates, we use the subspace distance metric defined in \cite{golub2013matrix}. Let $$\U_1 = \U_{\P}(:,1:K),~~\U_2 = \U_{\P}(:,K+1:J).$$  We wish the algorithm-output $\G^{(r)}$ to be the basis of ${\cal R}(\U_1)$. The subspace distance between between $\cR(\U_1)$ and $\cR(\G^{(r)})$ is given by $${\rm dist} (\cR(\G^{(r)}), \cR(\U_1)) = \| \U_2^\T \G\|_2 .$$ The following theorem characterizes the convergence property of \texttt{CuteMaxVar} in the linear MAX-VAR GCCA case:
	\begin{theorem}\label{thm:linear_convergence}(Global Optimality of The Linear Case)
		Let the eigenvalues of $\bm P = \sum_{i=1}^I \X_i (\X_i^\T \X_i)^{-1} \X_i^\T \in \bbR^{J\times J}$ be $\lambda_1, \dots, \lambda_{J}$ in descending order. Assume $\lambda_{K} > \lambda_{K+1}$, and $\cR(\G^{(0)})$ is not orthogonal to any components in $\cR(\U_1)$, i.e., $\cos(\gamma)  = \sigma_{\rm min}(\U^\T \G^{(0)}) > 0$.

		Further, assume that one solves the $\Q_i$-subproblem in each iteration to an accuracy such that 
		$\bbE_{ \cE^{(r,T)} } [ \|\Q_i^{(r+1)} - \widetilde{\Q}_i^{(r+1)} \|_2 |\cB^{(r)} ] \leq \kappa,$
		where $\widetilde{\Q}_i^{(r+1)} = (\X_i^\T \X_i)^{-1} \X_i\G^{(r)}$. Then, with probability at least $1-\omega$, 
		$$ {\rm dist}\left(\cR(\G^{(r)}), \cR(\U_1)\right) \leq {\rm tan}(\gamma) \left( \frac{\lambda_{K+1}}{\lambda_{K}} \right)^r + C, $$
		where $C = \cO(\nicefrac{\lambda_{K}}{\lambda_{K} - \lambda_{K+1}})$ is a constant and 
		$\omega = \sum_{i=1}^I r \sigma_{\rm max}(\X_i) \kappa + \frac{2Ir\sqrt{1-\delta}}{\delta} \sqrt{ \sum_{t=0}^T (\alpha_{\btheta}^{(r,t)})^2 \sigma^2 + 2 K}.$
	\end{theorem}
	
	The proof of Theorem \ref{thm:linear_convergence} is relegated to Appendix \ref{app:proofthm_linear}. 
	{ In the proof, the algorithm under the linear GCCA case is regarded as a noisy version of the {\it orthogonal iterations} (OI) \cite{golub2013matrix}---a multi-component variation of the {\it power iterations}. The quantization and inexact solution to the least squares problem introduce noise to the OI process. Nonetheless, if the noise is bounded and reduced along the iterations, the noisy OI algorithm still converges to the principal components of $\sum_{i=1}^I \X_i \X_i^\dagger$, which are our optimal solutions.}
	
	Theorem \ref{thm:linear_convergence} shows that if the $\Q_i$-subproblem is solved to a good accuracy (i.e., if $T$ is large enough), and the compressor incurs small compression error (i.e., if $\delta$ is close to 1), then \texttt{CuteMaxVar} converges to a neighborhood of the global optimal solution at a geometric rate with high probability. 
    
    \color{black}
	
	\section{Numerical Results}\label{sec:numerical_results}
	In this section, we showcase the effectiveness of \texttt{CuteMaxVar} using synthetic and real data experiments for both the linear and the deep cases. {Source code of the experiments is made available on \url{https://github.com/XiaoFuLab/federated_max_var_gcca.git}}
	
	\subsection{Experiment Settings of Linear GCCA} 
	\subsubsection{Synthetic Data Generation} Following the setting in \cite{fu2017scalable}, we generate each view of the synthetic data using $\X_i = \Z \A_i + \nu_i \bm N_i,$ where $\Z \in \bbR^{J \times D}$ is the latent factor with $J\geq D$, $\A_i \in \bbR^{D \times M_i}$ is a ``mixing matrix'', $\nu_i^2 \in \bbR$ is the noise variance, and $\bm N_i$ is the noise. Here, $\Z, \A_i$ and $\bm N_i$ for all $i$ are sampled from the i.i.d. standard normal distribution. 
	
	\subsubsection{Baselines} We use a couple of algorithms that are designed for large-scale MAX-VAR GCCA, namely, the \texttt{AltMaxVar} algorithm \cite{fu2017scalable} and the \texttt{MVLSA} algorithm \cite{rastogi2015multiview}, respectively. Our algorithm in the linear GCCA case can be understood as an exchanging information-quantized version of \texttt{AltMaxVar}. Hence, observing the communication efficiency using \texttt{AltMaxVar} as a benchmark is particularly important. The \texttt{MVLSA} algorithm is not an iterative algorithm. We present its result here for as a reference of the GCCA optimization performance.

	\subsubsection{Evaluation Metric} 
	As in conventional GCCA works, we also observe the GCCA performance by observing the cost value attained for Problem~\eqref{eq:maxvar_general}. Other than that, to evaluate the communication efficiency,
	we define compression ratio (\texttt{CR}) of \texttt{CuteMaxVar} as follows. Let $R_{C}$ and $R_A$ be the numbers of iterations required by \texttt{CuteMaxVar} and \texttt{AltMaxVar} to attain a certain convergence criterion, respectively. Then, we define \texttt{CR} as: 
	$$ \texttt{CR} = 1 - \frac{2q JK R_{C}}{2\times q_{\rm full} JK R_A}=1 - \frac{qR_{C}}{q_{\rm full} R_A},$$
	where $q$ is the number of bits per scalar used by the compressor, and $q_{\rm full}$ represents the full precision---which is 32 in our case. The terms $2q JK$ and $2\times 32JK$ represent the numbers of bits exchanged between the server and a node in each iteration in the quantized and full precision cases, respectively.
	For example, \texttt{CR}$=0.9$ means that the algorithm with compression attains the same convergence criterion (e.g., cost value) but only uses 10\% of the communication cost compared to the uncompressed version.

	To further observe the communication efficiency of the proposed method, we define ``bits per (exchanged) variable'' as the cumulative sum of bits used to communicate a variable up to a certain iteration. One can express the bits per variable (\texttt{BPV}) measure of \texttt{CuteMaxVar} as
    $$ \texttt{BPV}(r) = q_{\rm full} + rq,$$
    where $r$ is the number of iterations, and $q_{\rm full}$ appears because initialization requires a round of full precision information exchange.
    Similarly, the \texttt{BPV} of \texttt{AltMaxVar} is $ \texttt{BPV}(r) = q_{\rm full} + rq_{\rm full}$.
    We will show the cost value against $\texttt{BPV}(r)$ in this section to demonstrate communication saving.

	\subsubsection{Results}
    
	\begin{figure}
		\centering
		\includegraphics[width=0.7\linewidth]{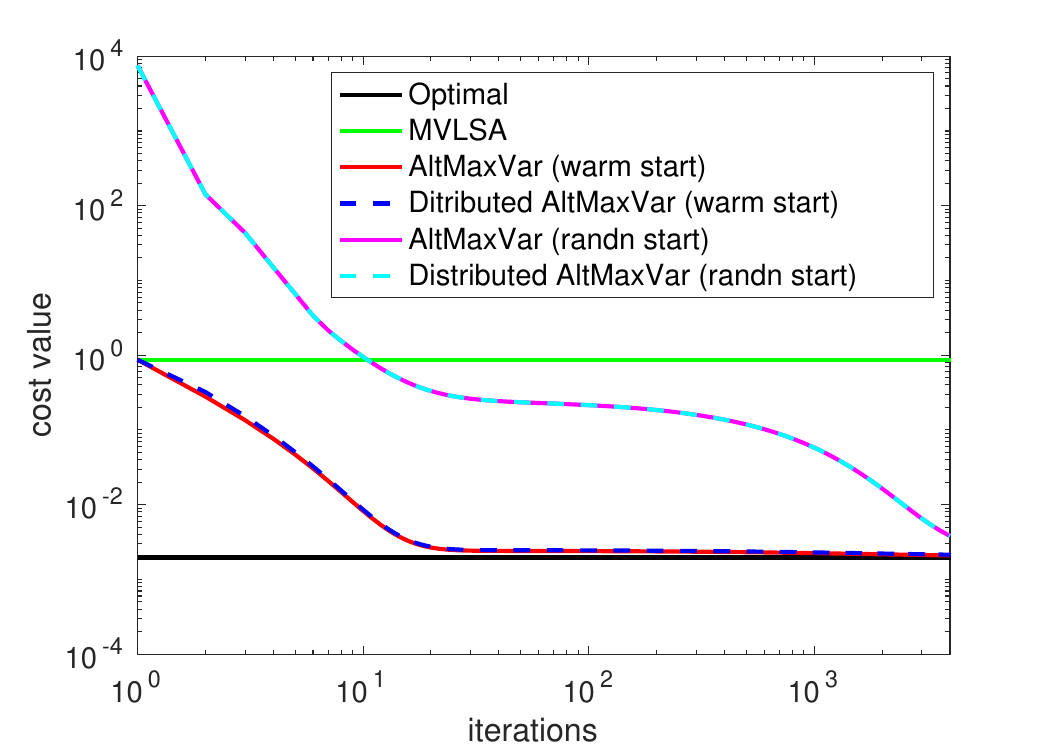}
		\caption{Convergence behavior of the proposed method and the baselines; $(J,N,D,K,I) = (500, 25, 20, 5, 3)$.}
		\label{fig:linear_synthetic_small_conv}
	\end{figure}

    Fig.~\ref{fig:linear_synthetic_small_conv} shows the results of a case where $(J,N,D,K,I) = (500, 25, 20, 5, 3)$ with $N_i = N$ for all $i$, $\nu_i = 0.01$, and $\widetilde{K}=10$ for \texttt{MVLSA}.
    For \texttt{CuteMaxVar}, we use SGD for the $\Q_i$-update with a batch size of 150. We use full gradient for \texttt{AltMaxVar} and set the step size to $1/\lambda_{\max}(\X_i^\T\X_i)$, following the setting in \cite{fu2017scalable}.  We use $q=3$ which corresponds to using 3 bits per scalar. We run 50  Monte Carlo trials and average the results. From the figure, one can observe that the convergence curves of \texttt{AltMaxVar} and \texttt{CuteMaxVar} when initialized randomly (denoted by ``randn start'') and by the solution of \texttt{MVLSA} (denoted by ``warm start''), respectively. One can see that both \texttt{AltMaxVar} and \texttt{CuteMaxVar} converge to the optimal solution. More importantly, the two algorithms converge with essentially the same rate. This suggests that our heavy compression of the exchanged information (3 bits per variable v.s. 32 bits per variable) is almost lossless in both speed and accuracy.

	\begin{figure}
		\centering
		\includegraphics[width=\linewidth]{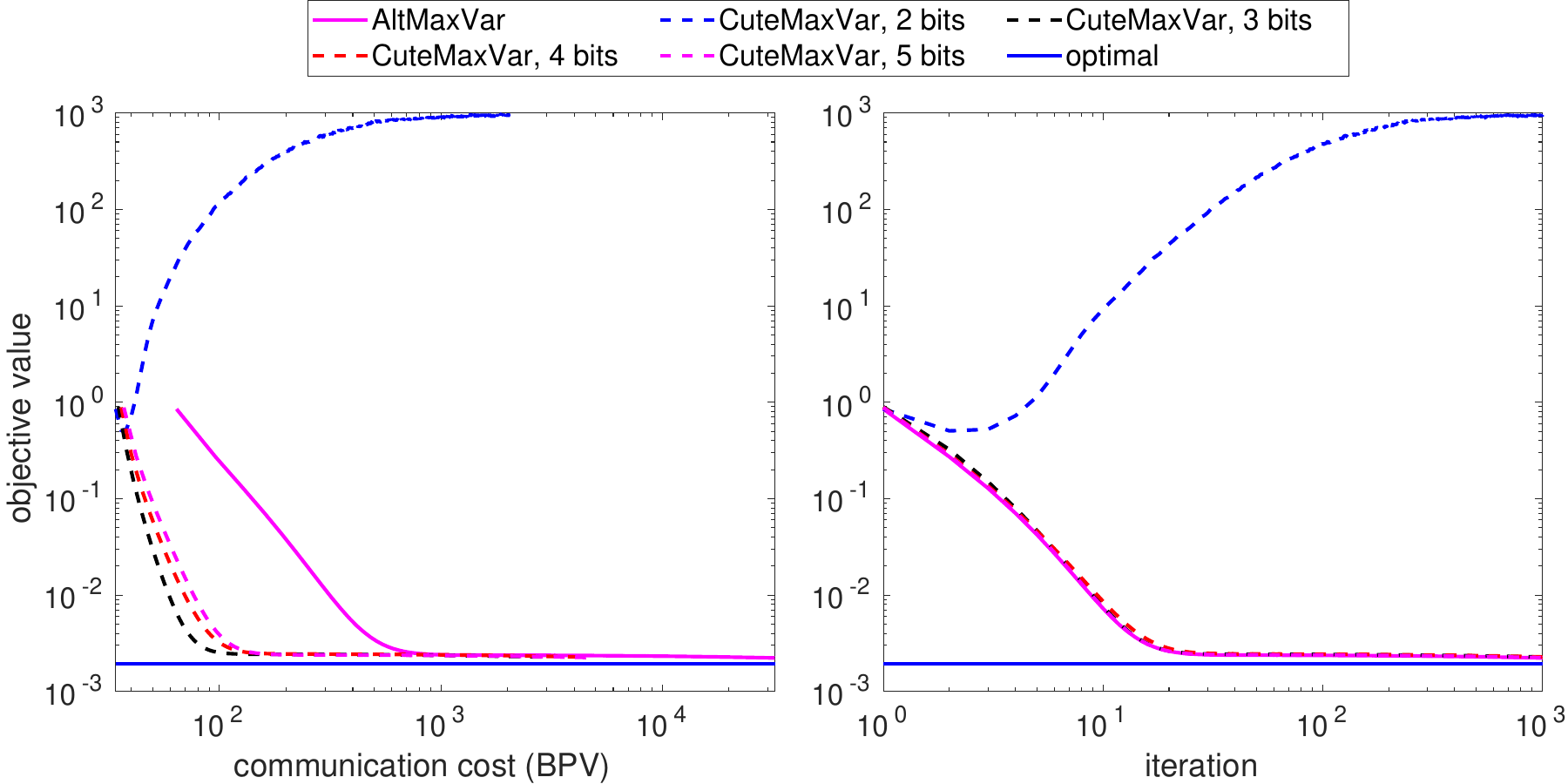}
		\caption{Communication cost [left] and convergence rate [right] under various compression levels; $(J,N,D,K,I) = (500, 25, 20, 5, 3)$. }
		\label{fig:linear_synthetic_varying_bits}
	\end{figure}

	Fig. \ref{fig:linear_synthetic_varying_bits} shows the communication cost (in terms of \texttt{BPV}$(r)$) and the convergence rate (in iterations) under various levels of compression. The settings follow those in Fig.~\ref{fig:linear_synthetic_small_conv}. One can see that with $q=3$, we get the least communication cost without any loss in the convergence rate. Note that $q=2$ corresponds to $S=1$. Under such circumstances, the compressor is less likely to satisfy the condition in Fact~\ref{fact:delta_compressor_condition} and thus convergence may be a challenge. 
 
	\begin{table}[t]
	    \centering
	    \caption{Compression ratios obtained by \texttt{CuteMaxVar} for the experiment in Fig.~\ref{fig:linear_synthetic_varying_bits}.}
	    \begin{tabular}{|c|c|c|c|c|}
	    \hline
	                    & 2 bits & 3 bits & 4 bits & 5 bits  \\
	                    \hline
	        \texttt{CR} & - & 0.9062 & 0.8681 & 0.8438 \\ \hline
	    \end{tabular}
	    \label{tab:cr_linear_small}
	\end{table}
	\begin{figure}[t]
		\centering
		\includegraphics[width=0.65\linewidth]{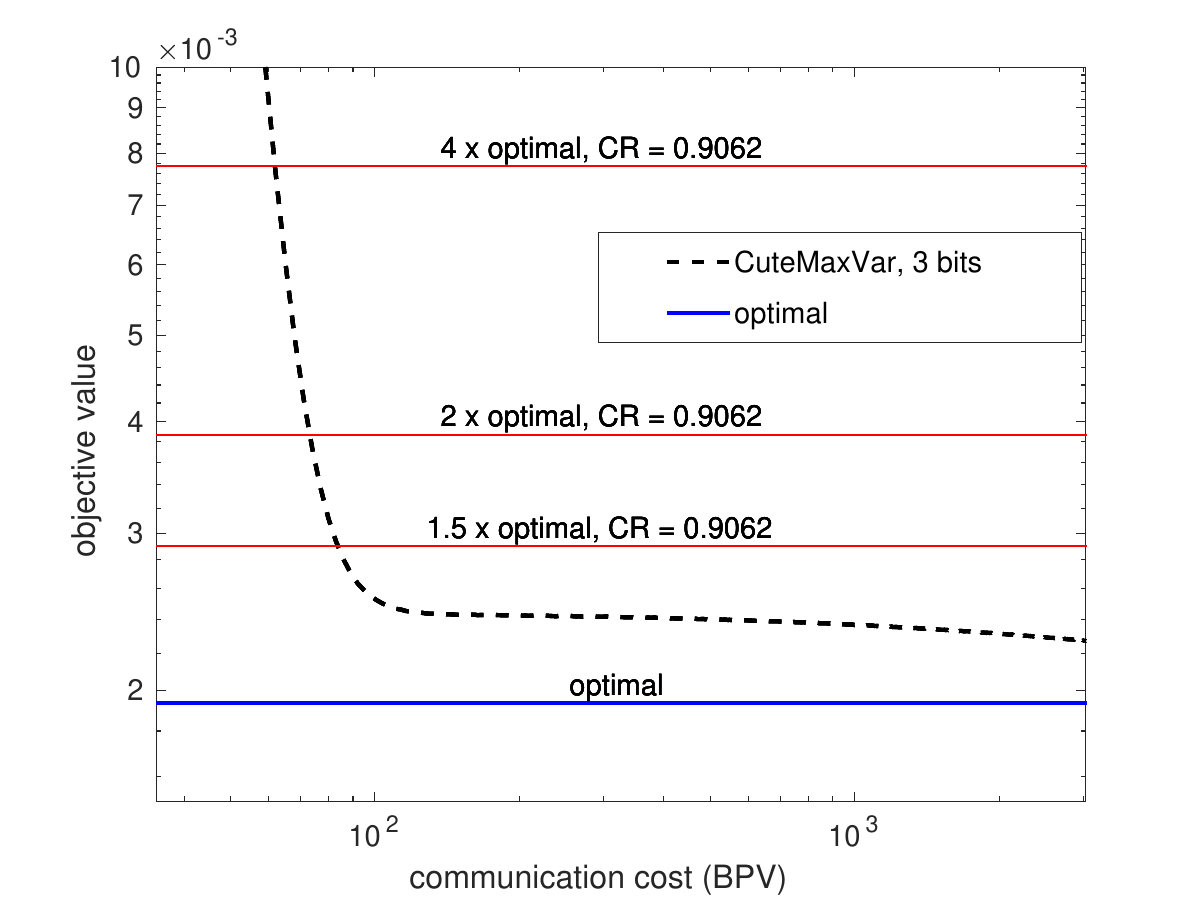}
		\caption{Compression ratios (\texttt{CR}s) attained by \texttt{CuteMaxVar} under various objective value levels close to the optimal value; $q=$3 bits.}
		\label{fig:cr_levels}
	\end{figure}

	Table~\ref{tab:cr_linear_small} shows the compression ratios for the experiment in Fig.~\ref{fig:linear_synthetic_varying_bits}. The \texttt{CR} is computed by considering the number of iterations required by the algorithms to reach $1.5\times v^\star$, where $v^\star$ is the optimal value of \eqref{eq:maxvar_general} (also see Fig.~\ref{fig:cr_levels}). In this case, using $q=$3 obtains the best \texttt{CR}, which is 0.9062.
	To get a closer look, we also zoom in to plot the convergence curves close to the optimal value in
	Fig.~\ref{fig:cr_levels}. 
	The figure
	shows the \texttt{CR}s for various levels of objective values under $q=3$. It shows that \texttt{CuteMaxVar} achieves \texttt{CR}$=0.9062$ consistently for all levels under consideration. This is because \texttt{CuteMaxVar} enjoys almost identical iteration complexity of \texttt{AltMaxVar} in this simulation.

    \begin{figure}[t!]
        \centering
        \includegraphics[width=\linewidth]{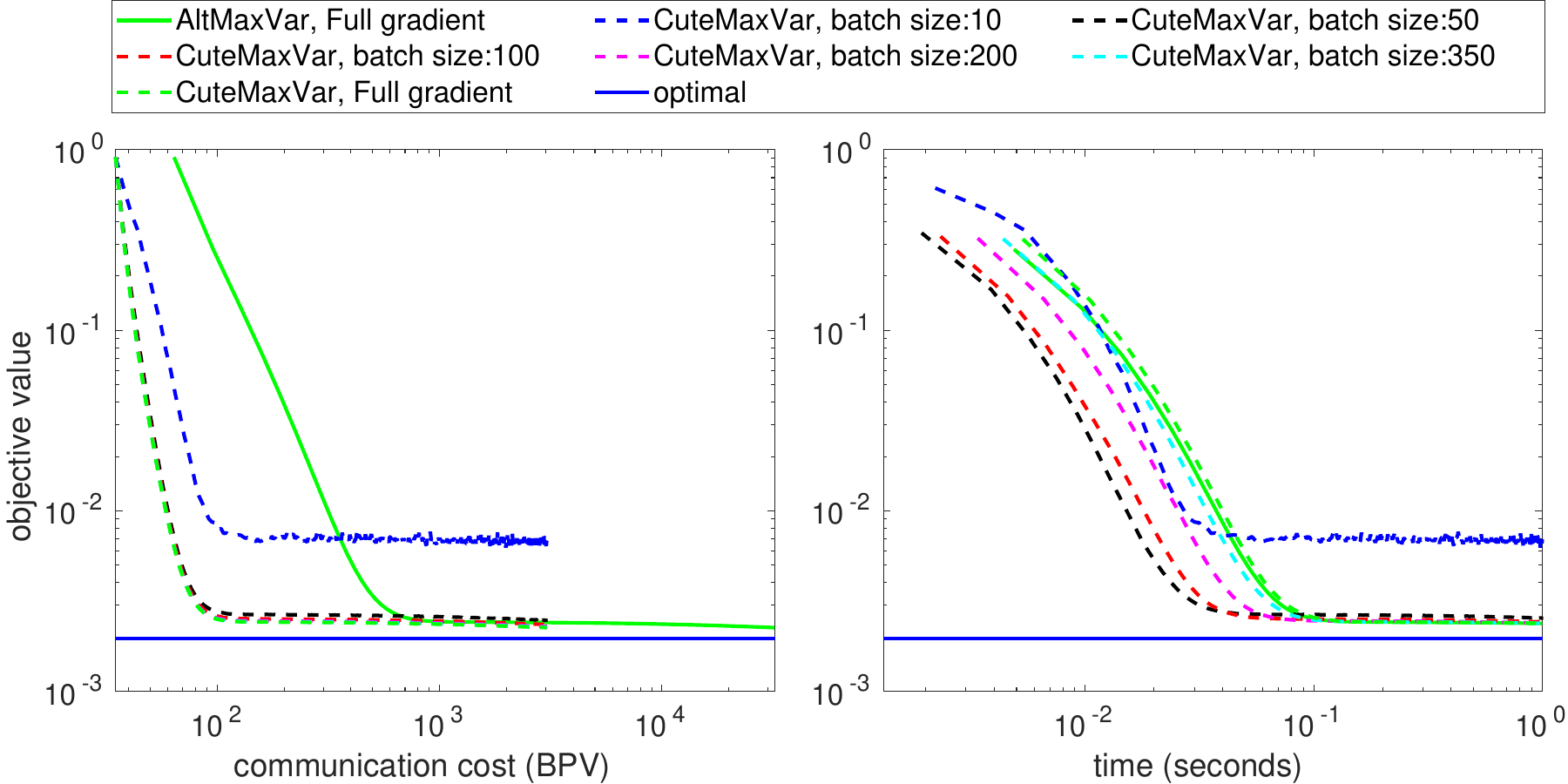}
        \caption{Communication cost [left] and convergence speed in time [right] under various batch sizes; $(J,N,D,K,I) = (500, 25, 20, 5, 3)$. }
        \label{fig:linear_synthetic_sgd}
    \end{figure}

    Fig. \ref{fig:linear_synthetic_sgd} shows the communication cost (in $\texttt{BPV}(r)$) and the convergence speed (in seconds) under various batch sizes used for the SGD-based $\Q$-subproblem solving. The setting for this experiment is the same as in Fig.~\ref{fig:linear_synthetic_small_conv}, including the same constant step size. One can see from Fig.~\ref{fig:linear_synthetic_sgd} [right] that with smaller batch sizes, \texttt{CuteMaxVar} converges faster in terms of time due to the reduced computation load at the nodes. However with very small batch sizes, e.g., 10, \texttt{CuteMaxVar} does not converge well. 
    This is expected, since using a larger batch size makes the variance of the gradient estimation (proportional to $\sigma^2$) smaller, which improves convergence, per Theorem~\ref{thm:linear_convergence}.

    \begin{figure}[t!]
    
        \centering
        \includegraphics[width=\linewidth]{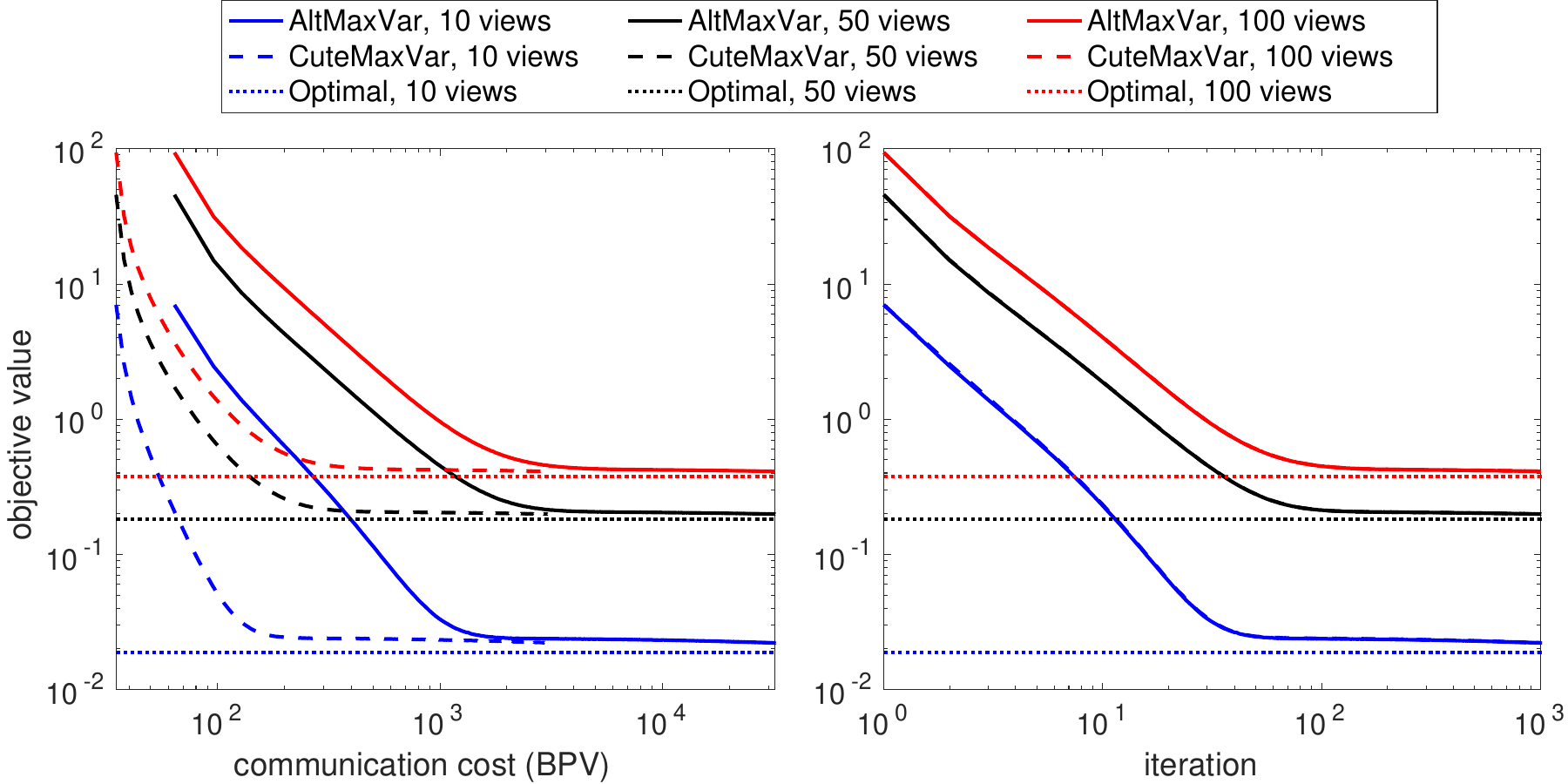}
        \caption{Communication cost [left] and convergence speed in time [right] under various number of views; $(J,N,D,K,q) = (500, 25, 20, 5, 3)$.}
        \label{fig:linear_synthetic_small_views}
    \end{figure}
    
    Fig.~\ref{fig:linear_synthetic_small_views} shows the communication cost (in $\texttt{BPV}(r)$) and the convergence speed (in seconds) under various numbers of views. One can see that the performance of the proposed method does not deteriorate when $I$ grows from 10 to 100. Even when $I=100$, the \texttt{CuteMaxVar} almost always has the same objective value as that of \texttt{AltMaxVar} in {\it every} iteration.

    \begin{figure}[t]
		\centering
		\includegraphics[width=\linewidth]{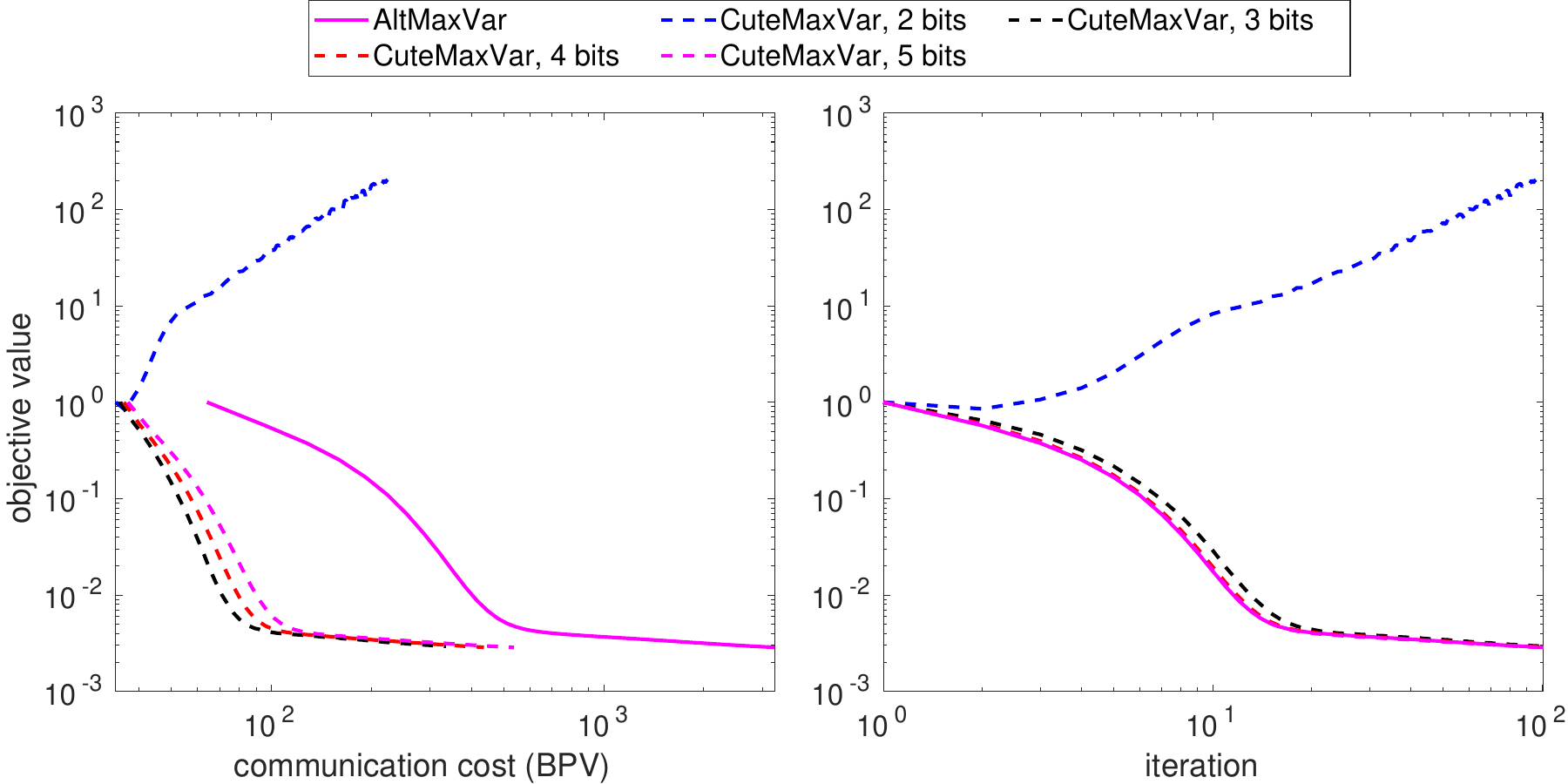}
		\caption{Communication cost [left] and convergence rate [right] under various compression levels; $(J,N,D,K,I) =(50000, 2000, 200, 5, 3)$.}
		\label{fig:linear_synthetic_large_bits}
	\end{figure}

	Figs.~\ref{fig:linear_synthetic_large_bits} uses a large-scale synthetic dataset to evaluate the proposed method. In both figures, $\Z$, $\A_i$, and $\bm N_i$, are sparse random Gaussian matrices whose nonzero elements have zero mean and unit variance. The density of $\X_i$ is defined as $\rho_i = \frac{{\rm nnz}(\X_i)}{JN}$, where ${\rm nnz}(\cdot)$ counts the number of non-zero elements of its argument. Due to the large size of this simulation, we present the results averaged from 5 Monte Carlo trials (as opposed to 50 in the previous smaller cases).
	In this simulation, we use $\rho =\rho_i= 0.02$ for all $i$.
	We set $(J,N_i,D,K,I,\nu_i) =$ $(50000, 2000, 200, 5, 3,0.01)$. We use $\widetilde{K}=50$ for \texttt{MSLVA}. 
	Fig.~\ref{fig:linear_synthetic_large_bits} shows the communication cost and the convergence rate (in iterations) under different compression levels. We observe that using $q=3$ costs the least communication overhead.
	
	\subsubsection{Real Data Experiment - Multilingual Data}
	We consider learning low-dimensional representations of sentences from different languages in the Europarl Corpus \cite{koehn2005europarl}. The sentences are from the translations of the same documents.
	Among the sentences that are aligned across all translations, we use 160,000 sentences for training and 10,000 as the test data. 
 {We use the data from ten languages as the $I=10$ views\footnote{The 10 languages are selected randomly from the 21 languages in the Europarl Corpus. Details can be found in our source code; see the link in the beginning of Sec.~\ref{sec:numerical_results}. }; i.e., $\X_i(j,:)$ corresponds to the $j$th sentence in the $i$th language. The raw features of the sentences are constructed following the same approach in \cite{kanatsoulis2018structured}. The dimension of each sentence in all languages is $N_i=$524,288. 
 }

{
For evaluation, we consider a cross-language sentence alignment task, where $\Q_i$'s are learned using GCCA over aligned views.
At the testing stage, we apply $\Q_i$'s to unaligned sentences to assist cross-language alignment.
To be specific, for each view $i \in [I]$, we compute $\X_i^{\rm (test)}(p,:)\Q_i, \forall p$. We consider $\Q_i$ and $\Q_m$ to be ``good'' representation learners if they correctly reflect the fact that $\X_i^{\rm (test)}(p,:)$ and $\X_m^{\rm (test)}(p,:)$ correspond to the same entity. 
To be precise, let
$${\rm dist}_{i,m}(p,q) = \|\X_i^{(\rm test)}(p,:)\Q_i - \X_m^{(\rm test)}(q,:)\Q_m \|_2$$
denote the distance between representations of the $p$th sentence in view $i$ and $q$th sentence in view $m$. Then we define \textit{alignment accuracy} \texttt{(AC)} as
\begin{align*}
  &\texttt{AC} = \nicefrac{1}{I(I -1)} \times \\
  &\sum_{(i,m) \in [I] \times [I], i \neq m} \nicefrac{\sum_{p \in [J]} \mathbb{I} \left[ {\rm dist}_{i,m}(p,p) \leq {\rm dist}_{i,m}(p,q), \forall q \in [J] \backslash p\right]}{J} , 
\end{align*}
where $\mathbb{I}[E]$ is the 0-1 indicator function of the event $E$.
Note that $\texttt{AC}$ is between 0 and 1, and a higher value of \texttt{AC} means a better performance.
}

 \begin{figure}[t]
 
		\centering
		\includegraphics[width=\linewidth]{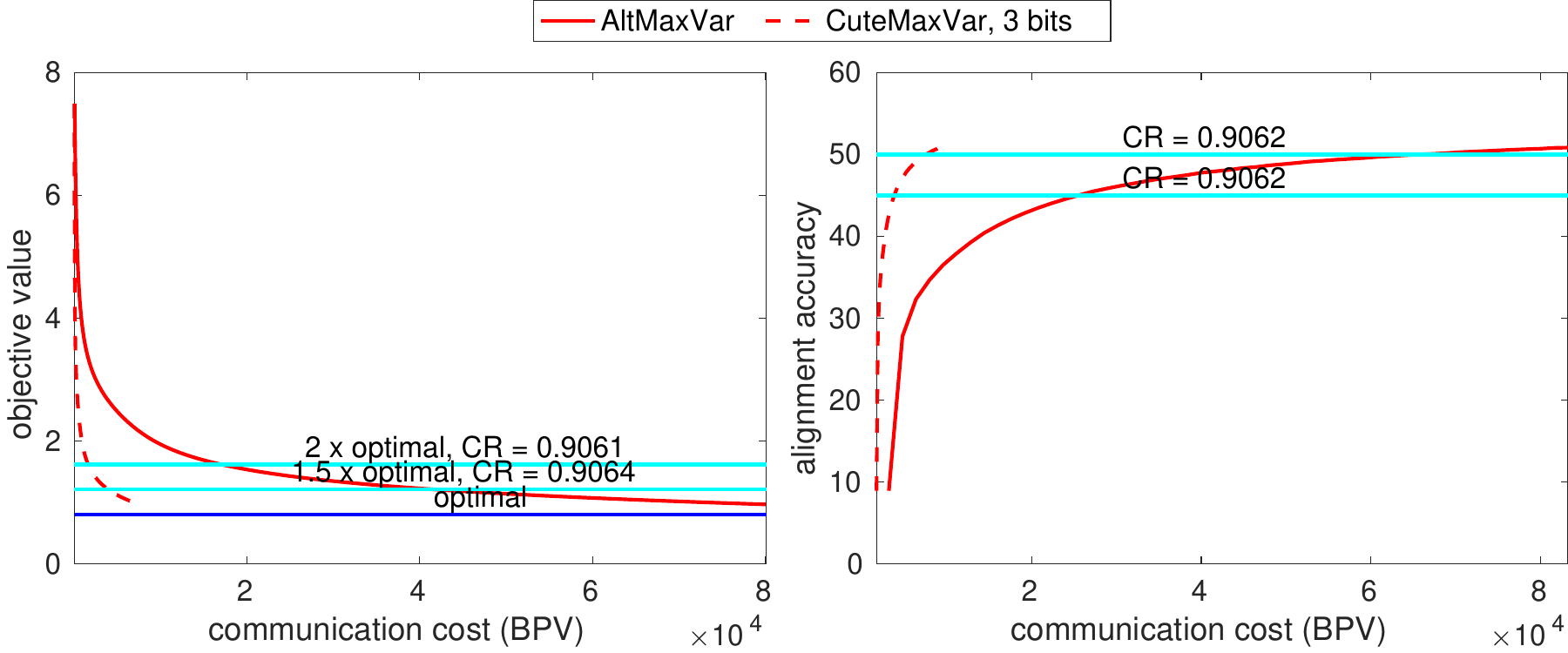}
		\caption{Training objective [left] and \texttt{AC} on the test set [right] vs. {\texttt{BPV}} for \texttt{CuteMaxvar} {and} \texttt{AltMaxVar}; $I=10$ languages are used.}
		\label{fig:linear_real}
	\end{figure}
    
    Fig.~\ref{fig:linear_real} shows the communication cost of \texttt{CuteMaxVar} and \texttt{AltMaxVar} for the aforementioned task. We use $q=3$, $T=20$, $K=10$ and $\widetilde{K} = 100$ for our method, and the same setting for the step size as in the synthetic case. We run both methods with SGD using a batch size of 2000 for the $\Q_i$-subproblem. Fig.~\ref{fig:linear_real} shows a  significant communication cost reduction by \texttt{CuteMaxVar} compared to \texttt{AltMaxVar}.
    Using the cost value reaching $1.5 \times {v}^\star$ as the stopping criterion, \texttt{CuteMaxVar} achieves a \texttt{CR} of 0.9064.  
    {If one uses the \texttt{AC} attaining 50\% as a stopping criterion, \texttt{CuteMaxVar} achieves a \texttt{CR} of 0.9062. }

    \subsection{Experiment Settings of Deep GCCA}
    \subsubsection{Baseline} In this section, we use the full precision version of \texttt{CuteMaxVar} for deep GCCA as our baseline. We denote \texttt{CuteMaxVar} by \texttt{CuteMaxVar-Deep} to emphasize the use of neural networks. Similarly, we use \texttt{AltMaxVar-Deep} to denote its uncompressed version. Note that this baseline algorithm deals with a similar objective as in \cite{benton2019deep}, but the algorithm there was not designed for {federated learning}.
    
	\subsubsection{Synthetic-Data Experiment Settings} The $\cQ(\cdot; \btheta_i)$ function used in our synthetic data experiment is a neural network with 2 hidden layers. The network has 128 and 64 fully connected neurons for the first and the second layers, respectively. We run SGD for the $\btheta_i$-subproblem with a batch size of 1000 using the \texttt{Adam} optimizer \cite{kingma2015adam}. For \texttt{Adam}, we set the initial step size to be $0.001$. We run the algorithms with $T=10$ inner iterations and $R=50$ outer iterations. We use $q=3$ bits for compression. Both \texttt{CuteMaxVar-Deep} and \texttt{AltMaxVar-Deep} use the same hyperparameter settings and the same neural architecture. The difference lies only in that the latter does not quantize the exchanged information.

	\subsubsection{Results}
\begin{figure}
		\centering
		\includegraphics[width=0.8\linewidth]{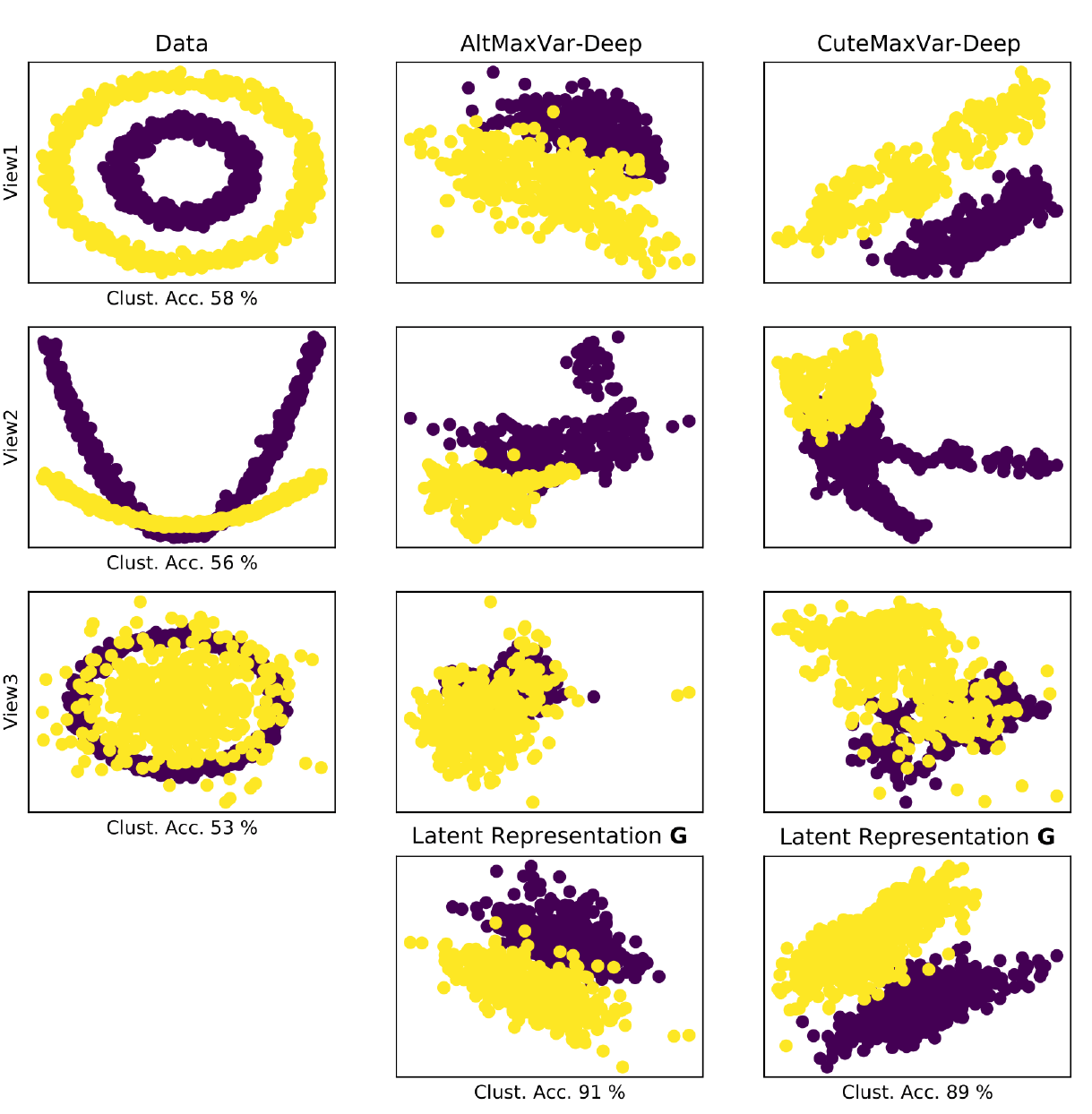}
		\caption{First, second, and third row: scatter plot of individual views [left], transformed views using \texttt{AltMaxVar}[middle], transformed views using \texttt{CuteMaxVar} [right]. Bottom row: scatter plot with clustering accuracy of the learned latent representation $\bm G$.}
		\label{fig:toy_dgcca_scatter}
	\end{figure}

	Fig. \ref{fig:toy_dgcca_scatter} shows an illustrative example. We generate synthetic data following \cite{benton2019deep}. As shown in the left column of Fig. \ref{fig:toy_dgcca_scatter}, there are $I=3$ views of $J=10,000$ samples with two-dimensional features, i.e., $(I, J, N_i) = (3, 10000, 2)$ for all $i=1,2,3$. 
	For each view, the points with the same color belong to the same cluster. However, simply using existing clustering algorithms, e.g., spectral clustering  \cite{ng2002spectral}, does not reveal the nodes' cluster membership well (see the first column in Fig.~\ref{fig:toy_dgcca_scatter}). In such problems, using deep GCCA, one can learn representations of the data points that form linearly separable clusters \cite{benton2019deep}, which can improve performance in downstream tasks.    
    From Fig.~\ref{fig:toy_dgcca_scatter}, with $K=2$, one can see that both methods can learn a $\bm G$ that clearly represent the cluster membership of the data. Further, \texttt{AltMaxVar-Deep} and \texttt{CuteMaxVar-Deep} achieve a clustering accuracy of 89\% and 91\%, respectively, using the same spectral clustering algorithm. That is, there is virtually no loss incurred by quantization.
 
\begin{figure}
	    \centering
	    \includegraphics[width=0.95\linewidth]{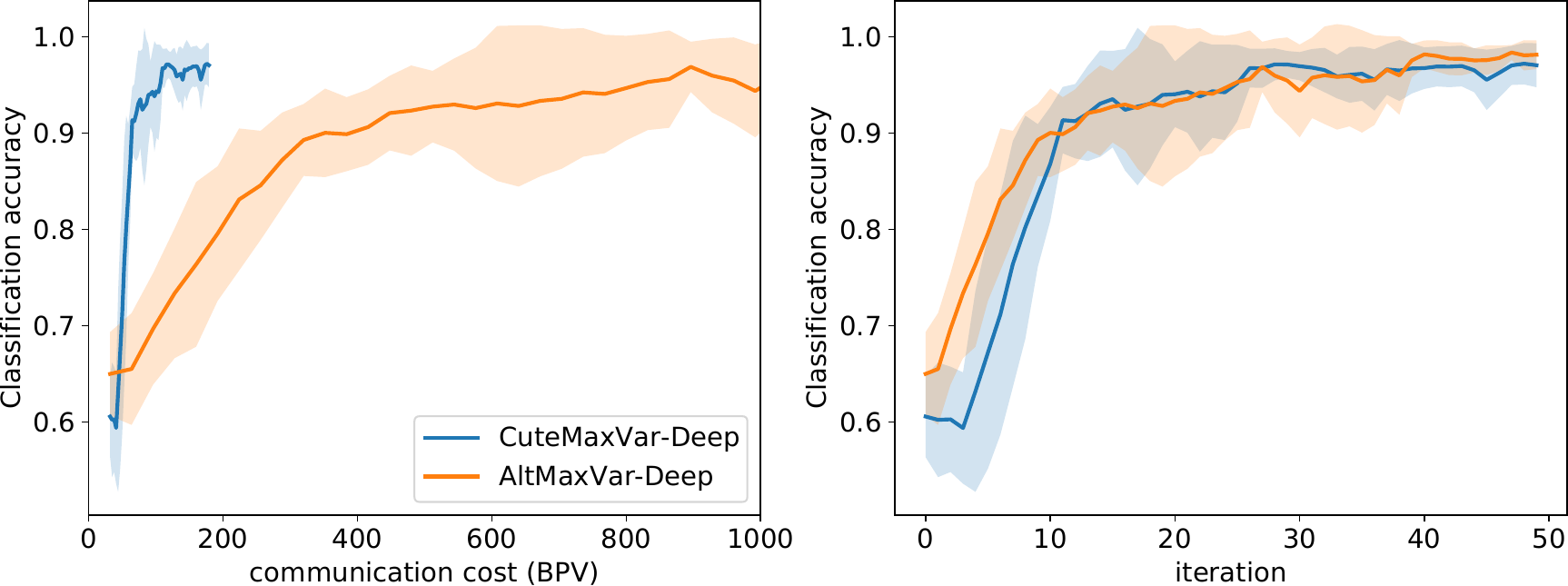}
	    \caption{Classification accuracy vs. {\texttt{BPV}} [left] and iteration [right].}
	    \label{fig:toy_class_acc}
	\end{figure}

    Fig.~\ref{fig:toy_class_acc} shows a plot of classification accuracy over unseen test data against the communication cost and the number of iterations, respectively, averaged over 10 trials. To measure the classification accuracy, we employ a linear classifier on the test set that has a size of 1000. To be specific, we use the support vector machines (SVM) trained over $\G^{(r)}$ that was learned from the training data with a size of $10000$.  Fig.~\ref{fig:toy_class_acc} [Right] shows that the classification accuracy of \texttt{CuteMaxVar-Deep} improves as fast as that of \texttt{AltMaxVar-Deep} when the number of iterations grows---which is consistent with our previous observations.
    Fig.~\ref{fig:toy_class_acc} [Left] demonstrates the communication reduction achieved by \texttt{CuteMaxVar-Deep} compared to the baseline. To attain $95\%$ classification accuracy, \texttt{CuteMaxVar-Deep} achieves a compression ratio of 0.906 ---i.e., less than 10\% of the communication overhead is used relative to \texttt{AltMaxVar-Deep}.
    
	\subsubsection{Real Data Experiment - EHR Data} 
	{ In multiview medical data analytics, data exchanging may be undesired due to the sensitive nature. In such a case, federated learning is well-motivated \cite{ma2021communication}.
	Here, we consider learning representations of different diagnoses from multiple views of an \textit{electronic health record} (EHR) dataset.} 
	
	We use EHR dataset from Centers for Medicare and Medicaid Services (CMS) \cite{cmsdataset}. It is a publicly available EHR dataset with patients' information protected. It contains claims data synthesized using a random sample of Medicare beneficiaries from 2008 to 2010. There are entries of more than 6,000,000 synthetic beneficiaries distributed across 20 files. Each file can act as an independent EHR dataset consisting of the following records: beneficiary summary, inpatient claims (hospitalized patient), outpatient claims, carrier claims, and prescription drug events. We utilize 3 files out of 20 to work as the three views of the data.  
    	
	We learn representations of the diagnoses from the 3 views as follows. Each diagnosis group can contain hundreds of diseases/diagnoses. For each diagnosis $j$ and view $i$, we construct its feature vector $\X_i(j,:)$ by using co-occurrence count with medications, i.e., $\X_i(j,n)$ is the co-occurrence of diagnosis $j$ and medication $n$ in view $i$. The CMS data utilizes the ICD-9 
	\footnote{Available: \url{https://www.cdc.gov/nchs/icd/icd9.htm}}
	coding system for diagnoses and HCPCS 
	\footnote{Available: {https://www.cms.gov/Medicare/Coding/HCPCSReleaseCodeSets}}
	for medication, which can represent around 13,000 diagnoses and more than 6000 medicines, respectively. However, most of the diagnoses do not occur at all in a given view. Therefore, we only use the most frequent 1045 diagnoses and $N_i=$511 medicines. Further, we discard diagnosis groups which have less than 40 diagnoses in the resulting diagnosis list. Therefore, we have 6 diagnosis groups, and a total of $J=$727 diagnoses. Finally, we shuffle all diagnoses inside each diagnosis group and hold out $200$ entities at random for testing the learned representations. Therefore, for training, we have $J=527$, $N_i=N=511$, whereas for prediction, we have $J=$200.
    We use a 4-hidden layer fully connected neural network with 512, 256, 128, and 64 neurons in the first to last hidden layers, respectively. We set the output layer size to be 10. We use the ReLU activation function and batch normalization after each layer except for the output layer. We run SGD for the $\btheta_i$-subproblem with a batch size of $250$ for 10 inner iterations. The step size $\alpha^{(r)}_{\bm \theta}$ is scheduled following the \texttt{Adam} rule.
    The \texttt{CuteMaxVar-Deep} and \texttt{AltMaxVar-Deep} algorithms share the same set of hyperparameters. For \texttt{CuteMaxVar-Deep}, we use $q = 4$.  
    
	\begin{figure}
		\centering
		\includegraphics[width=\linewidth]{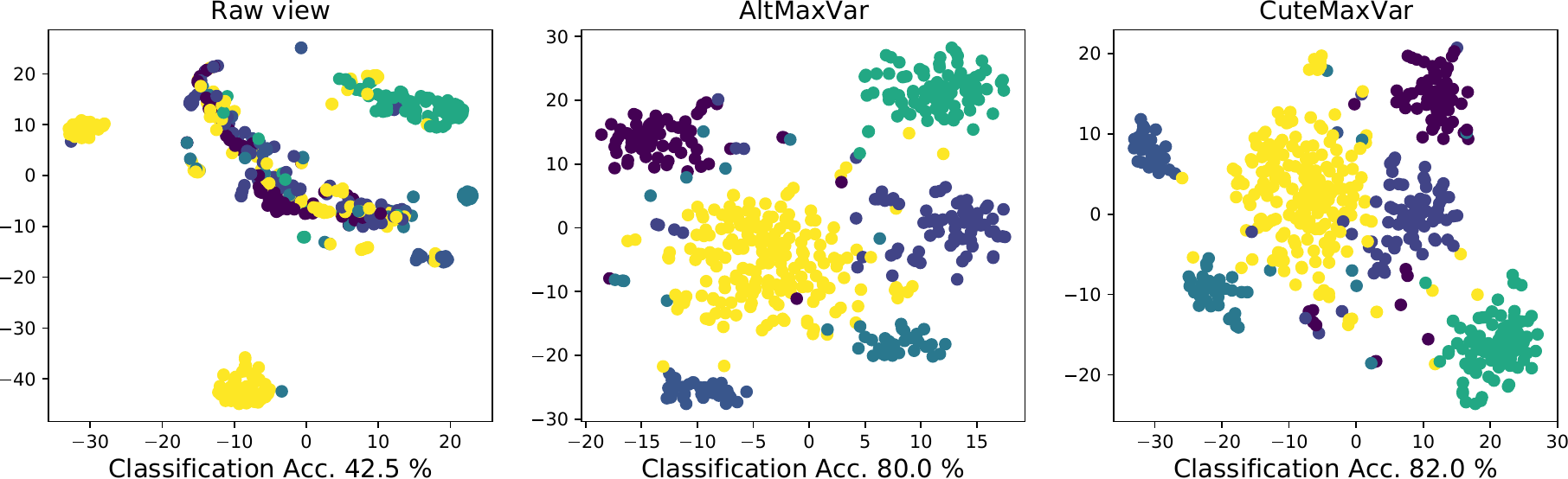}
		\caption{t-SNE plots with classification accuracy for the diagnosis groups classification task using raw data [left], \texttt{AltMaxVar} [middle], and \texttt{CuteMaxVar} [right].}
		\label{fig:tsne_ehr}
	\end{figure}
	
	Fig.~\ref{fig:tsne_ehr} shows the t-SNE plot for the representations learned by \texttt{CuteMaxVar-Deep} and \texttt{AltMaxVar-Deep} (i.e., $\bm G(j,:)$'s), and a raw view (i.e., $\X_1(j,:)$'s) for comparison. We use SVM with the radial basis functions as kernels to classify the learned representations. The classification accuracy using the raw data view is $42.5 \%$. However, \texttt{AltMaxVar-Deep} attains an $80.0 \%$ accuracy and \texttt{CuteMaxVar-Deep} $82.0\%$. First, this shows obvious benefits of representation learning using deep GCCA. Second, \texttt{CuteMaxVar-Deep} maintains good performance after heavily compressing the exchanging information, as we observed in the linear case.
	
	\begin{figure}
	    \centering
	    \includegraphics[width=0.95\linewidth]{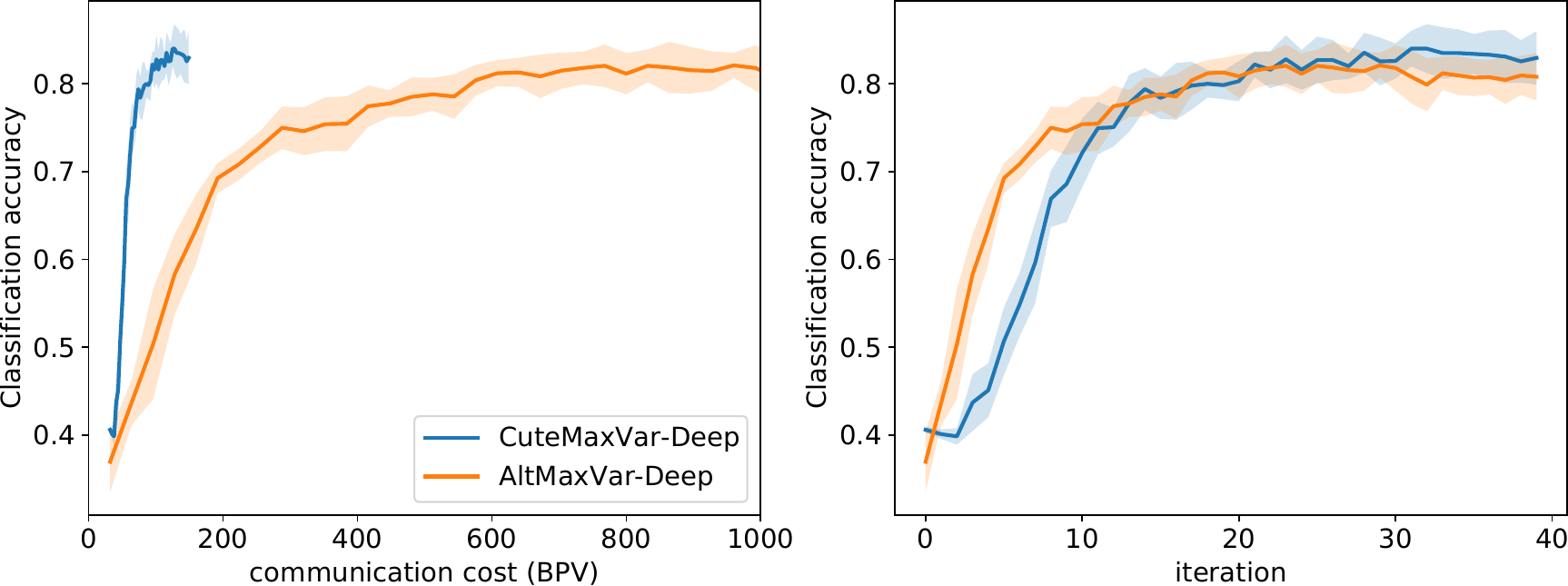}
	    \caption{Classification accuracy vs. {\texttt{BPV}} [left] and iteration [right]. }
	    \label{fig:class_acc_ehr}
	\end{figure}
	
    Fig.~\ref{fig:class_acc_ehr} shows the classification accuracy achieved by the algorithms on the test set against the communication cost and number of iterations, which are averaged over 10 trials due to the randomness of SGD. One can see that although \texttt{AltMaxVar-Deep} improves its classification accuracy using fewer iterations in the beginning, both methods reach the best accuracy using around the same number of iterations (Fig.~\ref{fig:class_acc_ehr} [right]). However, \texttt{CuteMaxVar-Deep} costs significantly less communication overhead to reach the same accuracy level. In particular, to attain an $80\%$ classification accuracy, \texttt{CuteMaxVar-Deep} works with $\texttt{CR}=$0.867.
	
	\section{Conclusion}
	In this work, we proposed a communication-efficient {federated learning} framework for linear and deep MAX-VAR GCCA. Our algorithm is designed for the scenario where the views are stored at computing agents and data sharing is not allowed.
	First, we integrated the idea of exchanging information quantization and error feedback to come up with a communication-economical algorithmic framework {for federated GCCA}, which was shown to save about 90\% communication overhead in both the linear and deep GCCA cases {in our empirical study}. Second, we offered rigorous convergence analysis to support our design. {As generic federated optimization results are not applicable to the GCCA problem}, we provided custom analyses to accommodate the special problem structure of linear/deep MAX-VAR GCCA. In addition to critical point convergence of the general framework, we also established approximate global optimality of the linear case under our {federated learning} scheme. We tested the algorithm on multiple synthetic and real datasets. The results corroborate our analyses and show promising performance.
	
\appendices
\section{Proof of Theorem \ref{thm:linear_convergence}}\label{app:proofthm_linear}
	Our proof idea is reminiscent of those in \cite{lu2014large,fu2017scalable}. To be specific, we treat the alternating optimization process as a noisy orthogonal iteration algorithm for subspace estimation \cite{golub2013matrix}. By bounding the noise, the desired solution can be shown.
	
	Instead of using $\bm G^{(r)}$, our algorithm uses $\widehat{\bm G}^{(r)}$ to update $\bm Q$. Hence, the optimal solution of the $\Q$-subproblem in iteration $r$ and node $i$, denoted by $\widetilde{\Q}_i^{(r+1)}$ is given by
	
	\begin{align*}
		\widetilde{\Q}_i^{(r+1)} & = (\X_i^\T \X_i)^{-1} \X^\T \widehat{\G}^{(r)} \\
		& = (\X_i^\T \X_i)^{-1} \X^\T \G^{(r)} + (\X_i^\T \X_i)^{-1} \X^\T(\widehat{\G}^{(r)}- \G^{(r)}) \\
		& = (\X_i^\T \X_i)^{-1} \X^\T \G^{(r)} + (\X_i^\T \X_i)^{-1}\X^\T \Z_{\G}^{(r)} .
	\end{align*}
	Instead of solving the $\Q$-subproblem to optimality, our algorithm uses SGD to obtain an inexact solution, i.e., $\Q_i^{(r+1)}$, which can be expressed as
	$$ \Q_i^{(r+1)} =(\X_i^\T \X_i)^{-1} \X^\T \G^{(r)} + (\X_i^\T \X_i)^{-1} \X^\T \Z_{\G}^{(r)} + \W_i^{(r)}, $$
	where $\W_i^{(r)}$ is an error term due to the solution inexactness.

	Quantization error is also introduced by quantizing and transmitting the update in $\X_i \Q_i^{(r+1)}$. At the server, the actually received signal from node $i$ is expressed as follows:
	$$ \widehat{\M}_i^{(r+1)} = \X_i\Q_i^{(r+1)} + \Z_{\Q_i}^{(r)}. $$
	Note that the solution of the $\G$ subproblem using SVD can be viewed as a change of bases. Hence, there is an invertible $\bm \Theta$ such that, the iterations can be written as 
	\begin{align*}
		& \G^{(r+1)} {\bm \Theta}^{(r+1)} \\
		& = \sum_{i=1}^I \widehat{\M}_i^{(r+1)} = \bm P \G^{(r)} + \underbrace{\sum_{i=1}^I \left( \bm P_i \Z_{\G}^{(r)} + \X_i \W_i^{(r)} + \Z_{\Q_i}^{(r)}  \right) }_{\E^{(r)}},	
	\end{align*}
	where $\P_i = \X_i (\X_i^\T \X_i)^{-1} \X_i^\T$ and $\P = \sum_{i=1}^I \P_i$.
	
    We can bound $\| \E^{(r)} \|_2$ as follows:
	\begin{align}\label{eq:error_bound}
		& \bbE \left\| \E^{(r)} \right\|_2 = \bbE\left\| \sum_{i=1}^{I} \left( \bm P_i \Z_{\G}^{(r)} + \X_i \W_i^{(r)} + \Z_{\Q_i}^{(r)}  \right)\right\|_2 \nonumber \\
		& \leq \left(\sum_{i=1}^I \bbE\left\|\Z_{\G}^{(r)} \right\|_2 + \sigma_{\rm max}(\X_i) \bbE\left\|\W_i^{(r)} \right\|_2 + \bbE\left\|\Z_{\Q_i}^{(r)} \right\|_2 \right) \nonumber \\
		& \stackrel{(a)}{\leq}  \sum_{i=1}^I \sigma_{\rm max}(\X_i) \kappa + \frac{2I\sqrt{1-\delta}}{\delta} \sqrt{ \sum_{t=0}^T \left(\alpha_{\btheta}^{(r,t)} \right)^2 \sigma^2 + 2 K} \nonumber \\
	\end{align}		
 
	where (a) follows by using Lemma~\ref{lemma:compression_error} and from the observation that $\G^{(r)}$ and $\G^{(r+1)}$ satisfy the manifold constraint $\G^\T \G = \I$.

	Now, using Markov inequality, one can see that $\| \E^{(r)} \|_2 < 1$ with probability at least $1-\tilde{\omega}$, where $\tilde{\omega} = \sum_{i=1}^I \sigma_{\rm max}(\X_i) \kappa + \nicefrac{2I\sqrt{1-\delta}}{\delta} \sqrt{ \sum_{t=0}^T (\alpha_{\btheta}^{(r,t)} )^2 \sigma^2 + 2 K}$.
	
	Recall that $\U_1 = \U_{\P}(:,1:K)$ and $\U_2 = \U_{\P}(:,K+1:J)$. Multiplying both sides by $\U_1$ and $\U_2$, we get
	\begin{align}\label{eq:subspaces}
		& \begin{bmatrix} \U_1^\T \G^{(r+1)} \\ \U_2^\T \G^{(r+1)} \end{bmatrix} {\bm \Theta}^{(r+1)}  = \begin{bmatrix} {\bm \Lambda}_1 \U_1^\T \G^{(r)} + \U_1^\T \E^{(r)} \\
			{\bm \Lambda}_2 \U_2^\T \G^{(r)} + \U_2^\T \E^{(r)} \end{bmatrix}.
	\end{align}
	The above equation implies
	\begin{align}\label{eq:subspace_ineq_chain}
		& \left\| \left( \U_2^\T \G^{(r+1)} \right) \left( \U_1^\T \G^{(r+1)} \right)^{-1} \right\|_2 \nonumber \\
		& = \left\| \left({\bm \Lambda}_2 \U_2^\T \G^{(r)} + \U_2^\T \E^{(r)} \right) \left({\bm \Lambda}_1 \U_1^\T \G^{(r)} + \U_1^\T \E^{(r)} \right)^{-1} \right\|_2 \nonumber\\
	\end{align}
	With probability $1-\tilde{\omega}$,
	\begin{align}
	    & \left\| \left( \U_2^\T \G^{(r+1)} \right) \left( \U_1^\T \G^{(r+1)} \right)^{-1} \right\|_2 \nonumber \\
		& \stackrel{(a)}{\leq} \left\| \left({\bm \Lambda}_2 \U_2^\T \G^{(r)} \right)\left({\bm \Lambda}_1 \U_1^\T \G^{(r)}\right)^{-1} \right\|_2 \nonumber \\
		& + C_1 \left( \|\U_1^\T \E^{(r)}\|_2 + \|\U_2^\T \E^{(r)}\|_2 \right)\nonumber \\
		& \leq \left(\frac{\lambda_{K+1}}{\lambda_{K}}\right)\left\| \left({\bm \Lambda}_2 \U_2^\T \G^{(r)} \right)\left({\bm \Lambda}_1 \U_1^\T \G^{(r)}\right)^{-1} \right\|_2 + 2 C_1,
	\end{align}
	where, for (a), we have used Taylor expansion of $({\bm \Lambda}_1 \U_1^\T \G^{(r)} + \U_1^\T \E^{(r)})^{-1} = ({\bm \Lambda}_1 \U_1^\T \G^{(r)})^{-1} - ({\bm \Lambda}_1 \U_1^\T \G^{(r)})^{-1} (\U_1^\T \E^{(r)}) ({\bm \Lambda}_1 \U_1^\T \G^{(r)})^{-1}+ (({\bm \Lambda}_1 \U_1^\T \G^{(r)})^{-1} \U_1^\T \E^{(r)})^2 ({\bm \Lambda}_1 \U_1^\T \G^{(r)})^{-1}- \dots $ and absorbed the higher order terms of $\|\E^{(r)}\|_2$ by $C_1$ because of $\| \E^{(r)}\|_2 < 1$. 

	Unrolling \eqref{eq:subspace_ineq_chain} for $r$ iterations, the following holds with probability at least $1- \omega$, where $\omega = r\tilde{\omega}$:
	\begin{align}\label{eq:proof_inter_subspace}
		& \left\| \left( \U_2^\T \G^{(r+1)} \right) \left( \U_1^\T \G^{(r+1)} \right)^{-1} \right\|_2 \nonumber \\
		& \leq  \left(\frac{\lambda_{K+1}}{\lambda_{K}}\right)^r \left\| \left( \U_2^\T \G^{(0)}\right)\left(\U_1^\T \G^{(0)}\right)^{-1}\right\|_2 + \sum_{j=0}^r \left(\frac{\lambda_{K+1}}{\lambda_{K}}\right)^j 2 C_1 \nonumber \\
		& \leq  \left(\frac{\lambda_{K+1}}{\lambda_{K}}\right)^r \left\| \left( \U_2^\T \G^{(0)}\right)\left(\U_1^\T \G^{(0)}\right)^{-1}\right\|_2 + \frac{1}{1- \left(\frac{\lambda_{K+1}}{\lambda_{K}}\right)} 2 C_1. 
	\end{align}
	
	Also, with probability at least $1-\omega$,
	\begin{align}
		& \left\|\U_2^\T \G^{(r+1)} \right\|_2 \leq \left\|\U_2^\T \G^{(r+1)} \left(\U_2^\T \G^{(r+1)}\right)^{-1} \right\|_2 \nonumber \\
		& \leq \left(\frac{\lambda_{K+1}}{\lambda_{K}}\right)^r \left\| \left( \U_2^\T \G^{(0)}\right)\left(\U_1^\T \G^{(0)}\right)^{-1}\right\|_2 + \frac{1}{1- \left(\frac{\lambda_{K+1}}{\lambda_{K}}\right)} 2 C_1 \nonumber \\
		& \leq \left(\frac{\lambda_{K+1}}{\lambda_{K}}\right)^r {\rm tan}(\gamma) + C,
	\end{align}
	where $C = \cO(\nicefrac{\lambda_{K}}{\lambda_{K} - \lambda_{K+1}})$, the first inequality is because $\|\U_1^\T \G^{(r+1)}\|_2 \leq 1 $, and the last inequality is because $\|\U_2^\T \G^{(0)}\|_2  = {\rm sin}(\gamma)$ and $\|\U_1^\T \G^{(r+1)}\|_2  = {\rm cos}(\gamma)$. 

\section{}\label{app:lemma1}
\subsection{Proof of Lemma \ref{lemma:compression_error}}
First, we bound the compression error in iteration $r$.  
 To this end, we take expectation with respect to $\zeta_{\btheta}^{(r)}$ conditioned on all the preceding random variables, i.e., $\cE^{(r,T)}$ and $ \cB^{(r)}$ for the SGD samples taken by the nodes in iteration $r$ and all the random variables used before iteration $r$, respectively.
	\begin{align*}
		& \bbE_{\zeta_{\btheta}^{(r)}}\left[ \left\|\Z_{\btheta_i}^{(r)}\right\|_{\rm F}^2 \big| \cE^{(r,T)}, \cB^{(r)}\right] \\
		& \leq  \bbE_{\zeta_{\btheta}^{(r)}}\left[\left\|{\bm \cC} \left( \bDelta_{\btheta_i}^{(r)} \right) - \bDelta_{\btheta_i}^{(r)}\right\|_{\rm F}^2 \big| \cE^{(r,T)}, \cB^{(r)}\right] \\
		& \stackrel{(a)}{\leq} {(1-\delta)} \left\| \bDelta_{\btheta_i}^{(r)} \right\|_{\rm F}^2 \\
		& \stackrel{(b)}{=} (1-\delta) \left\| \left( \cQ\left( \X_i; \btheta_i^{(r+1)} \right) - \cQ\left( \X_i; \btheta_i^{(r)} \right)\right) - \Z_{\btheta_i}^{(r-1)}\right\|_{\rm F}^2 \\
		& = (1-\delta) \bigg(\left\| \cQ\left( \X_i; \btheta_i^{(r+1)} \right) - \cQ\left( \X_i; \btheta_i^{(r)} \right) \right\|_{\rm F}^2 +\\
		& \left\| \Z_{\btheta_i}^{(r-1)}\right\|_{\rm F}^2 \bigg)  - 2 \left\langle \cQ\left( \X_i; \btheta_i^{(r+1)} \right) - \cQ\left( \X_i; \btheta_i^{(r)} \right),  \Z_{\btheta_i}^{(r-1)} \right\rangle \\
		& \stackrel{(c)}{\leq} (1-\delta) \left( (1+1/\eta)\left\| \cQ\left( \X_i; \btheta_i^{(r+1)} \right) - \cQ\left( \X_i; \btheta_i^{(r)} \right) \right\|_{\rm F}^2 \right) \\
		& + (1-\delta)\left((1 + \eta)\left\|\Z_{\btheta_i}^{(r-1)}\right\|_{\rm F}^2 \right) = (1-\delta) \times \\
		&\left((1+1/\eta) \sum_{t=0}^{T-1} \left(\alpha_{\btheta}^{(r,t)} \right)^2\left\| \g_{\btheta_i}^{(r,t)} \right\|_{\rm F}^2 + (1 + \eta)\left\|\Z_{\btheta_i}^{(r-1)}\right\|_{\rm F}^2 \right)
	\end{align*}
	where (a) follows from Assumption \ref{ass:delta_compressor}; (b) follows from \eqref{eq:deltatheta} and \eqref{eq:compression_error_theta}; and (c) follows by using Young's inequality for the elements of $\cQ( \X_i; \btheta_i^{(r+1)} ) - \cQ( \X_i; \btheta_i^{(r)} )$ and $ \Z_{\btheta_i}^{(r-1)} $ for $\eta > 0$.
	Next, we take expectation with respect to all previous seen random variables included in $\cE^{(r,T)}$ and $\cB^{(r)}$. Denote this total expectation by $\bbE[\cdot]$. This results in the following recursive relation: 
	\begin{align*}
		& \bbE  \left\|\Z_{\btheta_i}^{(r)}\right\|_{\rm F}^2 \leq (1-\delta) \left( (1+1/\eta) \sum_{t=0}^{T-1} \left(\alpha_{\btheta}^{(r,t)} \right)^2 \bbE \left\| \g_{\btheta_i}^{(r,t)} \right\|_{\rm F}^2\right) \\
		& + (1-\delta)\left((1 + \eta) \bbE \left\|\Z_{\btheta_i}^{(r-1)}\right\|_{\rm F}^2 \right).
	\end{align*}
	Consequently, we have
    \begin{align*}
		&\bbE\left[ \left\|\Z_{\btheta_i}^{(r)}\right\|_{\rm F}^2\right] {\leq} \sum_{\ell=0}^r [(1-\delta) (1+\eta)]^{r-\ell} (1-\delta)(1 + 1/\eta) \\
		& \quad\quad\quad \times \left(\sum_{t=0}^{T-1} \left(\alpha_{\btheta}^{(r,t)} \right)^2 \bbE \left\| \g_{\btheta_i}^{(r,t)} \right\|_{\rm F}^2 \right), \\
		& \stackrel{(a)}{\leq} \sum_{\ell=0}^\infty [(1-\delta) (1+\eta)]^{r-\ell} (1-\delta)(1 + 1/\eta)   \\
        & \quad\quad\quad \times \Big(\sum_{t=0}^{T-1} \left( \alpha_{\btheta}^{(r,t)} \right)^2  \bbE \left\| \g_{\btheta_i}^{(r,t)} \right\|_{\rm F}^2 \Big) \\
		& \stackrel{(b)}{\leq} \sum_{\ell=0}^\infty [(1-\delta) (1+\eta)]^{r-\ell} (1-\delta)(1 + 1/\eta) \sum_{t=0}^{T-1} \left( \alpha_{\btheta}^{(r,t)} \right)^2 \sigma^2  
  \end{align*}
  \begin{align*}
		& = \frac{(1-\delta)(1+ 1/\eta)}{\delta - \eta(1-\delta)} \sum_{t=0}^{T-1} \left( \alpha_{\btheta}^{(r,t)} \right)^2 \sigma^2,
    \end{align*}
	where (a) follows by realizing that the sequence with respect to $i$ consists of non-negative numbers and $\Z_{\btheta_i}^{(0)} = \zero$ as $\bDelta_{\btheta_i}^{(0)} = \zero$, and (b) follows from Assumption \ref{ass:sgd_variance}. Taking, $\eta = \frac{\delta}{2(1-\delta)}$ such that $1 + 1/\eta = (2-\delta)/\delta \leq 2/\delta$, the above inequality can be written as
	\begin{align*}
		\bbE \left\| \Z_{\btheta_i}^{(r)} \right\|_{\rm F}^2 \leq \frac{4(1-\delta)}{\delta^2} \sum_{t=0}^{T-1} \left( \alpha_{\btheta}^{(r,t)} \right)^2 \sigma^2.
	\end{align*}
	
	Proof of the compression error with respect to $\G$ update can be obtained by using $\bDelta_{\G}^{(r)} = \G^{(r+1)} - \G^{(r)} - \Z_{\G}^{(r-1)}$ and following the same steps used for bounding compression error with respect to $\btheta_i$ update. The details are omitted here to avoid repetition.
	
\subsection{Proof of \eqref{eq:Y_for_g_update}}\label{app:g_svd}
        We start with \eqref{eq:g_subproblem_proximal} that provides us $\G^{(r+1)}$ as follows:
        \begin{align*}
    		\arg \min_{ \substack{\G^\T\G = \bm I \\\mathbf{1}^\T \G/J = \mathbf{0}} }   \sum_{i=1}^I \frac{1}{2} \left\| \widehat{\M}_i^{(r+1)} - \G \right\|_{\rm F}^2 + \frac{1}{2\alpha_{\G}^{(r)}} \left\| \G - \G^{(r)} \right\|_{\rm F}^2.
        \end{align*}
        We can expand the objective as follows:
        \begin{align*}
    		 & \sum_{i=1}^I \frac{1}{2} \left\| \widehat{\M}_i^{(r+1)}\right\|_{\rm F}^2 - \left\langle \sum_{i=1}^I  \widehat{\M}_i^{(r+1)}, \G \right\rangle + \left\|\G \right\|_{\rm F}^2  \nonumber \\
    		& + \frac{1}{2\alpha_{\G}^{(r)}} \| \G \|_{\rm F}^2 -\left\langle \frac{1}{\alpha_{\G}^{(r)}} \G^{(r)}, \G \right\rangle + \frac{1}{2\alpha_{\G}^{(r)}} \left\|\G^{(r)} \right\|_{\rm F}^2 
        \end{align*}
        \begin{align*}
    		&  = \sum_{i=1}^I \frac{1}{2} \left\| \widehat{\M}_i^{(r+1)}\right\|_{\rm F}^2 - \left\langle \sum_{i=1}^I\widehat{\M}_i^{(r+1)} + \frac{1}{\alpha_{\G}^{(r)}} \G^{(r)}, \G \right\rangle \nonumber \\
    		& + \left\|\G \right\|_{\rm F}^2   + \frac{1}{2\alpha_{\G}^{(r)}} \| \G \|_{\rm F}^2  + \frac{1}{2\alpha_{\G}^{(r)}} \left\|\G^{(r)} \right\|_{\rm F}^2.
    	\end{align*}
    	Since $\G^\T \G = \I$ needs to be satisfied, $\|\G\|_{\rm F}^2 = {\rm Tr}(\G^\T \G)$ is a constant, and $\M_i^{(r+1)}, \forall i$ and $\G^{(r)}$ are also constants for the given objective.
    	Therefore we can re-write the objective as follows:
    	\begin{align*}
    	    & \arg \min_{\substack{\G^\T\G = \bm I \\\mathbf{1}^\T \G/J = \mathbf{0}} }  \frac{1}{2} \left\| \sum_{i=1}^I \widehat{\M}_i^{(r+1)} + \frac{1}{\alpha_{\G}^{(r)}} \G^{(r)} \right\|_{\rm F}^2\\
    	    & - \left\langle \sum_{i=1}^I\widehat{\M}_i^{(r+1)} + \frac{1}{\alpha_{\G}^{(r)}} \G^{(r)}, \G \right\rangle + \left\|\G \right\|_{\rm F}^2  \\
    	    & = \arg \min_{\substack{\G^\T\G = \bm I \\\mathbf{1}^\T \G = \mathbf{0}} }  \frac{1}{2} \left\| \sum_{i=1}^I \widehat{\M}_i^{(r+1)} + \frac{1}{\alpha_{\G}^{(r)}} \G^{(r)}  - \G \right\|_{\rm F}^2.
    	\end{align*}
    	Since $\G^{(r)}$ is already mean-centered, using \cite[Lemma 1]{lyu2020nonlinear} to the above problem gives us \eqref{eq:Y_for_g_update}.
        
% Generated by IEEEtran.bst, version: 1.14 (2015/08/26)

\vfill
\begin{mdframed}
    {
    Proof of Theorems \ref{thm:critical_point}-\ref{thm:convergence_rate} and all Facts are in the supplementary materials (under ``Media'' of the IEEE Xplore page). Also see \url{https://arxiv.org/pdf/2109.12400.pdf}.
    }
\end{mdframed}

    \begin{IEEEbiography}[{\includegraphics[width=1in,height=1.25in,clip,keepaspectratio]{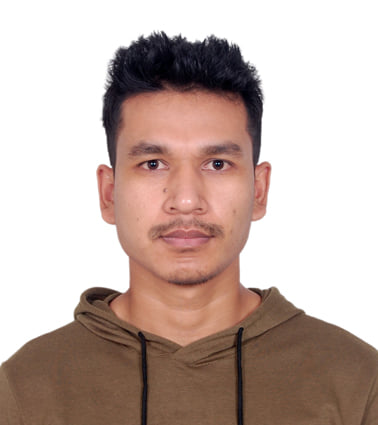}}]
	{Sagar Shrestha} received his B.Eng. in Electronics and Communication Engineering from Pulchowk Campus of Tribhuvan University, Kathmandu, Nepal, in 2016. He is currently working towards a Ph.D.
    degree in Computer Science at the Department of Electrical Engineering and Computer Science, Oregon State University, Corvallis, OR, USA. His current research interests are in the
    broad area of representation learning, statistical machine learning and signal
    processing.
    \end{IEEEbiography}
    % \vspace{-16cm}
    \begin{IEEEbiography}[{\includegraphics[width=1in,height=1.25in,clip,keepaspectratio]{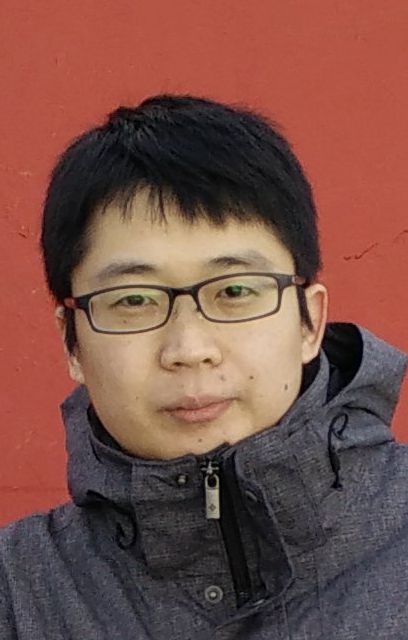}}]
        {Xiao Fu} (Senior Member, IEEE) received the the Ph.D. degree in
Electronic Engineering from The Chinese University of Hong Kong (CUHK), Shatin, N.T., Hong Kong, in 2014. He was a Postdoctoral Associate with the Department of Electrical and Computer Engineering, University of Minnesota, Minneapolis, MN, USA, from 2014 to 2017. Since 2017, he has been an Assistant Professor with the School of Electrical Engineering and Computer Science, Oregon State University, Corvallis, OR, USA. His research interests include the broad area of signal processing and machine learning. 

Dr. Fu received a Best Student Paper Award at ICASSP 2014. His coauthored papers received Best Student Paper Awards from IEEE CAMSAP 2015 and IEEE MLSP 2019, respectively. He received the 2022 IEEE Signal Processing Society (SPS) Best Paper Award and the 2022 IEEE SPS Donald G. Fink Overview Paper Award. He was a recipient of the Outstanding Postdoctoral Scholar Award at University of Minnesota in 2016, and the National Science Foundation (NSF) CAREER Award in 2022.
He serves as a member of the IEEE SPS Sensor Array and Multichannel Technical Committee  (SAM-TC) and the Signal Processing Theory and Methods Technical Committee (SPTM-TC). He is currently an Editor of {\sc Signal Processing} and an Associate Editor of {\sc IEEE Transactions on Signal Processing}. He was a tutorial speaker at ICASSP 2017 and SIAM Conference on Applied Linear Algebra 2021.
    \end{IEEEbiography}
% \begin{bibunit}[IEEEtran]
    
\clearpage
\begin{center}
    {\bf Supplementary Material of ``Communication-Efficient {Federated} Linear and Deep Generalized Canonical Correlation Analysis''}\\
    Sagar Shrestha and Xiao Fu
\end{center}

\section{Proof of Facts}\label{app:facts}
\subsection{Proof of Fact \ref{fact:delta_compressor}}
First, one can see that $\bbE[h(\bDelta(j,k), S)]$
\begin{align}\label{eq:unbiased_h}
% 	&   \nonumber \\
& = \frac{p}{S}\left( 1 - \frac{|\bDelta(j,k)|}{\| \bDelta \|_{\rm max} } + p\right) + \frac{p+1}{S} \left( \frac{|\bDelta(j,k)|}{\| \bDelta \|_{\rm max}} - p \right) \nonumber \\
& = \nicefrac{|\bDelta(j,k)|}{\| \bDelta\|_{\rm max}}. 
\end{align}
With this, {$\bbE[h(\bDelta(j,k), S)^2]$}
\begin{align*}
% 		&  \nonumber \\
    & = \bbE[h(\bDelta(j,k), S)]^2 + \bbE[\left( h(\bDelta(j,k),S) - \bbE[h(\bDelta(j,k), S)]^2 \right)] \nonumber \\
    & \stackrel{(a)}{\leq}\frac{|\bDelta(j,k)|^2}{\| \bDelta\|_{\rm max}^2} + \frac{1}{4S^2} ,
\end{align*}
where the second term in (a) follows by taking the maximum variance of any Bernoulli random variable, and the expectation is with respect to random variable associated with the compressor. 
Now,
\begin{align}\label{eq:c_tilde_bound}
    & \bbE\left[\|\widetilde{\bm \cC}(\bDelta)\|_{\rm F}^2\right]  = \sum_{j=1}^J \sum_{k=1}^K \bbE \left[ \|\bDelta\|_{\rm max}^2 h(\bDelta(j,k), S)^2\right] \nonumber \\
    & \leq \|\bDelta\|_{\rm max}^2 \sum_{j=1}^J \sum_{k=1}^K \left( \frac{|\bDelta(j,k)|^2}{\| \bDelta\|_{\rm max}^2} + \frac{1}{4S^2} \right) \nonumber \\
    & = \left( \frac{\|\bDelta\|_{\rm F}^2}{\|\bDelta\|_{\rm max}^2} + \frac{JK}{4S^2}\right) \|\bDelta \|_{\rm max}^2 = \left( 1 + \frac{JK}{4S^2}\frac{\|\bDelta\|_{\rm max}^2}{\|\bDelta\|_{\rm F}^2}\right) \|\bDelta \|_{\rm F}^2 .
\end{align}

From \eqref{eq:unbiased_h} and \eqref{eq:qsgd}, we can see that 
\begin{align}\label{eq:unbiased_c_tilde}
    \bbE[[\widetilde{\bm \cC}(\bDelta)]_{j,k} ] &= \|\bDelta\|_{\rm max} {\rm sgn}(\bDelta(j, k)) \bbE [ h(\bDelta(j,k), S)]  \nonumber \\
    & = {\rm sgn}(\bDelta(j, k)) | \bDelta(j,k)|  = \bDelta(j,k)  \nonumber \\
    & \implies \bbE[\widetilde{\bm \cC}(\bDelta)] = \bDelta.
\end{align} 

Taking $u = \left( 1 + \frac{JK}{4S^2}\frac{\|\bDelta\|_{\rm max}^2}{\|\bDelta\|_{\rm F}^2}\right) $, consider the following:
\begin{align*}
    & \bbE \left\| \frac{1}{u} \widetilde{\bm \cC}(\bDelta) - \bDelta\right\|_{\rm F}^2 = \frac{1}{u^2} \bbE \left\|\widetilde{\bm \cC}(\bDelta) \right\|_{\rm F}^2 - \frac{2}{u} \left\langle \bbE \left[\widetilde{\bm \cC}(\bDelta)\right] , \bDelta \right\rangle \\
    & + \left\| \bDelta \right\|_{\rm F}^2  \stackrel{(a)}{\leq} \frac{1}{u} \left\| \bDelta \right\|_{\rm F}^2 - \frac{2}{u} \left\| \bDelta \right\|_{\rm F}^2 + \left\| \bDelta \right\|_{\rm F}^2 = \left(1 - \frac{1}{u}\right) \|\bDelta \|_{\rm F}^2,
\end{align*} 
where (a) follows \eqref{eq:c_tilde_bound},  and we have used $ \bbE[\widetilde{\bm \cC}(\bDelta) ] = \bDelta$.

This concludes that ${\bm \cC}(\cdot)$ in \eqref{eq:quantization} is a $\frac{1}{u}$-compressor.
\subsection{Proof of Fact \ref{fact:delta_compressor_condition}}
Since
\begin{align*}
\bbE[\| \widetilde{\bm \cC}(\bDelta) - \bDelta \|_{\rm F}^2] & =  \bbE [\| \widetilde{\bm \cC}(\bDelta)\|_{\rm F}^2] - 2 \langle \bbE [ \widetilde{\bm \cC}(\bDelta) ], \bDelta \rangle  \\
+ \|\bDelta\|_{\rm F}^2
& =  \bbE [\| \widetilde{\bm \cC}(\bDelta)\|_{\rm F}^2] - \|\bbE[ \widetilde{\bm \cC}(\bDelta)]\|_{\rm F}^2,
\end{align*}
where we have used \eqref{eq:unbiased_c_tilde} in the first and the second equality, 
\eqref{eq:c_tilde_bound} leads to
% \begin{equation*}
    $$\bbE[\| \widetilde{\bm \cC}(\bDelta) - \bDelta \|_{\rm F}^2] \leq \left( \nicefrac{JK\|\bDelta\|_{\rm max}^2}{4s^2\|\bDelta\|_{\rm F}^2}\right) \|\bDelta \|_{\rm F}^2 .$$
% \end{equation*}
We can see that for $\widetilde{\bm \cC}(\cdot)$ to be a $\delta$-compressor, $\frac{JK\|\bDelta\|_{\rm max}^2}{4s^2\|\bDelta\|_{\rm F}^2} < 1$ should hold.

\subsection{Proof of Fact~\ref{fact:step_size}}
    Denote $(\alpha_{\G}^{(r)})^2 v(r,t)$ by $\widehat{\alpha}^{(r)}$.
    We wish to show that for any $\alpha^{(r)}$ satisfying the step size rule \eqref{eq:robins_monroe}, 
    if we use $\alpha_{\G}^{(r)} = \alpha_{\btheta}^{(r,t)} = \tau \alpha^{(r)}$ for some appropriate constant $\tau > 0$,  we get that $\widehat{\alpha}^{(r)}>0$ and satisfies step size rule in \eqref{eq:robins_monroe}.
    First, we show that  that $\widehat{\alpha}^{(r)}$ satisfies the step size rule in \eqref{eq:robins_monroe} when $\alpha_{\G}^{(r)} = \alpha_{\btheta}^{(r,t)} = \tau \alpha^{(r)}$. To that end, note that $\widehat{\alpha}^{(r)}$ is a polynomial that can be expressed as $\widehat{\alpha}^{(r)} = C_1 \alpha^{(r)} - C_2 \left(\alpha^{(r)}\right)^2 - C_3 \left(\alpha^{(r)}\right)^3,$
    where $C_i$ for $i=1,2,3$ are constants.

    If $\alpha^{(r)}$ satisfies \eqref{eq:robins_monroe}, one can see that 
    \begin{align*}
        \sum_{r=0}^\infty \widehat{\alpha}^{(r)} & =  C_1 \sum_{r=0}^\infty {\alpha}^{(r)} - C_2 \sum_{r=0}^\infty \left({\alpha}^{(r)}\right)^2 - C_3 \sum_{r=0}^\infty \left({\alpha}^{(r)}\right)^3  = \infty,
    \end{align*}
    since the first term is not bounded.
    We also have $
        \sum_{r=0}^\infty \left(\widehat{\alpha}^{(r)}\right)^2 < \infty,$
    as $ \left(\widehat{\alpha}^{(r)}\right)^2 $ is a polynomial consisting of $\left(\alpha^{(r)} \right)^2$ and higher order terms. Since $\left(\alpha^{(r)} \right)^2$ is summable, the higher order terms must also be summable.
    
    Now, we show that for some appropriate choice of $\tau$, we can ensure that $\widehat{\alpha}^{(r)} > 0, \forall r$, i.e., $\forall r$, 
    \begin{align*}
        & \left(\alpha_{\G}^{(r)}\right)^2 v(r,t) > 0 \Rightarrow \tau \alpha^{(r)} - I \left(\tau \alpha^{(r)}\right)^2 - c \left(\tau \alpha^{(r)}\right)^3 > 0 \\
        & \Rightarrow  \left(\tau \alpha^{(r)}\right)^2 + \frac{I}{c} \tau\alpha^{(r)} < 1 \Rightarrow  \tau \alpha^{(r)} < \nicefrac{I}{c}\left(\sqrt{\nicefrac{c^2}{I^2} + 1} - 1\right)
    \end{align*}
    Since by assumption $\alpha^{(r)} < 1$, using $\tau = \nicefrac{I}{c}(\sqrt{\nicefrac{c^2}{I^2} + 1} - 1)$ satisfies the above inequality.
    
    To ensure that $\alpha_{\btheta}^{(r,t)} \leq \nicefrac{1}{L}, \forall r,t$, we can rescale $\tau$ as  
    $\tau = \nicefrac{I}{c}(\sqrt{\nicefrac{c^2}{I^2} + 1} - 1){\rm min}\{1, 1/L\}.$

\subsection{Proof of Fact~\ref{fact:potential_func}}
    Consider the following:
    \begin{align*}
        & \Gamma^{(r)}  = \sum_{t=0}^{T-1} \sum_{i=0}^I \left\| \nabla_{\btheta_i} f_i(\btheta_i^{(r,t)}, \G^{(r)})\right\|_{\rm F}^2  +  \Big\|I \G^{(r)} \nonumber \\
        & \quad - \sum_{i=1}^I \cQ\left( \X_i; \btheta_i^{(r+1)} \right) + \G^{(r+1)} \bLambda^{(r+1)} + \blambda^{(r+1)} \one^\T \Big\|_{\rm F}^2 \\
        &= \sum_{i=0}^I \left\| \nabla_{\btheta_i} f_i(\btheta_i^{(r)}, \G^{(r)})\right\|_{\rm F}^2  + \sum_{t=1}^{T-1} \sum_{i=0}^I \left\| \nabla_{\btheta_i} f_i(\btheta_i^{(r,t)}, \G^{(r)})\right\|_{\rm F}^2 \\
        & \quad + \big\| \nabla_{\G}f(\btheta^{(r+1)}, \G^{(r)}) + \G^{(r+1)} \bLambda^{(r+1)} + \blambda^{(r+1)} \one^\T \big\|_{\rm F}^2 \\
        & = \|{\bm \varPhi}(\btheta^{(r)}, \G^{(r)})\|_{\rm F}^2 + \sum_{t=1}^{T-1} \sum_{i=0}^I \left\| \nabla_{\btheta_i} f_i(\btheta_i^{(r,t)}, \G^{(r)})\right\|_{\rm F}^2.
    \end{align*}
    Then above implies that when $\Gamma^{(r)} \to 0$, ${\bm \varPhi}(\btheta^{(r)}, \G^{(r)}) \to 0$.

\section{Proof of Theorem \ref{thm:critical_point}}\label{app:proofthm_critical}
{ First,} the $\btheta_i$ update, in iteration $(r,t)$, can be written as:
\begin{equation*}
    \btheta_i^{(r,t+1)} \leftarrow \arg \min_{\btheta_i}  \left\langle \g_{\btheta_i}^{(r,t)}, \btheta_i - \btheta_i^{(r,t)}\right\rangle + \frac{1}{2\alpha_{\btheta}^{(r,t)}} \left\| \btheta_i - \btheta_i^{(r,t)}\right\|_{\rm F}^2.
\end{equation*}
Further, let us define $\widehat{f} (\btheta^{(r+1)}, \{\widehat{\M}_i^{(r+1)}\}_{i=1}^I, \G^{(r)}) = \sum_{i=1}^I \nicefrac{1}{2} \| \widehat{\M}_i^{(r+1)} - \G^{(r)} \|_{\rm F}^2$.
For brevity, we denote $\widehat{f}(\btheta^{(r+1)}, \{\widehat{\M}_i^{(r)}\}_{i=1}^I, \G^{(r)})$ by $\widehat{f}(\widehat{\btheta}^{(r+1)}, \G^{(r)})$, where we use $\widehat{\bm \theta}^{(r+1)}$ to represent the quantized information $\widehat{\M}^{(r+1)}_i$ related to $\bm \theta^{(r+1)}$.
Therefore the proximal update of $\G$-subproblem is as follows:
\begin{align}\label{eq:g_proximal_step}
    & \G^{(r+1)} = \arg\min_{\substack{\G^\T\G=\bm I,\\ \bm 1^\T\bm G=\bm 0 } } \widehat{f}\left(\widehat{\btheta}^{(r+1)}, \G^{(r)}\right) + \frac{1}{2 \alpha_{\G}^{(r)}} \|\G - \G^{(r)}\|_{\rm F}^2. 
\end{align}

\subsection{The $\btheta$-update}
We first show that the $\theta$-update with quantized information can still decrease the objective function as if the exchange information was not quantized.
Since $L$ is the lipschitz continuity constant of $\nabla_{\btheta_i} f_i (\btheta_i, \G)$ for all $i$, we have the following:
\begin{align}\label{eq:q_lipschitz}
    & f(\btheta^{(r,t+1)}, \G^{(r)}) - f(\btheta^{(r,t)}, \G^{(r)}) \leq \sum_{i=1}^I \Big\langle \nabla_{\btheta_i} f_i(\btheta_i^{(r,t)}, \G^{(r)}), \nonumber \\
    & \btheta_i^{(r, t+1)} - \btheta_i^{(r,t)} \Big\rangle + \sum_{i=1}^{I} \frac{L}{2} \left\| \btheta_i^{(r, t+1)} - \btheta_i^{(r, t)} \right\|_{\rm F}^2.
\end{align}
Note that \eqref{eq:q_lipschitz} holds for all pairs of $\bm \theta$ and $\bm \theta'$, which has nothing to do with the updating rule.
On the other hand, under the designed algorithm,
we have $\btheta_i^{(r, t+1)} - \btheta_i^{(r, t)} = - \alpha_{\btheta}^{(r,t)} \bm g_{\btheta_i}^{(r,t)}$. Plugging this relation into \eqref{eq:q_lipschitz}, the right hand side of \eqref{eq:q_lipschitz} becomes
\begin{align*}
    & \sum_{i=1}^I- \alpha_{\btheta}^{(r,t)}\left\langle \nabla_{\btheta_i} f_i\left(\btheta_i^{(r,t)}, \G^{(r)}\right),\g_{\btheta_i}^{(r,t)}\right\rangle +  \frac{L (\alpha_{\btheta}^{(r,t)})^2}{2} \left\| \g_{\btheta_i}^{(r,t)} \right\|_{\rm F}^2.
\end{align*}
Notice that $\g_{\btheta_i}^{(r,t)}$ is the SGD computed with respect to $f_i(\btheta_i^{(r,t)}, \widehat{\G}^{(r)})$ but not $f_i(\btheta_i^{(r,t)}, \G^{(r)})$. However, we can utilize the Lipchitz continuity of the gradient to show that $\nabla_{\btheta_i} f_i(\btheta_i^{(r,t)}, \widehat{\G}^{(r)})$ is not far from $\nabla_{\btheta_i} f_i(\btheta_i^{(r,t)}, \G^{(r)})$---which will help get rid of $\widehat{\bm G}^{(r)}$ in the final expression.
To see this,we first take conditional expectation of the last inequality w.r.t. $\xi^{(r,t)}$ to get
\begin{align*}
    & \bbE_{\xi^{(r,t)}} \left[f\left(\btheta^{(r,t+1)}, \G^{(r)}\right) \big| \cE^{(r,t)}, \cB^{(r)} \right]  - f\left(\btheta^{(r,t)}, \G^{(r)}\right) \nonumber \\
    & \stackrel{(a)}{\leq}- \sum_{i=1}^I \alpha_{\btheta}^{(r,t)} \left\langle \nabla_{\btheta_i} f_i\left(\btheta_i^{(r,t)}, \G^{(r)}\right), \nabla_{\btheta_i} f_i\left(\btheta_i^{(r,t)}, \widehat{\G}^{(r)}\right) \right\rangle \nonumber \\
    & + y(r,t) \leq - \sum_{i=1}^I \alpha_{\btheta}^{(r,t)} \left\| \nabla_{\btheta_i} f_i\left(\btheta_i^{(r,t)}, \G^{(r)}\right) \right\|_{\rm F}^2 + y(r,t) - \\
    & \sum_{i=1}^I \alpha_{\btheta}^{(r,t)}  \bigg\langle \nabla_{\btheta_i} f_i\left(\btheta_i^{(r,t)}, \G^{(r)}\right), \\
    & \nabla_{\btheta_i} f_i\left(\btheta_i^{(r,t)}, \widehat{\G}^{(r)}\right) - \nabla_{\btheta_i} f_i\left(\btheta_i^{(r,t)}, \G^{(r)}\right)\bigg\rangle. \\
    & \stackrel{(b)}{\leq} - \sum_{i=1}^I \alpha_{\btheta}^{(r,t)} \left( 1- \frac{\eta}{2} \right)\left\| \nabla_{\btheta_i} f_i\left(\btheta_i^{(r,t)}, \G^{(r)}\right) \right\|_{\rm F}^2 + y(r,t) \nonumber \\
    & + \frac{\alpha_{\btheta}^{(r,t)}}{2 \eta}\sum_{i=1}^I \left\|\nabla_{\btheta_i} f_i\left(\btheta_i^{(r,t)}, \widehat{\G}^{(r)}\right) - \nabla_{\btheta_i} f_i\left(\btheta_i^{(r,t)}, \G^{(r)}\right) \right\|_{\rm F}^2 \nonumber \\ 	
    & \stackrel{(c)}{\leq} - \sum_{i=1}^I \alpha_{\btheta}^{(r,t)} \left( 1- \frac{\eta}{2} \right)\left\| \nabla_{\btheta_i} f_i\left(\btheta_i^{(r,t)}, \G^{(r)}\right) \right\|_{\rm F}^2 + y(r,t)
\end{align*}
\begin{align*}
    & + \frac{\alpha_{\btheta}^{(r,t)}L^2}{2 \eta}\sum_{i=1}^I \left\| \widehat{\G}^{(r)} - \G^{(r)}\right\|_{\rm F}^2,
\end{align*}
where $y(r,t) = \nicefrac{I L \left(\alpha_{\btheta}^{(r,t)}\right)^2 \sigma^2}{2}$, $\eta>0$. The right hand side of (a) is obtained by using Fact \ref{fact:unbiased_gq} for the first term and Assumption \ref{ass:sgd_variance} for the last term, respectively, (b) is obtained by applying Young's inequality (cf. \cite{young1912classes}) onto the last term of the previous inequality, and (c) is obtained by Assumption \ref{ass:lipschitz}. 

Taking expectation w.r.t. the filtration $\cE^{(r,t)}$ on both sides of the last inequality, we get
\begin{align}
    & \bbE_{\xi^{(r,t)}\cE^{(r,t)}} \left[f\left(\btheta^{(r,t+1)}, \G^{(r)}\right) \big| \cB^{(r)}\right]  - \bbE_{\cE^{(r,t)}} \left[f\left(\btheta^{(r,t)}, \G^{(r)}\right)\big| \cB^{(r)} \right] \nonumber \\
    & \leq - \sum_{i=1}^I \alpha_{\btheta}^{(r,t)} \left( 1- \frac{\eta}{2} \right)\bbE_{\cE^{(r,t)}} \left[ \left\| \nabla_{\btheta_i} f_i\left(\btheta_i^{(r,t)}, \G^{(r)}\right) \right\|_{\rm F}^2 \big| \cB^{(r)}\right] \nonumber \\
    & \quad + y(r,t) + \frac{\alpha_{\btheta}^{(r,t)}L^2}{2 \eta}\sum_{i=1}^I \left\| \widehat{\G}^{(r)} - \G^{(r)}\right\|_{\rm F}^2.
\end{align}
In the last inequality, we have obtained a bound for the ``sufficient decrease'' resulted from the update of $\btheta^{(r,t)}$ to $\btheta^{(r,t+1)}$. Summing the above from $t = 0 $ to $ t=T-1$ and using $\eta=1$ leads to
\begin{align}\label{eq:expected_q_update}
    & \bbE_{\cE^{(r,T)}} \left[f\left(\btheta^{(r+1)}, \G^{(r)}\right) \big| \cB^{(r)} \right]  - f(\btheta^{(r)}, \G^{(r)}) \nonumber \\
    & \leq -\frac{1}{2} \sum_{t=0}^{T-1}\sum_{i=1}^I  \alpha_{\btheta}^{(r,t)} \bbE_{\cE^{(r,T)}} \left[ \left\| \nabla_{\btheta_i} f_i\left(\btheta_i^{(r,t)}, \G^{(r)}\right) \right\|_{\rm F}^2 \big| \cB^{(r)} \right] \nonumber \\
    & + \sum_{t=0}^{T-1} y(r,t) + \sum_{t=0}^{T-1} \frac{\alpha_{\btheta}^{(r,t)}I L^2}{2}\left\|\Z_{\G}^{(r-1)}\right\|_{\rm F}^2.
\end{align}
Notice that we have used $\Z_{\G}^{(r-1)} = \widehat{\G}^{(r)} - \G^{(r)}$ in the above inequality. This provides us a bound for the sufficient decrease of the cost function (in conditional expectation) when updating from $\btheta^{(r)}$ to $\btheta^{(r+1)}$.
\subsection{The $\G$-update}
The proposed method tries to minimize the objective value $\widehat{f}(\widehat{\btheta}^{(r+1)}, \G^{(r+1)}) $ instead of the true objective, $f(\btheta^{(r+1)}, \G^{(r+1)})$. However, we hope to bound the change in the true objective after each $\G$-iteration. To this end, the following holds:
\begin{align}\label{eq:inter_bound_g}
    & {f}(\btheta^{(r+1)}, \G^{(r+1)}) - {f}(\btheta^{(r+1)}, \G^{(r)}) \leq {f}(\btheta^{(r+1)}, \G^{(r+1)})  \nonumber\\
    & - \widehat{f}(\widehat{\btheta}^{(r+1)}, \G^{(r+1)})   + \widehat{f}(\widehat{\btheta}^{(r+1)}, \G^{(r+1)}) - \widehat{f}(\widehat{\btheta}^{(r+1)}, \G^{(r)}) \nonumber \\
    & + \widehat{f}(\widehat{\btheta}^{(r+1)}, \G^{(r)}) - {f}(\btheta^{(r+1)}, \G^{(r)}).
\end{align}
To bound the right hand side, we use \eqref{eq:g_proximal_step} to get
\begin{align}\label{eq:inter_f_bar}
& \widehat{f}(\widehat{\btheta}^{(r+1)}, \G^{(r+1)}) - \widehat{f}(\widehat{\btheta}^{(r+1)}, \G^{(r)}) \nonumber \\
& \leq   -\nicefrac{1}{2\alpha_{\G}^{(r)}} \big\| \G^{(r+1)} - \G^{(r)}\big\|_{\rm F}^2.
\end{align}
In any iteration $r$ and for any $\G$, the difference between $f(\btheta^{(r+1)}, \G)$ and $\widehat{f}(\widehat{\btheta}^{(r+1)}, \G)$ is expressed as follows:
\begin{align}
    &f(\btheta^{(r+1)}, \G) - \widehat{f}(\widehat{\btheta}^{(r+1)}, \G) \nonumber \\
    & = \sum_{i=1}^I \left\| \M_i^{(r+1)} - \G \right\|_{\rm F}^2  - \sum_{i=1}^I \left\| \widehat{\M}_i^{(r+1)}  - \G \right\|_{\rm F}^2.
\end{align}
This leads to $ f(\btheta^{(r+1)}, \G^{(r+1)}) - \widehat{f}(\widehat{\btheta}^{(r+1)}, \G^{(r+1)}) - (f(\btheta^{(r+1)}, \G^{(r)}) - \widehat{f}(\widehat{\btheta}^{(r+1)}, \G^{(r)} )) $
\begin{align}\label{eq:inter_f_f_fbar}
    & =  \sum_{i=1}^I \bigg( \| \M_i^{(r+1)} - \G^{(r)} \|_{\rm F}^2 - \| \widehat{\M}_i^{(r+1)} -\G^{(r)} \|_{\rm F}^2 \nonumber \\
    & - \left( \| \M_i^{(r+1)} - \G^{(r+1)} \|_{\rm F}^2 - \|\widehat{\M}_i^{(r+1)} - \G^{(r+1)}\|_{\rm F}^2\right) \bigg) \nonumber \\
    & \stackrel{(a)}{=} \sum_{i=1}^I 2 \left\langle \widehat{\M}_i^{(r+1)} - \M_i^{(r+1)}, \G^{(r)} - \G^{(r+1)} \right\rangle \nonumber \\
    & \stackrel{(b)}{\leq}\sum_{i=1}^I \left(\left\| \Z_{\btheta_i}^{(r)} \right\|_{\rm F}^2 + \left\| \G^{(r+1)} - \G^{(r)} \right\|_{\rm F}^2\right),
\end{align}
where (a) follows by expanding and simplifying the previous equality, and (b) follows from Young's inequality (cf. \cite{young1912classes}). Using \eqref{eq:inter_f_bar}, \eqref{eq:inter_f_f_fbar} and \eqref{eq:inter_bound_g}, we obtain\\
    $f\left(\btheta^{(r+1)}, \G^{(r+1)}\right) - f\left(\btheta^{(r+1)}, \G^{(r)}\right) $
\begin{align}
    & \leq \sum_{i=1}^I \left\| \Z_{\btheta_i}^{(r)} \right\|_{\rm F}^2 - \left( \nicefrac{1}{2\alpha_{\G}^{(r)}} - I\right) \left\| \G^{(r+1)} - \G^{(r)} \right\|_{\rm F}^2.\nonumber
\end{align}
Now, taking conditional expectation with respect to ${\zeta_{\btheta}^{(r)}}$ given the seen random variables before updating $\G^{(r)}$ and using step size rule in \eqref{eq:robins_monroe}, we get
    $\bbE_{\zeta_{\btheta}^{(r)}} \left[f\left(\btheta^{(r+1)}, \G^{(r+1)}\right)\big| \cE^{(r,T)}, \cB^{(r)}\right] - f\left(\btheta^{(r+1)}, \G^{(r)}\right) $
\begin{align*}
    & \leq \sum_{i=1}^I \bbE_{\zeta_{\btheta}^{(r)}}\left[ \left\|\Z_{\btheta_i}^{(r)}\right\|_{\rm F}^2 \big| \cE^{(r,T)}, \cB^{(r)}\right]  \nonumber \\
    & - \left( \nicefrac{1}{2\alpha_{\G}^{(r)}} - I\right)  \bbE_{\zeta_{\btheta}^{(r)}} \left[\left\|\G^{(r+1)} - \G^{(r)}\right\|_{\rm F}^2\big| \cE^{(r,T)}, \cB^{(r)}\right].
\end{align*}
In addition,
taking expectation over the filtration $\cE^{(r,T)}$, we get
\begin{align}\label{eq:expected_g_update}
    & \bbE_{\cE^{(r,T)}} \left[ \bbE_{\zeta_{\btheta}^{(r)}} f(\btheta^{(r+1)}, \G^{(r+1)})  - f(\btheta^{(r+1)}, \G^{(r)}) \big| \cB^{(r)} \right]\nonumber \\
    & \leq \sum_{i=1}^I \bbE_{\zeta_{\btheta}^{(r)} \cE^{(r,T)}}\left[ \|\Z_{\btheta_i}^{(r)}\|_{\rm F}^2 \big| \cB^{(r)}\right]  \\
    & - \left( \nicefrac{1}{2\alpha_{\G}^{(r)}} - I\right)  \bbE_{\zeta_{\btheta}^{(r)} \cE^{(r,T)}} \left[\|\G^{(r+1)} - \G^{(r)}\|_{\rm F}^2\big| \cB^{(r)}\right],\nonumber
\end{align}
which characterizes the sufficient decrease induced by the $\G$-update.

\subsection{Putting Together}
Now, we combine the change in objective values in \eqref{eq:expected_q_update} with \eqref{eq:expected_g_update} and get
% \begin{align}
\begin{align}\label{eq:full_update}
    & \bbE_{\zeta_{\btheta}^{(r)}\cE^{(r,T)}} \left[ f\left(\btheta^{(r+1)}, \G^{(r+1)}\right) |  \cB^{(r)} \right] - f(\btheta^{(r)}, \G^{(r)}) \nonumber \\
    & \leq - \frac{1}{2}\sum_{t=0}^{T-1}\sum_{i=1}^I \alpha_{\btheta}^{(r,t)} \bbE_{\cE^{(r,T)}} \left[\| \nabla_{\btheta_i} f_i\left(\btheta_i^{(r,t)}, \G^{(r)}\right) \|_{\rm F}^2 \big| \cB^{(r)} \right] \nonumber \\
    & + \sum_{t=0}^{T-1} y(r,t)  + \sum_{t=0}^{T-1} \frac{\alpha_{\btheta}^{(r,t)}I L^2}{2} \left\| \Z_{\G}^{(r-1)}\right\|_{\rm F}^2 \nonumber \\
    &+  \sum_{i=1}^I \bbE_{\zeta_{\btheta}^{(r)} \cE^{(r,T)}}\left[ \left\|\Z_{\btheta_i}^{(r)}\right\|_{\rm F}^2 \big| \cB^{(r)}\right] \nonumber \\
% \end{align}
    & - \left( \nicefrac{1}{2\alpha_{\G}^{(r)}} - I\right)   \bbE_{\zeta_{\btheta}^{(r)} \cE^{(r,T)}} \left[\|\G^{(r+1)} - \G^{(r)}\|_{\rm F}^2\big| \cB^{(r)}\right]. 
\end{align}
Taking expectation with respect to filtration $\cB^{(r)}$ on both sides and using Lemma \ref{lemma:compression_error}, we have the following:
\begin{align}
    & \bbE \left[ f\left(\btheta^{(r+1)}, \G^{(r+1)}\right) \right] - \bbE \left[f(\btheta^{(r)}, \G^{(r)})\right] \nonumber \\
    & \leq - \frac{1}{2}\sum_{t=0}^{T-1}\sum_{i=1}^I \alpha_{\btheta}^{(r,t)} \bbE\left[\left\| \nabla_{\btheta_i} f_i\left(\btheta_i^{(r,t)}, \G^{(r)}\right) \right\|_{\rm F}^2 \right] \nonumber \\
    & + \sum_{t=0}^{T-1} \left( w(r,t) + \frac{2\alpha_{\btheta}^{(r,t)} I L^2 (1-\delta)}{ \delta^2} \bbE \left\| \G^{(r)} - \G^{(r-1)}\right\|_{\rm F}^2 \right) \nonumber \\
    & - \left( \nicefrac{1}{2\alpha_{\G}^{(r)}} - I\right)   \bbE_{\zeta_{\btheta}^{(r)} \cE^{(r,T)}} \left[\|\G^{(r+1)} - \G^{(r)}\|_{\rm F}^2\big| \cB^{(r)}\right].  \nonumber
\end{align}
Take total expectation of both sides.
Then, summing $r$ from $r=0$ to $r=R$ and rearranging terms lead to 
\begin{align}\label{eq:sufficient_decrease}
    & \sum_{r=0}^{R} \sum_{t=0}^{T-1}\sum_{i=1}^I \frac{\alpha_{\btheta}^{(r,t)}}{2} \bbE \left[\left\| \nabla_{\btheta_i} f_i\left( \btheta_i^{(r,t)},\G^{(r)} \right) \right\|_{\rm F}^2 \right] \nonumber \\
    & + \sum_{r=0}^{R} \left( \nicefrac{1}{2\alpha_{\G}^{(r)}} - I\right)  \bbE \left[\left\|\G^{(r+1)} - \G^{(r)}\right\|_{\rm F}^2\right] \nonumber \\
    & \leq f\left( \btheta^{(0)}, \G^{(0)} \right) - \bbE \left[ f\left( \btheta^{(R+1)}, \G^{(R+1)} \right) \right] + \sum_{r=0}^{R} \sum_{t=0}^{T-1} w(r,t) \nonumber \\
    & + \sum_{r=1}^{R} \sum_{t=0}^{T-1} \frac{2\alpha_{\btheta}^{(r,t)} I L^2 (1-\delta)}{ \delta^2} \bbE \left\| \G^{(r)} - \G^{(r-1)}\right\|_{\rm F}^2.\nonumber \\
     & \implies \sum_{r=0}^{R} \sum_{t=0}^{T-1}\sum_{i=1}^I \frac{\alpha_{\btheta}^{(r,t)}}{2} \bbE \left[\left\| \nabla_{\btheta_i} f_i\left( \btheta_i^{(r,t)},\G^{(r)} \right) \right\|_{\rm F}^2 \right] \nonumber \\
    & +  \sum_{r=0}^R v(r,t) \bbE \left\| \G^{(r+1)} - \G^{(r)}\right\|_{\rm F}^2 \leq f\left( \btheta^{(0)}, \G^{(0)} \right) \nonumber\\
    & - \bbE \left[ f\left( \btheta^{(R+1)}, \G^{(R+1)} \right) \right]  + \sum_{r=0}^{R} \sum_{t=0}^{T-1} w(r,t) ,
\end{align}
where $v(r,t)=( \nicefrac{1}{2\alpha_{\G}^{(r)}} - I - \sum_{t=0}^{T-1} \nicefrac{2\alpha_{\btheta}^{(r+1,t)} I L^2 (1-\delta)}{ \delta^2} ).$
{Since $f$ is lower bounded and $\alpha_{\btheta}^{(r,t)}$ satisfies the step-size rule in \eqref{eq:robins_monroe}, as $R \to \infty$ the left hand side of the above equation is bounded (i.e., $ < \infty$).

Since the left hand side of the above inequality contains the sum of two positive terms, the two terms should individually be bounded, i.e.,}
\begin{equation}\label{eq:theta_conv}
    \sum_{r=0}^{\infty} \sum_{t=0}^{T-1}\sum_{i=1}^I \frac{\alpha_{\btheta}^{(r,t)}}{2} \bbE \left[\left\| \nabla_{\btheta_i} f_i\left(\btheta_i^{(r,t)}, \G^{(r)}\right) \right\|_{\rm F}^2 \right] < \infty,
\end{equation}
\begin{equation}
    \begin{aligned}
        &\sum_{r=0}^{\infty} v(r,t) \bbE \left[\left\|\G^{(r+1)} - \G^{(r)}\right\|_{\rm F}^2\right] < \infty. 
    \end{aligned}
\end{equation}
Let us define $\H^{(r)} = \nicefrac{1}{\alpha_{\G}^{(r)}} ( \G^{(r)} - \G^{(r+1)})$. Then
\begin{align} \label{eq:h_conv}
    \sum_{r=0}^\infty \left(\alpha_{\G}^{(r)} \right)^2 v(r,t) \bbE [ \| \H^{(r)} \|_{\rm F}^2 ] < \infty.
\end{align}
Since we have assumed that $(\alpha_{\G}^{(r)} )^2 v(r,t)$ satisfies step size rule in \eqref{eq:robins_monroe}, from \eqref{eq:h_conv} and \eqref{eq:theta_conv}, and using \cite[Lemma {A.5}]{mairal2013stochastic}

\begin{subequations}\label{eq:lim_inf}
\begin{align}
    \lim_{r \to \infty} \inf \bbE [ \| \H^{(r)} \|_{\rm F}^2 &= 0 \\
    \lim_{r \to \infty} \inf \bbE \left[\left\| \nabla_{\btheta_i} f_i\left(\btheta_i^{(r,t)}, \G^{(r)}\right) \right\|_{\rm F}^2 \right] &= 0, \quad \forall, i, t.
\end{align}
\end{subequations}

\subsection{Critical Point Convergence}
First, observe that the update rule in \eqref{eq:g_proximal_step} implies that there exists a $\bLambda^{(r+1)}$ and $\blambda^{(r+1)}$ such that the following optimality condition holds
\begin{align}\label{eq:g_diff}
    & \nabla_{\G} \widehat{f}\left(\widehat{\btheta}^{(r+1)}, \G^{(r)}\right) + \nicefrac{1}{\alpha_{\G}^{(r)}}(\G^{(r+1)} - \G^{(r)}) + \G^{(r+1)} \bLambda^{(r+1)} \nonumber  \\
    &  + \blambda^{(r+1)} \one^\T = \mathbf{0} \quad \Rightarrow \quad  I \G^{(r)} - \sum_{i=1}^I \widehat{\M}_i^{(r+1)} \\
    & + \nicefrac{1}{\alpha_{\G}^{(r)}}(\G^{(r+1)} - \G^{(r)}) + \G^{(r+1)} \bLambda^{(r+1)} + \blambda^{(r+1)} \one^\T  = \mathbf{0} 
    \nonumber .
\end{align}
Using the definition in \eqref{eq:subdifferential}, $\| \bm \varPhi(\btheta^{(r+1)}, \G^{(r+1)})\|_{\rm F}^2 =$
\begin{align*}
    &  \sum_{i=1}^{I} \left\| \nabla_{\btheta_i} f(\btheta^{(r+1)}, \G^{(r+1)}) \right\|_{\rm F}^2 +  \iota  ~\leq~ 2\widehat{\iota} +\\
    &  \sum_{i=1}^{I} \Big\| \nabla_{\btheta_i} f(\btheta^{(r+1)}, \G^{(r+1)}) \Big\|_{\rm F}^2   + 2 \Big\| \sum_{i=1}^I  (\widehat{\M}_i^{(r+1)} -   \M_i^{(r+1)}) \Big\|_{\rm F}^2 \\
% \end{align*}
% \begin{align*}
    & \stackrel{(a)}{\leq}\sum_{i=1}^{I} \left\| \nabla_{\btheta_i} f(\btheta^{(r+1)}, \G^{(r+1)}) \right\|_{\rm F}^2 + 2\left\| \H^{(r)} \right\|_{\rm F}^2 \\
    & + 2^I  \sum_{i=1}^I \left\| \Z_{\btheta_i}^{(r)} \right\|_{\rm F}^2
    \leq \sum_{i=1}^{I} \left\| \nabla_{\btheta_i} f(\btheta^{(r+1)}, \G^{(r+1)}) \right\|_{\rm F}^2  \\
    &+ 2\left\| \H^{(r)} \right\|_{\rm F}^2 + \frac{2^I 4(1-\delta) ( \alpha_{\btheta}^{(r,t)} )^2 T \sigma^2}{\delta^2},
\end{align*}
where $\iota = \|I \G^{(r)} - \sum_{i=1}^I \M_i^{(r+1)} + \G^{(r+1)} \bLambda^{(r+1)}  + \blambda^{(r+1)} \one^\T  \|_{\rm F}^2$  and $\widehat{\iota} = \|I \G^{(r)} - \sum_{i=1}^I \widehat{\M}_i^{(r+1)} + \G^{(r+1)} \bLambda^{(r+1)}  + \blambda^{(r+1)} \one^\T  \|_{\rm F}^2$, (a) follows from \eqref{eq:g_diff}. Taking total expectation, with respect to all random variables up to iteration $r+1$ on both sides, and taking the $\lim_{r \to \infty} \inf$ on both sides leads to $ \lim_{r \to \infty} \inf \left\| \bm \varPhi\left(\btheta^{(r+1)}, \G^{(r+1)}\right)\right\|_{\rm F}^2 \leq$
\begin{align*}
    & \sum_{i=1}^{I} \lim_{r \to \infty} \inf \left\| \nabla_{\btheta_i} f(\btheta^{(r+1)}, \G^{(r+1)}) \right\|_{\rm F}^2 + 2 \lim_{r \to \infty} \inf \left\| \H^{(r)} \right\|_{\rm F}^2 \\
    &  + 2^I   \sum_{i=1}^I \frac{4(1-\delta)}{\delta^2} \lim_{r \to \infty} \inf \left( \alpha_{\btheta}^{(r,t)} \right)^2 T \sigma^2.
\end{align*}
From \eqref{eq:lim_inf} and \eqref{eq:robins_monroe}, one can see that the right hand side is zero. Therefore
$	\lim_{r \to \infty} \inf \| \bm \varPhi(\btheta^{(r+1)}, \G^{(r+1)})\|_{\rm F}^2 = 0 .$
This completes the proof. 

\section{Proof of Theorem \ref{thm:convergence_rate}}\label{app:proofthm_rate}
We have from \eqref{eq:sufficient_decrease} that 	
\begin{align*}
    &  \sum_{r=0}^{R} \sum_{t=0}^{T-1}\sum_{i=1}^I \frac{\alpha_{\btheta}}{2} \bbE \left[\left\| \nabla_{\btheta_i} f_i\left( \btheta_i^{(r,t)},\G^{(r)} \right) \right\|_{\rm F}^2 \right] \\
    & +  \sum_{r=0}^R\left( \frac{1}{2\alpha_{\G}} -I - \frac{2T\alpha_{\btheta} I L^2 (1-\delta)}{ \delta^2} \right) \bbE \left\| \G^{(r+1)} - \G^{(r)}\right\|_{\rm F}^2 \nonumber \\
    & \leq f\left( \btheta^{(0)}, \G^{(0)} \right) - \bbE \left[ f\left( \btheta^{(R+1)}, \G^{(R+1)} \right) \right] 
\end{align*}
\begin{align*}
    & + \sum_{r=0}^{R} \frac{IT\alpha_{\btheta}^2 \sigma^2 (\delta^2 L + 8(1-\delta))}{2\delta^2}.
\end{align*}

Let $v = \frac{1}{2\alpha_{\G}} -I -\frac{2T\alpha_{\btheta} I L^2 (1-\delta)}{ \delta^2} $. Using \eqref{eq:g_diff} in the above inequality leads to 
\begin{align}\label{eq:inter_rate}
    & \sum_{r=0}^{R} \sum_{t=0}^{T-1}\sum_{i=1}^I \frac{\alpha_{\btheta}}{2} \bbE \left[\left\| \nabla_{\btheta_i} f_i\left( \btheta_i^{(r,t)},\G^{(r)} \right) \right\|_{\rm F}^2 \right] \nonumber  + \sum_{r=0}^{R}  v \alpha_{\G}^2 \bbE[~\widehat{\iota}~]  \\
    & \leq f\left( \btheta^{(0)}, \G^{(0)} \right) - \bbE \left[ f\left( \btheta^{(R+1)}, \G^{(R+1)} \right) \right]  \nonumber \\
    & + \sum_{r=0}^{R}\frac{TI\alpha_{\btheta}^2 \sigma^2 (\delta^2 L + 8(1-\delta))}{2\delta^2}. 
\end{align}
Consider the following chain of inequalities:
\begin{align*}
& 2 \bbE [~\iota~] \stackrel{(a)}{\geq} \bbE[~ \iota ~]- 2 \bbE \bigg\| \sum_{i=1}^I\widehat{\M}_i^{(r+1)} - \sum_{i=1}^I\M^{(r+1)}\bigg\|_{\rm F}^2 \nonumber \\ 
    & \stackrel{(b)}{\geq}\bbE[~ \iota ~] - 2^I \sum_{i=1}^I \frac{4(1-\delta)}{\delta^2} \left( \alpha_{\btheta} \right)^2 T \sigma^2,
\end{align*}
where (a) follows from Young's inequality, using $\eta = 1$, and (b) follows from Lemma \ref{lemma:compression_error}.
Using the above in \eqref{eq:inter_rate} leads to 
\begin{align}
    & \sum_{r=0}^{R} \sum_{t=0}^{T-1}\sum_{i=1}^I \frac{\alpha_{\btheta}}{2} \bbE \left[\left\| \nabla_{\btheta_i} f_i\left( \btheta_i^{(r,t)},\G^{(r)} \right) \right\|_{\rm F}^2 \right] \nonumber \\
    &+ \sum_{r=0}^{R}  \frac{v \alpha_{\G}^2}{2} \widehat{\iota} \leq f\left( \btheta^{(0)}, \G^{(0)} \right) - \bbE \left[ f\left( \btheta^{(R+1)}, \G^{(R+1)} \right) \right]  \nonumber \\
    & +   \frac{RIT\alpha_{\btheta}^2 \sigma^2 (\delta^2 L + 8(1-\delta))}{2\delta^2} + 2^{I-1}   \frac{4R I (1-\delta)\alpha_{\btheta}^2 v\alpha_{\G}^2 T \sigma^2}{\delta^2} . \nonumber
\end{align}
Let $d = {\rm min} \left\{  \frac{\alpha_{\btheta}}{2}, \frac{v \alpha_{\G}^2}{2} \right\}$. Then $ \frac{1}{R+1}\sum_{r=0}^R d \bbE[\Gamma^{(r)}] $
\begin{align}\label{eq:rate_expression_proof}
    & \leq \frac{1}{R+1} \left(f\left( \btheta^{(0)}, \G^{(0)} \right) - \bbE \left[ f\left( \btheta^{(R+1)}, \G^{(R+1)} \right) \right] \right) \nonumber\\
    & + \frac{IT\alpha_{\btheta}^2 \sigma^2 (\delta^2 L + 8(1-\delta))}{2\delta^2}  + \frac{2^{I+1}IT(1-\delta)\alpha_{\btheta}^2 v\alpha_{\G}^2 \sigma^2}{\delta^2}.
\end{align}
We have $\alpha_{\btheta} = \alpha_{\G} = \alpha =  \nicefrac{1}{\sqrt{R+1}} (\nicefrac{I}{c} (\sqrt{1 + \nicefrac{c}{I^2}} - 1)){\rm min}\{ 1/L, 1 \}$. 
Let $\alpha = \nicefrac{1}{\sqrt{R+1}} C_1$, where $C_1$ is a  constant. Using Fact \ref{fact:step_size}, we can see that $v \alpha_{\G}^2 \geq 0$ and can be upper-bounded by a constant, i.e., $v \alpha_{\G}^2  \leq C_2$, where $C_2$ is a constant. 
Therefore, one can write \eqref{eq:rate_expression_proof} as $\frac{1}{R+1}\sum_{r=0}^R d \bbE[\Gamma^{(r)}]$
\begin{align*}
    & \leq \frac{1}{R+1} \left(f\left( \btheta^{(0)}, \G^{(0)} \right) - \bbE \left[ f\left( \btheta^{(R+1)}, \G^{(R+1)} \right) \right] \right) \nonumber \\
    & + \frac{IT C_1^2 \sigma^2 (\delta^2 L + 8(1-\delta))}{2\delta^2(R+1)}  + \frac{2^{I+1}IT(1-\delta) C_1^2 C_2\sigma^2}{\delta^2(R+1)}. 
\end{align*} 
Further, $d$ is of the order $\cO(\nicefrac{1}{\sqrt{R+1}})$. Therefore the above inequality leads to 
\begin{align*}
    & \frac{1}{R+1}\sum_{r=0}^R \bbE[\Gamma^{(r)}] \nonumber \\
    & \leq  \cO\left(\frac{1}{\sqrt{R+1}}\right) \bigg[f\left( \btheta^{(0)}, \G^{(0)}\right)  - \bbE \left[ f\left( \btheta^{(R+1)}, \G^{(R+1)} \right) \right] \\
    & + \frac{IT \sigma^2 (\delta^2 L + 8(1-\delta))}{2\delta^2} \nonumber + \frac{2^{I+1} IT(1-\delta)\sigma^2}{\delta^2} \bigg]. \nonumber
\end{align*}
This completes the proof.

	\ifCLASSOPTIONcaptionsoff
	\newpage
	\fi
\end{document}